\documentclass[11pt,draftclsnofoot,onecolumn]{IEEEtran}

\usepackage{graphicx}
\usepackage{amsmath,amssymb} 
\usepackage{multirow}
\usepackage{color}
\usepackage{colortbl}
\usepackage{xcolor}
\usepackage{rotating}
\usepackage{setspace}

\graphicspath{{ss_figure/}}

\title{A Data-driven Approach for Human Pose Tracking Based on Spatio-temporal Pictorial Structure}
\author{Soumitra Samanta and Bhabatosh Chanda
\thanks{ECSU, Indian Statistical Institute, Kolkata 700108, India.}}

\begin{document}

\maketitle

\begin{abstract}
In this paper, we present a data-driven approach for human pose tracking in video data. We formulate the human pose tracking problem
as a discrete optimization problem based on spatio-temporal pictorial structure model and
solve this problem in a greedy framework very efficiently. We propose the model to track the human pose by combining the human pose estimation from single image
and traditional object tracking in a video. Our pose tracking objective function consists of the following
terms: likeliness of appearance of a part within a frame, temporal displacement of the part from previous frame to the current
frame, and the spatial dependency of a part with its parent in the graph structure. Experimental evaluation on benchmark datasets
(VideoPose2, Poses in the Wild and Outdoor Pose) as well as on our newly build ICDPose dataset shows the usefulness of our proposed method.
\end{abstract}

\section{Introduction}
%
%
%
%
Human pose tracking is an important problem in computer vision due to its application in human action recognition and
surveillance from video data. Visual appearance of any human action is a sequence of various human poses. We propose that if we can track
those poses, then human action could be determined accurately. Human body is a symmetric articulated structure consisting of several parts
connected pairwise.
We define human pose as a combination of $n$ parts ($n$ depends on visible portion of a body). Let $p_{i}$ denotes the $i$-th body part and
$\mathbf{x}_{i}^{t}$ = $(u_{i}^{t}, v_{i}^{t})$ ($i = 1:n$) its position in $t^{th}$ frame. Where $(u, v)$ is the image co-ordinate.
Our aim is to track human pose in a video, i.e., to estimate the positions of these parts in every frame of the video.
We write this as an optimization problem given by
\begin{equation}
    \mathbf{x}_{1}^{t*}, ..., \mathbf{x}_{n}^{t*} = arg \min_{\mathbf{x}_{1}^{t}, ..., \mathbf{x}_{n}^{t}}f(\mathbf{x}_{1}^{t}, ..., \mathbf{x}_{n}^{t} | \mathbf{x}_{1}^{t-1}, ..., \mathbf{x}_{n}^{t-1})
    \label{eqn:chap5:x_1_t_*_to_x_n_t_*}
\end{equation}
In general, due to exponential search space of all body parts in all frames in a video, solving Eq.~(\ref{eqn:chap5:x_1_t_*_to_x_n_t_*})
is NP-hard. Researchers impose various constraints to limit the search space. Another major challenge in detecting these parts and subsequently
the pose structure is \textit{double counting}, which occurs due to symmetry in human body. In fact, the problem of double counting is resulted
when detection score for each of the pair of symmetric parts becomes high at the same location because of occlusion. If not solved, this problem
may affect subsequent processing such as pose estimation~\cite{HyunSooParkICCV11} and action recognition~\cite{LambertoBallanIEEETM12, BangpengYaoECCV12, YuGangJiangIEEETM15}.

In this paper we model the human pose as a tree structure and develop a pose tracking algorithm where position of each body part is estimated based
on its appearance in the current frame, position of its ancestor in the current frame, and its own position in the previous frame. Note that our
model is different from the tree structure model proposed in~\cite{YiYangIEEETPAMI13}. We also propose a novel local part descriptor as the
appearance of body part.
Thus, our pose tracking algorithm is the combination of traditional object tracking~\cite{DorinComaniciuCVPR00} and pose estimation in still
image~\cite{MatthiasDantoneIEEETPAMI14}. Our main contributions in this paper are as follows.
\begin{itemize}
  \item We propose a new human part descriptor based on sum of intensities and gradients over annular region that is computed efficiently using integral image.
  \item We propose a new objective function for human pose tracking in video data.
  \item In addition, we introduce a new full body human pose tracking dataset called ICDPose,
  which is more challenging and bigger than many state-of-the-art datasets.
  The dataset is available at~\cite{ICDPose} for research purpose.
\end{itemize}

Rest of the paper is organized as follows. Related works are briefly described in Section~\ref{sec:Related_works}. We present our proposed method in
Section~\ref{sec:Proposed_method}
which includes human part description, and model formulation for pose tracking. We evaluate the performance of our pose tracking method in
Section~\ref{sec:Experimental_result}
which includes description of benchmark dataset, experimental setup, and comparison with the
state-of-the-art methods. Finally, Section~\ref{sec:Conclusion} concludes the paper.

\section{Related works}
\label{sec:Related_works}

Equation~(\ref{eqn:chap5:x_1_t_*_to_x_n_t_*}) suggests that the position of a part in the current frame depends on its own position in the previous frame.
This constraint helps in reducing the search space. If information from previous frame is not utilized, resultant algorithm can estimate human pose from a single
image~\cite{YiYangIEEETPAMI13, MatthiasDantoneIEEETPAMI14, AndreasMLehrmannICCV13, AlexanderToshevCVPR14, JensPuweinECCV14, VarunRamakrishnaECCV14, MartinKiefelECCV14}.
Almost all of these methods use pictorial structure model~\cite{PedroFFelzenszwalbIJCV05} explicitly for articulated human pose estimation, and differ primarily
from one another in determining appearance and modeling the interaction among different parts in terms of constraints and a priori. For example,
Yang et al.~\cite{YiYangIEEETPAMI13} capture the local part as a mixture of different parts, while Dantone et al.~\cite{MatthiasDantoneIEEETPAMI14} consider HOG
features~\cite{NavneetDalalCVPR05} and linear SVM as part appearance template. On the other hand, Kiefel and Gehler~\cite{MartinKiefelECCV14} model the presence
and absence of body parts at every possible location of the image at any orientation and any scale. This results in a large number of binary random variables,
which is handled more or less efficiently by approximate inference approach.
Though many approaches employ efficient optimization solver for pose estimation (e.g., branch and bound based algorithm by Puwein et al.~\cite{JensPuweinECCV14}),
Ramakrishna et al.~\cite{VarunRamakrishnaECCV14} have shown that modular framework along with symmetry property (left and right legs etc.) may lead to easy implementation
and efficient inference without any efficient optimization solver.
Dantone et al.~\cite{MatthiasDantoneIEEETPAMI14} find the human pose using pairwise interdependencies of the parts and co-occurrence based  joint regressors, while
Ramakrishna et al.~\cite{VarunRamakrishnaECCV14} exploit spatial interaction among multiple parts. Interdependencies of the parts are also handled by non-parametric
Bayesian network~\cite{AndreasMLehrmannICCV13}. In essence, all these methods adopt a common approach for human pose estimation, that is, by simultaneous identification
of body parts (joints) as well as their interdependencies. On the other hand, Toshev et al.~\cite{AlexanderToshevCVPR14} adopted a holistic approach for human pose
estimation using deep neural network. Deep convolutional neural network is also used for pose estimation~\cite{TomasPfisterACCV14} where temporal information from
multiple frames is exploited.
Human pose estimation from still image usually incur high computational cost (roughly $1$ second for an image~\cite{YiYangIEEETPAMI13}). Second, these methods do not
make use of temporal dependencies between locations of a part in subsequent frames. So, these methods are not directly employed to track human pose in video data.

In this paper our objective is to track the human pose in a video. To achieve this goal, our strategy is to track all the parts in the video subject to maintaining
the tree structure representing the human body. Object tracking in a video has a rich repertoires of algorithms.
Traditional object tracking~\cite{LauraSevilla-LaraCVPR12, WeiZhongIEEETIP14, Seung-HwanBaeIEEETIP14, AlexandreHeiliIEEETIP14,
 JunseokKwonIEEETPAMI13a, KaihuaZhangECCV14}
\cite{ShuWangIEEETIP14, ZhaoweiCaiIEEETIP14, YuweiWuIEEETIP15, ShunliZhangIEEETM15, KaihuaZhangIEEETPAMI14, EmilioMaggioWiley11, ArnoldWMSmeuldersIEEETPAMI14, ChaZhangICPR06, LexingXieIEEETM13, TengXuAVSS12, XinguoYuIEEETM06, JiTaoIJIG07, ChunTeChuIEEETM13, YangCongIEEETCSVT15, JingMingGuoIEEETCSVT12, YuanYuanIEEETM15}
algorithms generally search the target object in the current ($t^{th}$) frame within a search window around the target object
location in the previous $(t-1)^{th}$ frame. The target object is located by finding the maximum matching
score between the target object template obtained from $(t-1)^{th}$ frame and the patch at different locations within the search window in $t^{th}$ frame.
Variation in tracking strategy may lead to multiple object tracking~\cite{Seung-HwanBaeIEEETIP14} and nonrigid object tracking~\cite{JunseokKwonIEEETPAMI13a}.
Zhang et al.~\cite{KaihuaZhangECCV14} present a tracking method based on spatio-temporal context learning. They formulate the object tracking model
as spatio-temporal relationships between the object of interest and its locally dense contexts in a Bayesian framework. They have used Fast Fourier
Transform (FFT) to speedup the tracking process. Various other approaches such as Markov random field model~\cite{ AlexandreHeiliIEEETIP14}
and sparse formulation~\cite{WeiZhongIEEETIP14, MarkBarnardIEEETM14} are also adopted to track human figure.

Human pose tracking is significantly different from the traditional object tracking, as the former is a structurally combined representation of local parts. So,
here the tracking method should not only track the local parts, but also have to maintain the global structure in terms of connectivity.
Though an early work in this direction may be found in 1996~\cite{ShanonXJuICAFGR96}, because of complexity of the problem not many works on human pose tracking
from video data has been reported so far. Some works consider restricted view of the homan pose. For example, in \cite{DevaRamananCVPR05} the authors assume that the
people tend to take on a fixed set of canonical poses during activities and their algorithm can successfully detect the body parts in lateral walking pose.
However, for the said task we may borrow the concept of still image pose estimation and incorporate inter-frame dependencies to perform tracking.
Recently few researchers tried the same~\cite{VarunRamakrishnaCVPR13, AnoopCherianCVPR14}. Ramakroshna et al.~\cite{VarunRamakrishnaCVPR13} have modeled human
body as a combination of singleton parts (e.g., head, neck) and symmetric pair of parts (e.g, left and right feet). So they formulate the pose tracking problem as
a multi-target (parts) tracking problem where targets are related by an articulated structure. The appearance model and the optimization technique used to solve the
problem incurs high computational cost.
In~\cite{AnoopCherianCVPR14} the authors propose a pose estimation model for video data by incorporating optical flow information in the pictorial structure model
for still image~\cite{YiYangIEEETPAMI13}. As a result, computational complexity of the method becomes high.
Some works on human pose tracking based on 3D data~\cite{JonathanDeutscherIJCV05}
\cite{RaduHoraudIEEETPAMI08, JanBandouchIJCV12, LeonidSigalIJCV12, ChristianSchmaltzMVA12, ZhengZhangIEEETM13, TewodrosABiresawNeurocomputing15}
are also available in the literature.
We propose that we would estimate the human pose in the first frame of the video and then onward track the pose throughout the video. For the latter part of the task,
we may employ object tracking algorithm to each part locally maintaining spatial relationship between pair of parts guided by a tree-structure.

\section{Proposed method}
\label{sec:Proposed_method}

\begin{figure}
  \begin{center}
	\includegraphics[width = 0.65\textwidth]{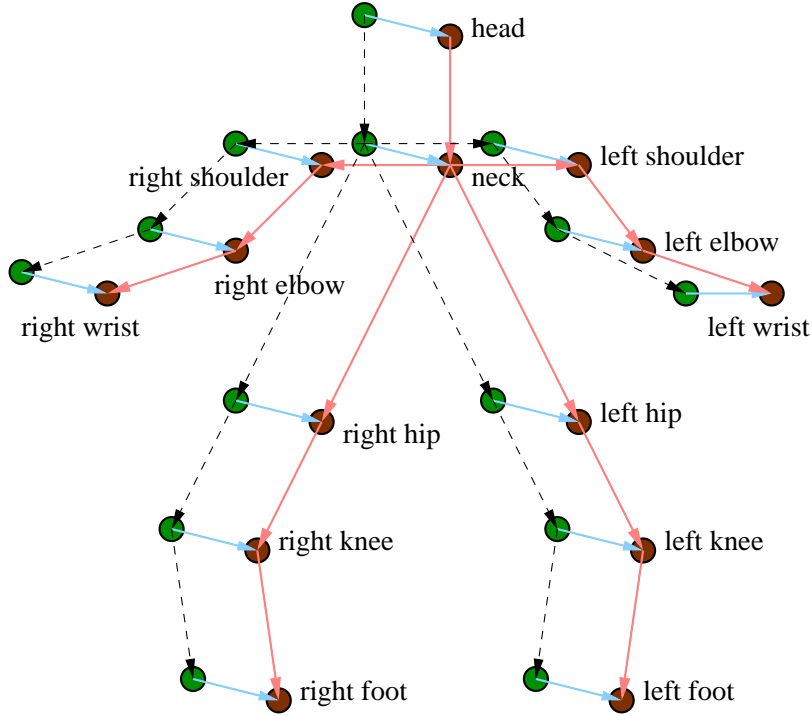}
    \caption{Proposed human pose tracking structure. Brown colored nodes represent the tree structure in the $t^{th}$ frame that we want to estimate, while the green colored nodes are the same in the $(t-1)^{th}$ frame, which are already known. The direct dependencies of a node are marked with arrow sign $(\rightarrow)$.}
    \label{fig:chap5:human_pose_tracking_structure}
  \end{center}
\end{figure}

We simplify the problem stated in  Eq.~(\ref{eqn:chap5:x_1_t_*_to_x_n_t_*}) with some rational assumptions and try to solve it within a reasonable time. We consider
a human body pose in an image or frame as a graphical tree structure model, where each part or, more specifically  `joint' corresponds to a node of the graph and
dependencies or physical link (based on human anatomy structure) between two parts define an edge of the graph (see brown colored structure in Fig.~\ref{fig:chap5:human_pose_tracking_structure}).
In this model we consider head as the root of the tree structure, because among all the body parts head is unique and mostly visible, and can be detected with highest
certainty~\cite{YiYangIEEETPAMI13}.

Our main motivation of approximating the solution of~(\ref{eqn:chap5:x_1_t_*_to_x_n_t_*}) for human pose
tracking is as follows. In human pose tracking, our ability to detect the position of each part in the current frame
depends on the three factors: (i) the position of its ancestor in the current frame, (ii) its own position in the previous frame, and
(iii) its appearance in the current frame. In Fig.~\ref{fig:chap5:human_pose_tracking_structure}, brown colored nodes represent the tree structure in the $t^{th}$ frame
that we want to estimate, while the green colored nodes are the same in the $(t-1)^{th}$ frame, which are already known. The direct dependencies
of a node are marked with arrow sign $(\rightarrow)$. So, if we know the position of the root of the human body tree structure in the current frame and
the nature of dependency of each node to its ancestors, then we can find the solution of~(\ref{eqn:chap5:x_1_t_*_to_x_n_t_*}) in polynomial time
using greedy approach.

Given that position of a part $p_{i}$ in $(t-1)^{th}$ frame is $\mathbf{x}_{i}^{(t-1)}$ and the position of its parent $p_{par(i)}$ in the $t^{th}$ frame is $\mathbf{x}_{par(i)}^{t}$, we find the position $\mathbf{x}_{i}^{t}$ of the part (node) $p_{i}$ in the $t^{th}$ frame as
\begin{equation}
    \mathbf{x}_{i}^{t*} = arg \min_{\mathbf{x}_{i}^{t}} \{ l_{i}(\mathbf{x}_{i}^{t}) + d_{i}(\mathbf{x}_{i}^{(t-1)}, \mathbf{x}_{i}^{t}) + d_{i, par(i)}(\mathbf{x}_{i}^{t}, \mathbf{x}_{par(i)}^{t})\}
    \label{eqn:chap5:x_i_t_*}
\end{equation}
where $l_{i}(\mathbf{x}_{i}^{t})$ measures the likeliness of appearance when the template of part $p_{i}$ is placed at location
$\mathbf{x}_{i}^{t}$ in the $t^{th}$ frame. Note that the image or feature representation of part $p_{i}$ in $(t-1)^{th}$ frame is used as the template for that
part in the $t^{th}$ frame. The function $d_{i}(\mathbf{x}_{i}^{(t-1)}, \mathbf{x}_{i}^{t})$ represents the amount of temporal displacement
of part $p_{i}$ from $(t-1)^{th}$ frame to $t^{th}$ frame. For a part $p_{i}$ given its parent
$p_{par(i)}$, $d_{i, par(i)}(\mathbf{x}_{i}^{t}, \mathbf{x}_{par(i)}^{t})$ is a function which measures the deviation from expected spatial
distance between part $p_{i}$ and its parent in the $t^{th}$ frame.

To find $\mathbf{x}_{i}^{t*}$ we need to know $\mathbf{x}_{par(i)}^{t*}$ first. Similarly to know $\mathbf{x}_{par(i)}^{t*}$ we
have to know $\mathbf{x}_{par(par(i))}^{t*}$ by solving the optimization problem~(\ref{eqn:chap5:x_i_t_*}) at appropriate level. In this way we recursively
reach the root node of the pose tree structure. So we find the position of the root part $p_{root}$ in $t^{th}$ frame ignoring
the term defining dependency to parent node in~(\ref{eqn:chap5:x_i_t_*}) as follows:
\begin{equation}
    \mathbf{x}_{root}^{t*} = arg \min_{\mathbf{x}_{root}^{t}} \{ l_{root}(\mathbf{x}_{root}^{t}) + d_{root}(\mathbf{x}_{root}^{(t-1)}, \mathbf{x}_{root}^{t})\}
    \label{eqn:chap5:x_root_t_*}
\end{equation}
where the function $l_{root}(\mathbf{x}_{root}^{t})$ measures the likeliness of appearance when the template of root part $p_{root}$ is
placed at location $\mathbf{x}_{root}^{t}$, and $d_{root}(\mathbf{x}_{root}^{(t-1)}, \mathbf{x}_{root}^{t})$ measures the amount
of temporal displacement as stated before.

We consider the objective functions~(\ref{eqn:chap5:x_i_t_*}) and~(\ref{eqn:chap5:x_root_t_*}) for pose tracking at possible
position $\mathbf{x}_{i}^{t} \in W_{i}$, where $W_{i}$ denotes a window around $\mathbf{x}_{i}^{(t-1)}$. The functions
$d_{i}(\mathbf{x}_{i}^{(t-1)}, \mathbf{x}_{i}^{t})$ and $d_{i, par(i)}(\mathbf{x}_{i}^{t}, \mathbf{x}_{par(i)}^{t})$ actually play
the role of constraints in estimating the part position $\mathbf{x}_{i}^{t}$, because in~(\ref{eqn:chap5:x_i_t_*}) $l_{i}(\mathbf{x}_{i}^{t})$ measures
the likeliness of appearance of part $p_{i}$ between $t^{th}$ and $(t-1)^{th}$ frames and we try to optimize it. Now depending on the speed of
the movement and the rate of change in appearance of the part, reliability of each of the above terms varies. So we rewrite our objective function for each part $p_{i}$ in a regularization form as,
\begin{equation}
\small
    \mathbf{x}_{i}^{t*} = arg \min_{\mathbf{x}_{i}^{t}} \{ l_{i}(\mathbf{x}_{i}^{t}) + \lambda_{1}d_{i}(\mathbf{x}_{i}^{(t-1)}, \mathbf{x}_{i}^{t}) + \lambda_{2}d_{i, par(i)}(\mathbf{x}_{i}^{t}, \mathbf{x}_{par(i)}^{t})\}
    \label{eqn:chap5:regularized_x_i_t_*}
\end{equation}
and
\begin{equation}
    \mathbf{x}_{root}^{t*} = arg \min_{\mathbf{x}_{root}^{t}} \{ l_{i}(\mathbf{x}_{root}^{t}) + \lambda_{1}d_{root}(\mathbf{x}_{root}^{(t-1)}, \mathbf{x}_{root}^{t})
    \label{eqn:chap5:regularized_x_root_t_*}
\end{equation}
where $\lambda_{1}$ and $\lambda_{2}$ are the regularization parameters controlling the importance of various terms in optimization.

To minimize the objective functions~(\ref{eqn:chap5:regularized_x_i_t_*}) and~(\ref{eqn:chap5:regularized_x_root_t_*}), we need to know the functions
$l_{i}(\mathbf{x}_{i}^{t})$, $d_{i}(\mathbf{x}_{i}^{(t-1)}, \mathbf{x}_{i}^{t})$, for $i = 1:n$ and $d_{i, par(i)}(\mathbf{x}_{i}^{t}, \mathbf{x}_{par(i)}^{t})$ for $i = 2:n$.
We learn these functions from the training data and describe this learning process in the subsequent subsections.

\subsection{Measure of likeliness of appearance}
\label{subsec:degree_of_appearance_mismatch}

Measure of likeliness of appearance $l_{i}(\mathbf{x}_{i}^{t})$ for each part $p_{i}$ ($i = 1:n$) in $t^{th}$ frame is an important term for object detection and
tracking. In human pose estimation this term is learned from the training data, where HOG features are widely used~\cite{YiYangIEEETPAMI13}. In human pose tracking,
use of fixed template may not work well because of movement, 3D to 2D projection and
occlusion. So people try to match the raw pixel values of the part between $(t-1)^{th}$ and $t^{th}$ frames using, say, sum of absolute differences (SAD)~\cite{Lai-ManPoTENCON96}.
Here we measure the likeliness of appearance using Euclidean distance between feature vectors $\phi(\mathbf{x}_{i}^{t})$ describing the appearance of the part $p_{i}$
at location $\mathbf{x}_{i}^{(t)}$ in $t^{th}$ frame and the corresponding template $\tau(\mathbf{x}_{i}^{t})$ as
\begin{equation}
    l_{i}(\mathbf{x}_{i}^{t}) = \parallel \phi(\mathbf{x}_{i}^{t}) - \tau(\mathbf{x}_{i}^{(t)})\parallel_{2}
\end{equation}
Traditional methods form $\phi(\mathbf{x}_{i}^{t})$ with raw pixel values and use $\tau(\mathbf{x}_{i}^{(t)}) = \phi(\mathbf{x}_{i}^{(t-1)})$.
Here we describe each human part using a novel rectangular feature using \textit{Integral image}. Thus the proposed feature can be computed more efficiently compared
to state-of-the-art features (see computational complexity in Subsection~\ref{subsec:Computational_complexity}). As our feature computation is based on Integral
image representation, we briefly describe it next.

\begin{figure}
  \begin{center}
    $\begin{array}{@{\hspace{3pt}}c@{\hspace{3pt}}c@{\hspace{3pt}}c}
	\includegraphics[width=0.24\textwidth]{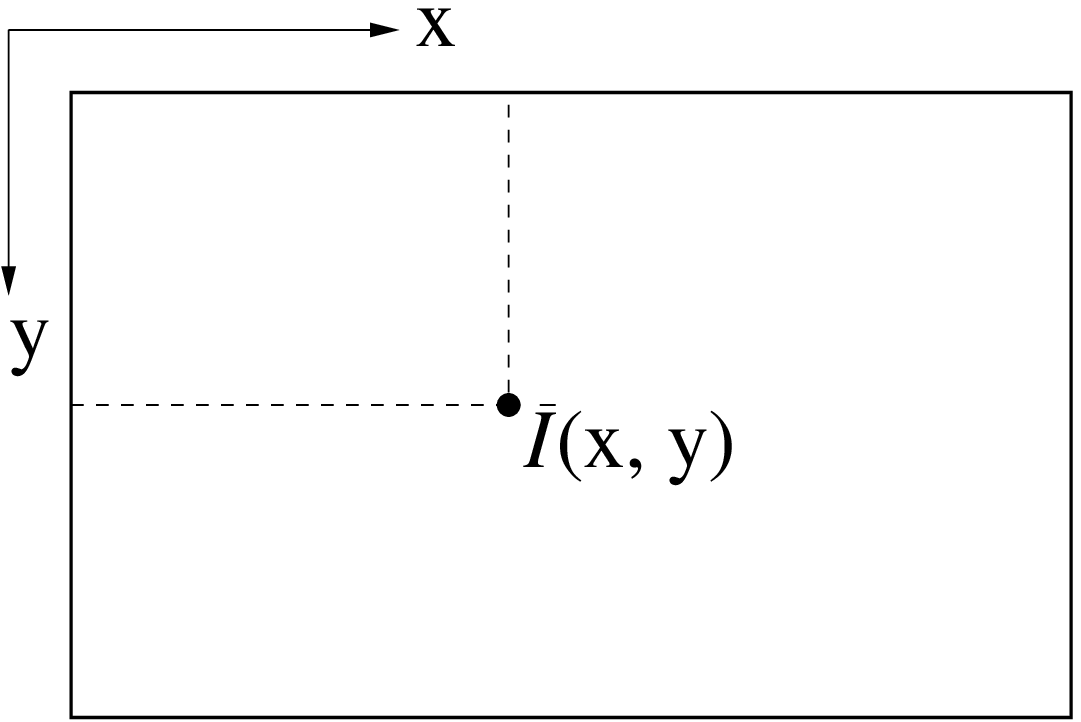} &
	\includegraphics[width=0.24\textwidth]{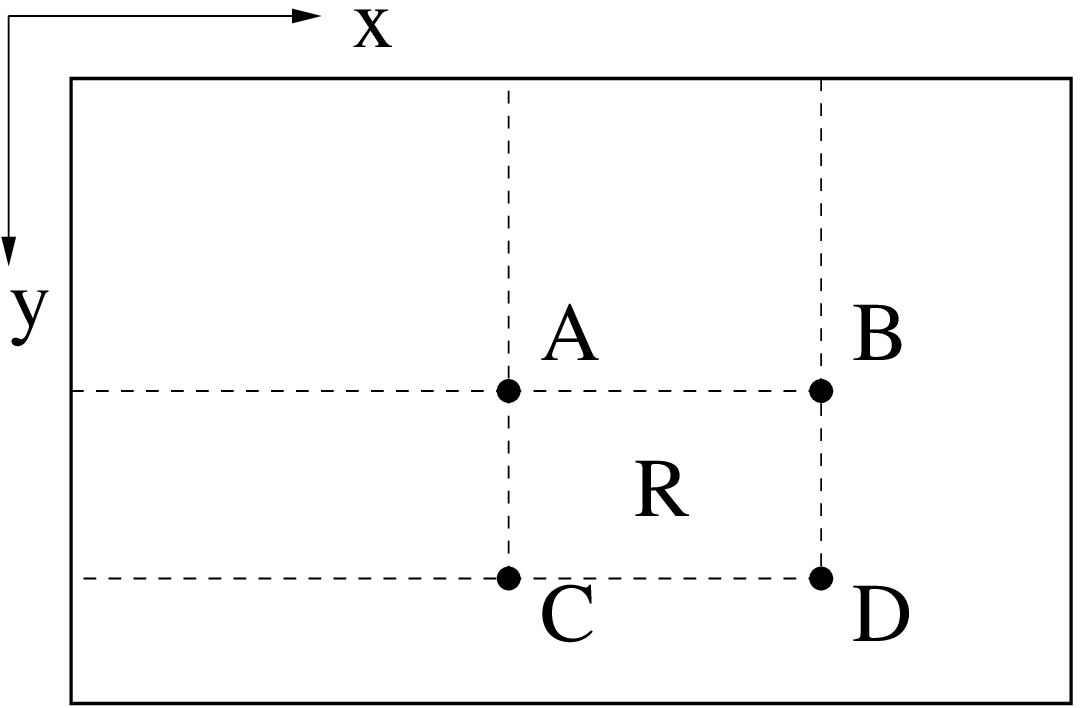} &
	\includegraphics[width=0.24\textwidth]{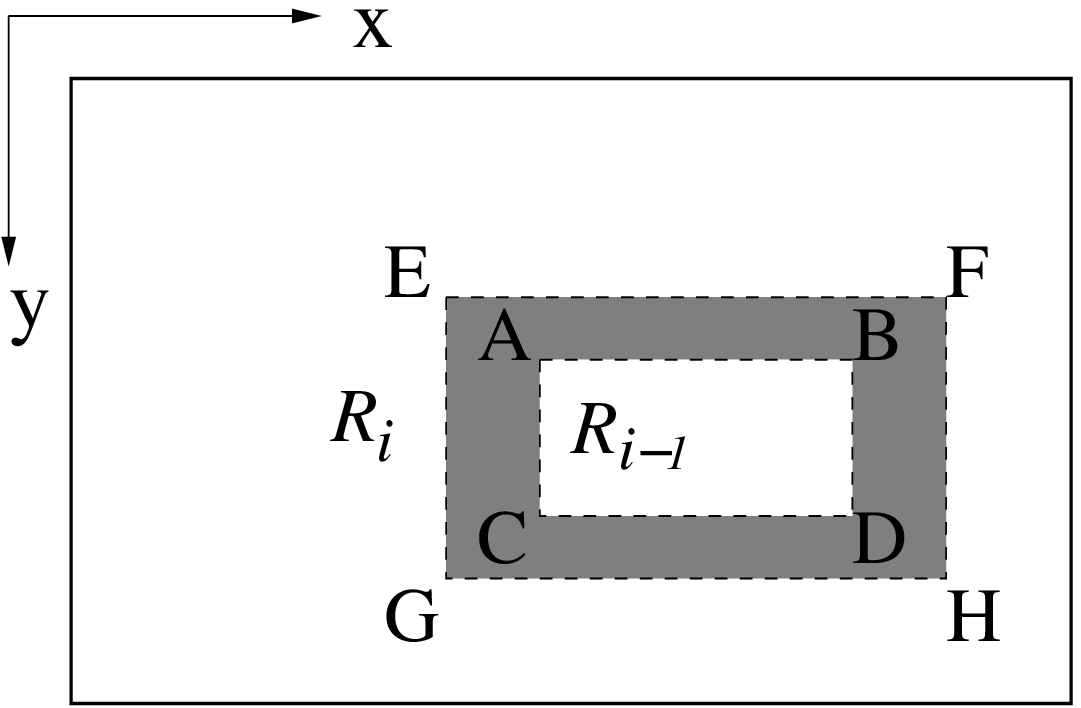}\\
	(a) & (b)	&	(c)
    \end{array} $
    \caption{Integral image representation and rectangular region feature calculation: (a) integral image, (b) sum of the pixels value within the
    region $R$ can be computed using [(A+D) - (B+C)], and (c) sum of pixel values within the region $R_{i}\setminus R_{i-1}$ (shaded region) can be
    computed using [\{(E+H) - (F+G)\} - \{(A+D) - (B+C)\}].}
    \label{fig:chap5:integral_image_representation}
  \end{center}
\end{figure}

{\bf Integral image representation:} Integral image was first appeared in graphics literature~\cite{FranklinCCrowACMSIGGRAPH84}
and became popular in computer vision community after successful application in face detection~\cite{PaulViolaCVPR01}. Let $I$ be an
input image. Then integral image $\bar{I}$ can be defined as
\begin{equation}
    \bar{I}(x, y) = \sum_{1 \leq r\leq x; 1 \leq c\leq y}I(r, c)			
    \label{equ:chap5:I_bar_x_y}
\end{equation}
i.e., $\bar{I}(x, y)$ stores the sum of all pixels above and left to the pixel $(x, y)$ of input image $I$ [Fig.~\ref{fig:chap5:integral_image_representation}(a)].
Integral image $\bar{I}$ can be computed in a single pass over the input image $I$  using the following recurrence relations.
\begin{equation}
    S(x, y) = S(x, y-1) + I(x, y)			
    \label{equ:chap5:I_bar_recurrence_x_y}
\end{equation}
\begin{equation}
    \bar{I}(x, y) = \bar{I}(x-1, y) + S(x, y)			
    \label{equ:chap5:I_bar_recurrence_x_y}
\end{equation}
where $S$ is the cumulative row sum of image $I$ with $S(x, 0) = 0$ and $\bar{I}(0, y) = 0$.
The advantage of $\bar{I}$ is that the sum of pixel values in any rectangular region ABCD [Fig.~\ref{fig:chap5:integral_image_representation}(b)] of $I$ can be
computed as $\bar{I}(D) + \bar{I}(A) - \bar{I}(B) - \bar{I}(C)$, i.e., using only three arithmetic operations, which is quite fast.

\begin{figure}[!ht]
  \begin{center}
    $\begin{array}{@{\hspace{3pt}}c@{\hspace{4pt}}c}
	\includegraphics[width=0.30\textwidth]{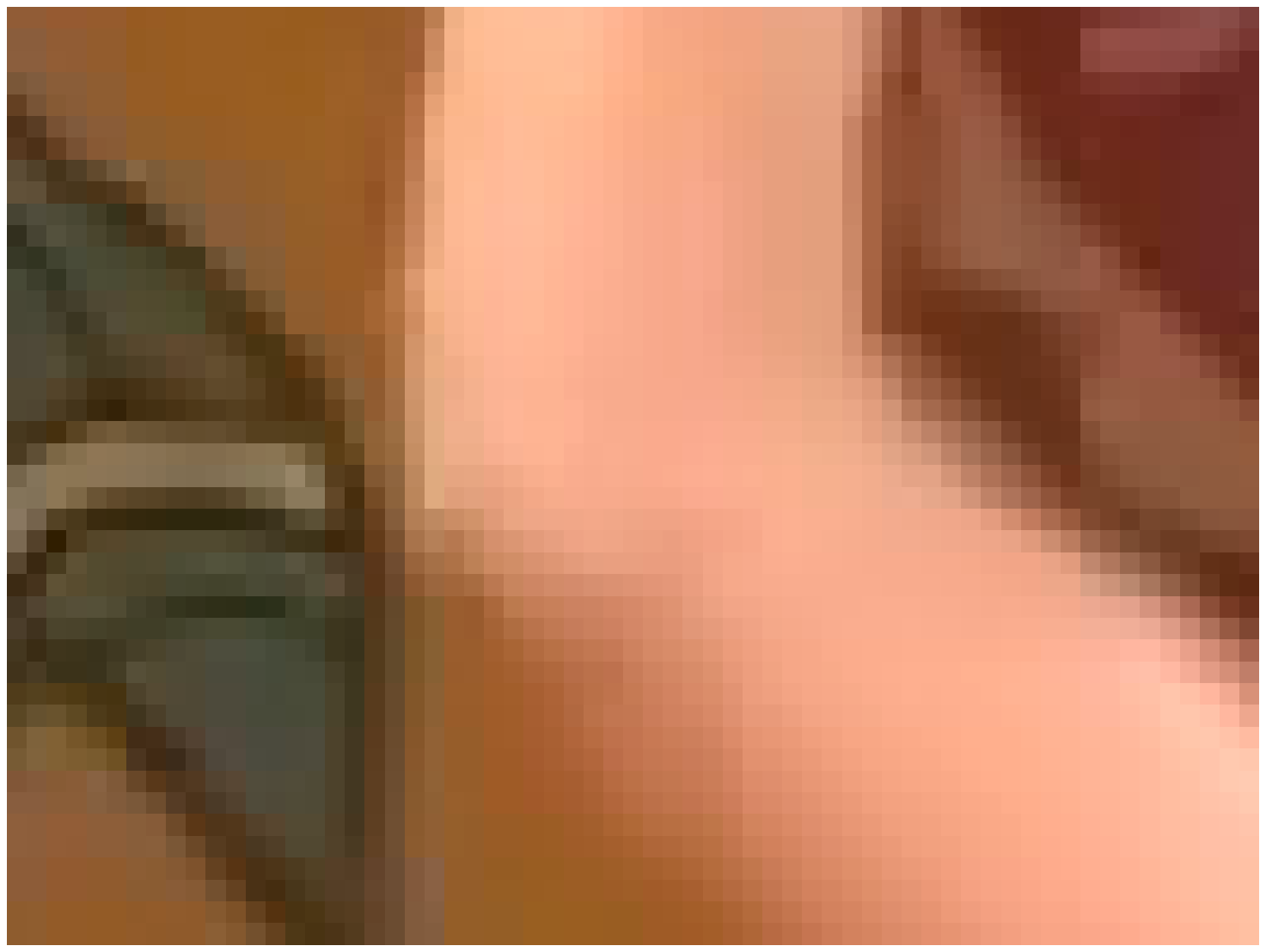} &
	\includegraphics[width=0.30\textwidth]{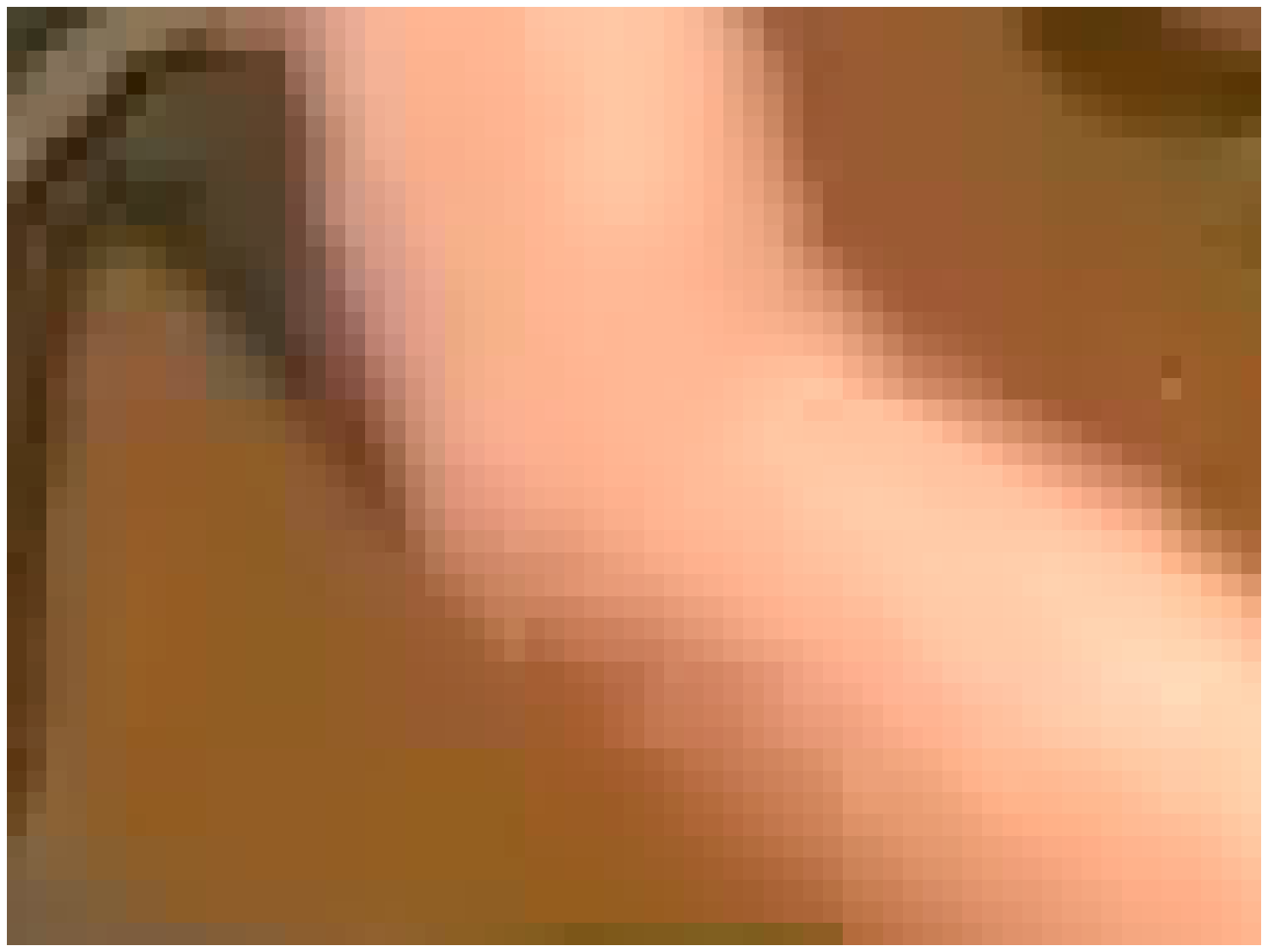} \\
	(a) & (b)
    \end{array} $
    \caption{Elbow part in two consecutive frames of ICDPose dataset.}
    \label{fig:chap5:icdpose_elbow_part}
  \end{center}
\end{figure}

{\bf Proposed local feature computation: } For human body part description the main challenge is the deformation in appearance as well as its rotation. For example,
consider the \textit{elbow} part in action where joint angle changes frequently (see Fig.~\ref{fig:chap5:icdpose_elbow_part}), and due to that
the traditional features (e.g, raw pixel values or different variants of gradient histogram based features ) may not work well. So to represent a human body part we
consider $m$ concentric rectangular annular strip around the center of that part as shown in Fig.~\ref{fig:chap5:integral_image_representation}(c) and mark them
as $R_{1}$, . . ., $R_{m}$ in ascending order of their areas.
Now sum of pixel values within the annular region $a_i = R_{i}\setminus R_{i-1}$ $(i = 1:m,$ with $R_{0} = \Phi)$ can be computed very efficiently using integral
image representation. We calculate the sum of pixel values within the region $a_i$ by subtracting sum over $R_{i-1}$ from the sum over $R_{i}$.

For each color channel we compute $m$ sum values for intensity, magnitude of horizontal gradient and magnitude of vertical gradient separately. We normalize these
sum values by their corresponding area. Thus we get $9m$ dimensional feature vector at each location of human body part.

Occlusion of parts causes a big problem in human pose tracking. To overcome this issue we update the body part template based on the previous frames. This
learned template helps in describing the modified part more reliably under occlusion as well as deformation. After estimating $x_{i}^{t}$ in $t^{th}$ frame, we
update the template $\tau(\mathbf{x}_{i}^{t+1})$ for $p_{i}$ ($i = 1:n$) in the $(t+1)^{th}$ frame as
\begin{equation}
   \tau(\mathbf{x}_{i}^{t+1}) = \alpha \phi(\mathbf{x}_{i}^{t}) + (1 - \alpha)\tau(\mathbf{x}_{i}^{(t)})			
    \label{equ:chap5:phi_feature_update}
\end{equation}
where $\alpha = e^{-l_{i}(\mathbf{x}_{i}^{t})}$. Note that the proposed feature vector as well as the template as rotation, translation and flip invariant.

\subsection{Temporal displacement}
\label{subsec:degree_of_temporal_displacement}

\begin{figure}
  \begin{center}
	\includegraphics[width = 0.69\textwidth]{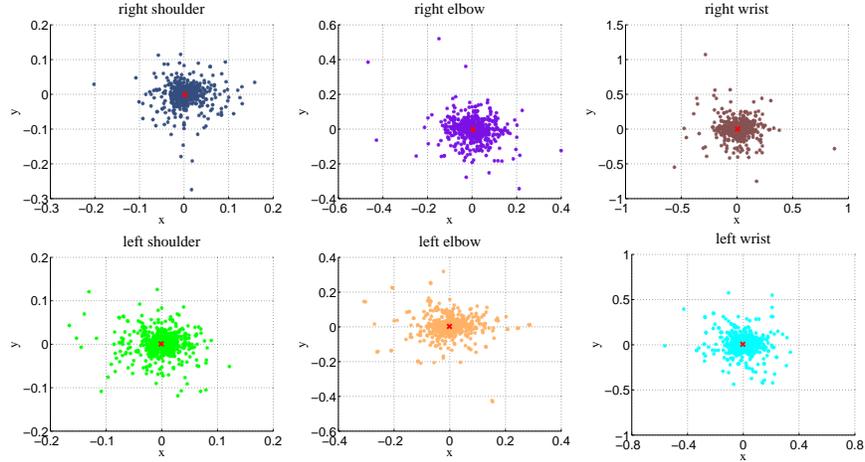}
    \caption{Temporal displacement of different parts of VideoPose2 training database.}
    \label{fig:chap5:videopose_upper_body_part_teporal_displacement}
  \end{center}
\end{figure}

\begin{figure}
  \begin{center}
	\includegraphics[width = 0.69\textwidth]{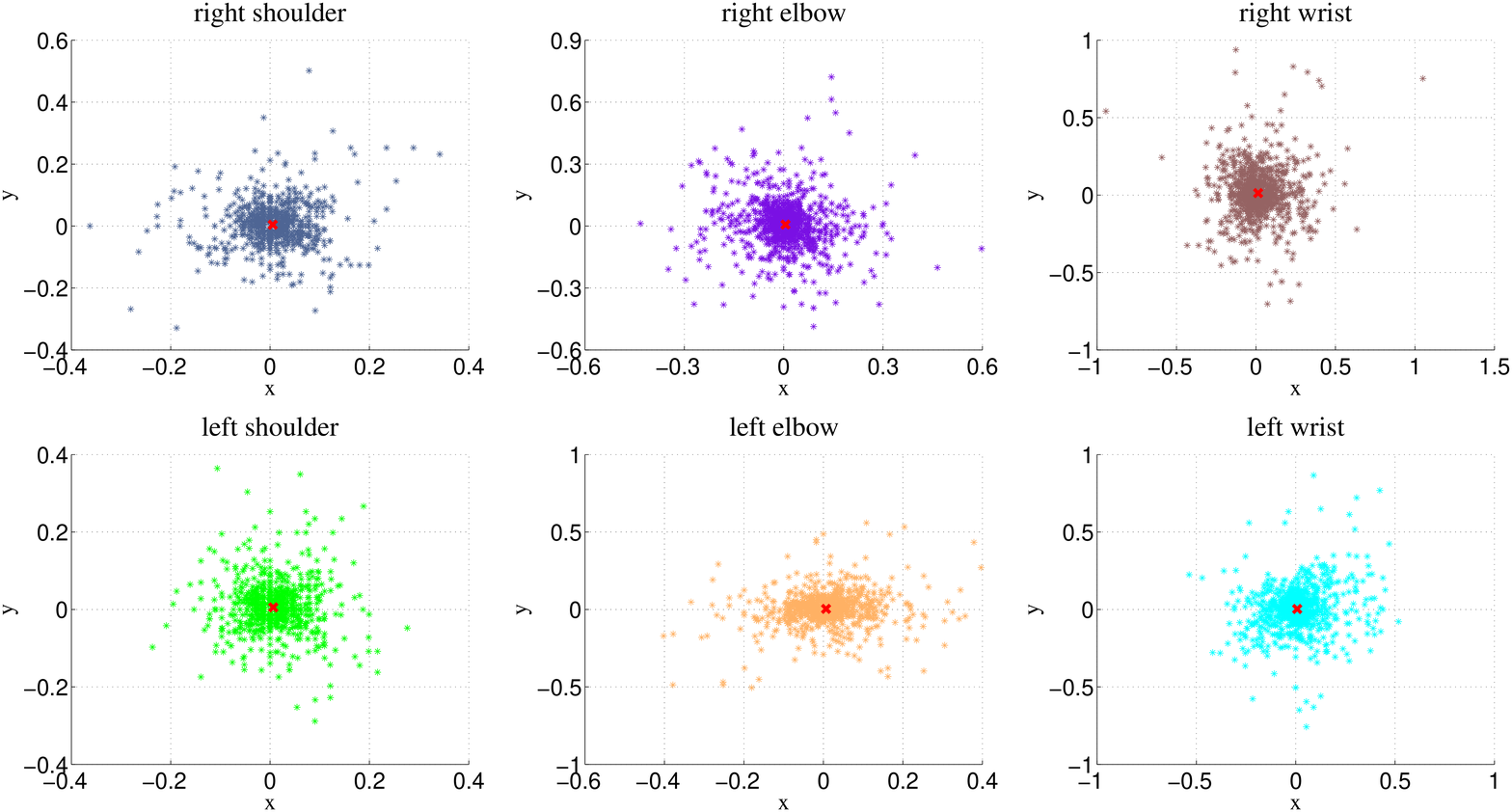}
    \caption{Temporal displacement of different upper body parts of ICDPose training dataset.}
    \label{fig:chap5:icdpose_upper_body_part_teporal_displacement}
  \end{center}
\end{figure}

\begin{figure}
  \begin{center}
	\includegraphics[width = 0.69\textwidth]{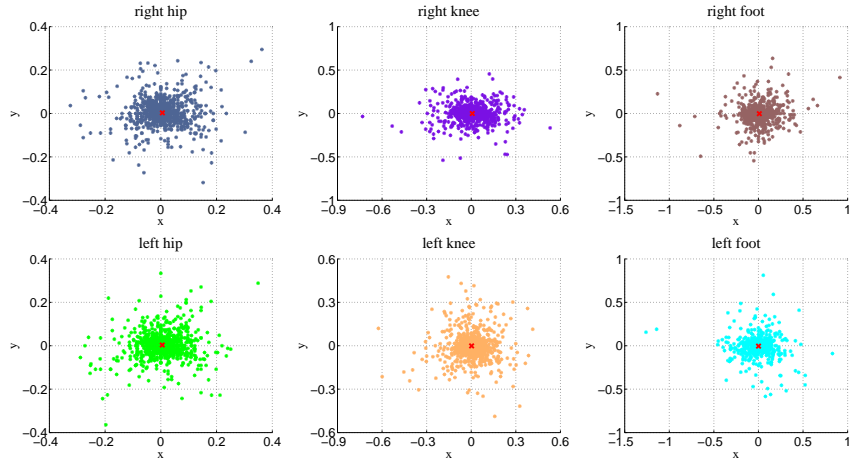}
    \caption{Temporal displacement of different lower body parts of ICDPose training dataset.}
    \label{fig:chap5:icdpose_lower_body_part_teporal_displacement}
  \end{center}
\end{figure}

We have defined the amount of temporal displacement $d_{i}(\mathbf{x}_{i}^{(t-1)}, \mathbf{x}_{i}^{t})$ of each part $p_{i}$ ($i = 1:n$) in the data driven framework.
From the training data we calculate temporal displacement $\mathbf{e}_{i} = \mathbf{x}_{i}^{t} - \mathbf{x}_{i}^{(t-1)}$ from $(t-1)^{th}$ frame to $t^{th}$ frame. Thus, for each part
$p_{i}$ we have a set of temporal displacement $\mathbf{e}_{i}$. Fig.~\ref{fig:chap5:videopose_upper_body_part_teporal_displacement} show the temporal
displacement of different parts (left and right shoulders, left and right elbows, and left and right wrists) from VideoPose2 training dataset. Temporal displacement of
ICDPose training data for different parts are shown in Fig.~\ref{fig:chap5:icdpose_upper_body_part_teporal_displacement}
and~\ref{fig:chap5:icdpose_lower_body_part_teporal_displacement}. Observing these three figures we postulate that a bivariate Gaussian distribution may represent the
temporal displacement of each part. So, from the set of $\mathbf{e}_{i}$'s, we learn a Gaussian distribution ($\mu_{i}$, $\Sigma_{i}$) for each part $p_{i}$ ($i = 1:n$). We use
Mahalanobis distance from the learned distribution for $\mathbf{e}_{i}$'s to define $d_{i}(\mathbf{x}_{i}^{(t-1)}, \mathbf{x}_{i}^{t})$ of part $p_{i}$ in $t^{th}$ frame as,
\begin{equation}
    d_{i}(\mathbf{x}_{i}^{(t-1)}, \mathbf{x}_{i}^{t}) = (\mathbf{e}_{i} - \mathbf{\mu}_{i})^{T}\mathbf{\Sigma}_{i}^{-1}(\mathbf{e}_{i} - \mathbf{\mu}_{i})			
    \label{equ:chap5:d_i_t_t-1}
\end{equation}
where $\mathbf{e}_{i} = \mathbf{x}_{i}^{t} - \mathbf{x}_{i}^{(t-1)}$ is the temporal displacement of part $p_{i}$ from $(t-1)^{th}$ frame to $t^{th}$ frame.

\subsection{Spatial deformation}
\label{subsec:degree_of_spatial_deformation}

\begin{figure}
  \begin{center}
	\includegraphics[width = 0.69\textwidth]{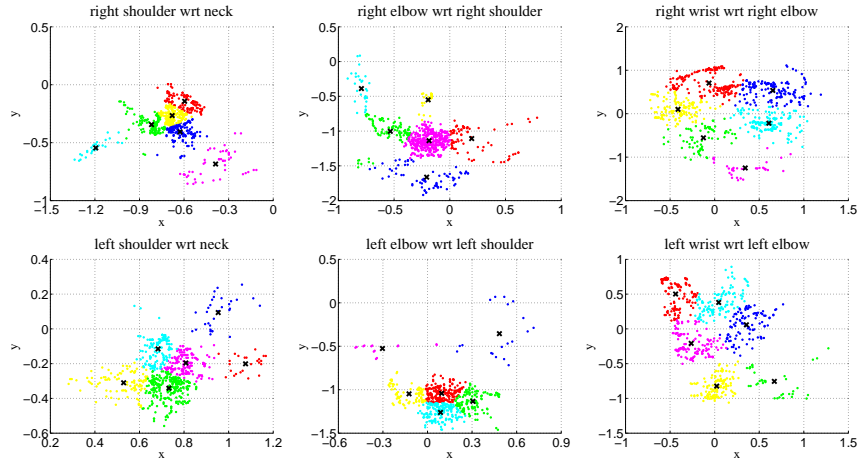}
    \caption{Different body part positions with respect to (wrt) its parent position of Videopose2 training database.}
    \label{fig:chap5:videopose_upper_body_part_position}
  \end{center}
\end{figure}
\begin{figure}
  \begin{center}
	\includegraphics[width = 0.69\textwidth]{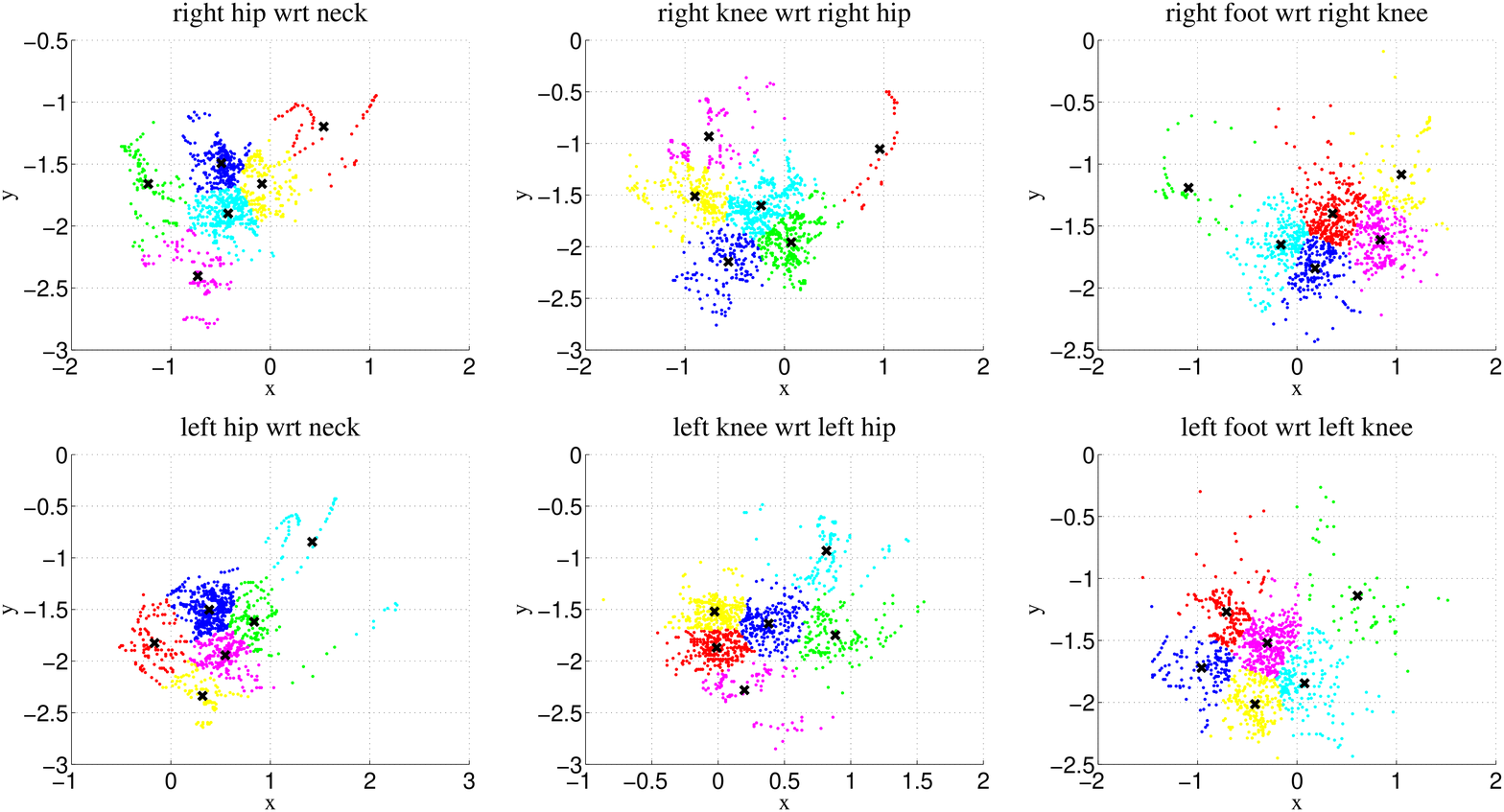}
    \caption{Different upper body part positions with respect to (wrt) its parent position of ICDPose training database.}
    \label{fig:chap5:icdpose_upper_body_part_position}
  \end{center}
\end{figure}
\begin{figure}
  \begin{center}
	\includegraphics[width = 0.69\textwidth]{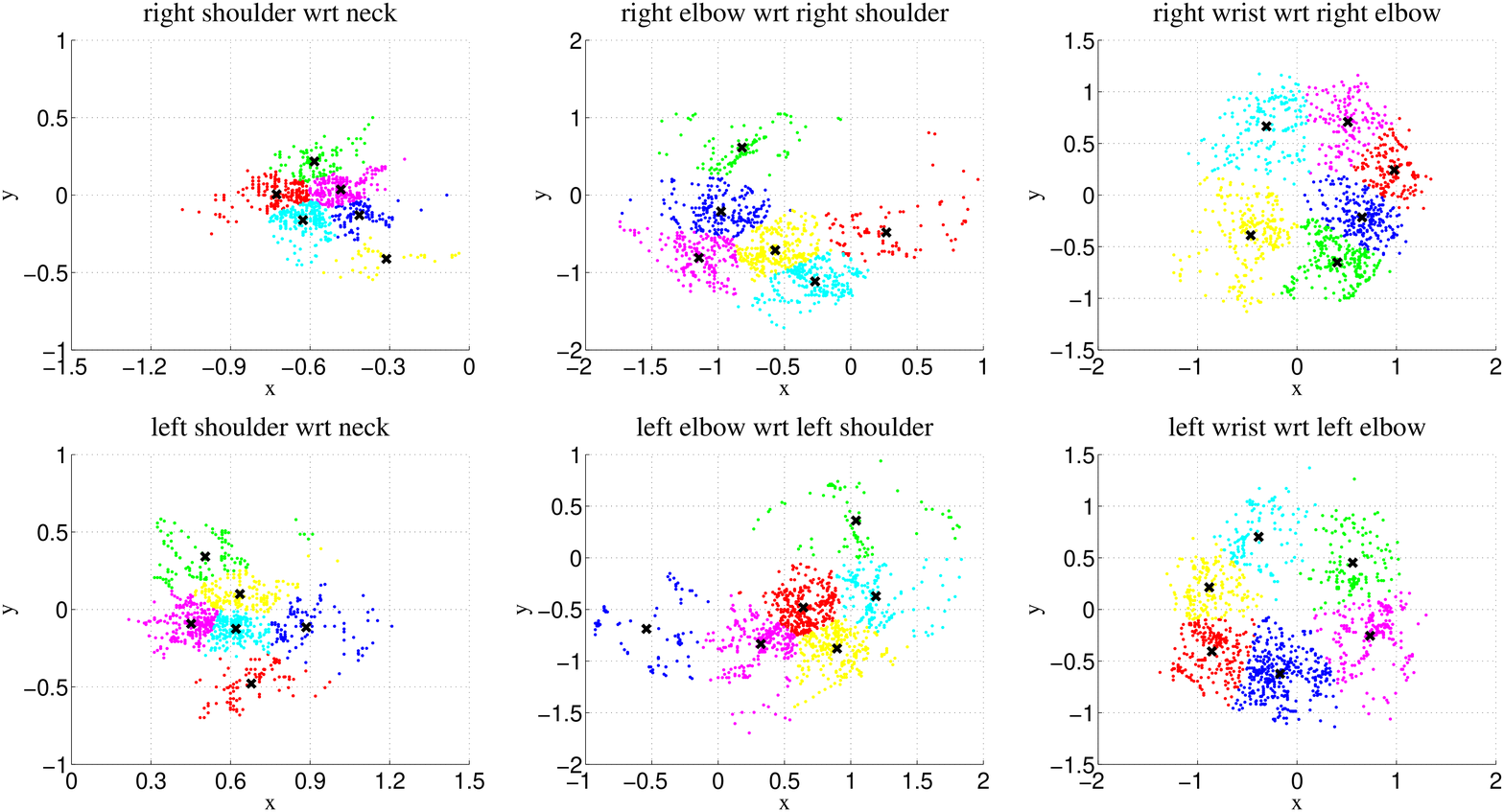}
    \caption{Different lower body part position with respect to (wrt) its parent position of ICDPose training database.}
    \label{fig:chap5:icdpose_lower_body_part_position}
  \end{center}
\end{figure}

Distance between two parts may change in 2D frame due to change in orientation of the portion of the body connecting two said parts in 3D. We call this change in distance
as a result of 3D to 2D mapping as \textit{spatial deformation}. This spatial deformation is handled through the dependency of a body part node of the pose tree
structure on its parent part. In traditional pose estimation or object recognition model, researchers have captured this dependency by relative position of that
part $p_{i}$ with respect to its parent connected by an edge. In~\cite{PedroFFelzenszwalbIJCV05} the degree of deformation of a part with respect to the other is
modeled by a Gaussian distribution of their relative position. We use similar idea with a little modification to model our part dependency relation in data driven
framework. We compute the relative position of part $p_{i}$ with respect to its parent $p_{par(i)}$ position from the training data.
Fig.~\ref{fig:chap5:videopose_upper_body_part_position} and~\ref{fig:chap5:icdpose_upper_body_part_position}  shows the relative positions of each parts (left and
right shoulder with respect to neck; left and right elbow to the left and right shoulder respectively; and left and right wrist to the left and right elbow
respectively) for VideoPose2 and ICDPose dataset respectively. Fig.~\ref{fig:chap5:icdpose_lower_body_part_position} shows the relative positions of lower body
parts (left and right hips and relatively lower) of ICDPose dataset. From these three figures we observe that one Gaussian distribution for each part is not sufficient
to capture its dependency relation. Instead we make the system learn multiple Gaussian distributions for each part to capture the part's spatial dependency.

To learn multiple Gaussian distributions, we use the data driven approach~\cite{YiYangIEEETPAMI13}. We first cluster the relative positions of each part using k-means
clustering algorithm. Figs.~\ref{fig:chap5:videopose_upper_body_part_position},~\ref{fig:chap5:icdpose_upper_body_part_position},
and~\ref{fig:chap5:icdpose_lower_body_part_position} show the clusters of relative locations of different parts of VideoPose2 and ICDPose datasets. Then we learn a
Gaussian distribution ($\mathbf{\mu}_{i, par(i)}^{c}$, $\mathbf{\Sigma}_{i, par(i)}^{c}$; $c = 1:N_{i}$) ($i = 2:n$) for each cluster separately, where $N_{i}$ is
the number of clusters of relative position for part $p_{i}$. Similar to temporal displacement, here also we use Mahalanobis distance measure to define the degree
of deformation $d_{i, par(i)}(\mathbf{x}_{i}^{t}, \mathbf{x}_{par(i)}^{t})$ of a part $p_{i}$ ($i = 2:n$) with respect to its parent $p_{par(i)}$ as,
\small
\begin{eqnarray}
    d_{i, par(i)}(\mathbf{x}_{i}^{t}, \mathbf{x}_{par(i)}^{t}) &=& \min_{\substack{\mathbf{\mu}_{i,par(i)}^{c}, \mathbf{\Sigma}_{i,par(i)}^{c} \\ c = 1:N_{i}}} \{(\mathbf{e}_{i,par(i)} - \mathbf{\mu}_{i,par(i)}^{c})^{T} \nonumber\\
    & &(\mathbf{\Sigma}_{i,par(i)}^{c})^{-1}(\mathbf{e}_{i,par(i)} - \mathbf{\mu}_{i,par(i)}^{c})\}		
    \label{equ:chap5:d_i_par_i_t_t}
\end{eqnarray}
\normalsize
where $\mathbf{e}_{i,par(i)} = \mathbf{x}_{i}^{t} - \mathbf{x}_{par(i)}^{t}$ is the relative displacement of part $p_{i}$ with respect to its parent $p_{par(i)}$.

\subsection{Tracking human pose in a video}
\label{subsec:tracking_human_pose_in_a_video}

After getting all the parameters for the functions $l_{i}(\mathbf{x}_{i}^{t})$, $d_{i}(\mathbf{x}_{i}^{(t-1)}, \mathbf{x}_{i}^{t})$, ($i = 1:n$)
and $d_{i, par(i)}(\mathbf{x}_{i}^{t}, \mathbf{x}_{par(i)}^{t})$ ($i = 2:n$) we plugin the optimization problems~(\ref{eqn:chap5:regularized_x_i_t_*})
and~(\ref{eqn:chap5:regularized_x_root_t_*}) for the human pose tracking in a video. Now for tracking a human pose in a video, we need the
human pose at the first frame of that video. We may manually annotate the human pose at the first frame or may employ a good human pose
estimation algorithm for still image, and then track that pose through all the frames of that video using our proposed pose tracking model.
Thus our human pose tracking method for a video clip works as follows: We manually annotate a human pose in the first frame of a video and
our aim is to track that pose through all the frames of that video. As we have mentioned that human head is the root part of our pose structure.
So, for the second frame we first track the head part using~(\ref{eqn:chap5:regularized_x_root_t_*}) and then track all the remaining parts
using~(\ref{eqn:chap5:regularized_x_i_t_*}) in a greedy fashion. In a similar way, given a human pose in the $k^{th}$ frame we track the
pose in the $(k+1)^{th}$ frame of that video. Our algorithm first fixes the root node and travels from parent $p_{par(i)}$ position
$\mathbf{x}_{par(i)}^{t*}$  of a part $p_{i}$ to position $\mathbf{x}_{i}^{t*}$ of part $p_{i}$. So, computational complexity to track each
part is linear in the possible location of each part $p_{i}$ with a constant multiplier (number of clusters of relative location of part
$p_{i}$). Let for each part $p_{i}$ ($i = 1:n$) we have $M$ possible locations and $N$ number of clusters for each of these relative locations.
Then the time complexity of our proposed method is $O(nMN)$ per frame. We evaluate our proposed human pose tracking method using standard
benchmark datasets and compare with the state-of-the-art methods discuss in the next Section~\ref{sec:Experimental_result}.

\section{Experimental result}
\label{sec:Experimental_result}

We have implemented our algorithm in MATLAB2013a and evaluated in a system with Intel(R) Core(TM) i5-2430M CPU $@$ 2.40 GHz and 4GB RAM running Windows 7 operating system. We evaluate our
proposed method on benchmark datasets as well as on our new dataset. In this section we briefly describe each of the datasets followed by experimental settings.

\subsection{Datasets}
\label{subsec:Datasets}

Here we have used three benchmark datasets: VideoPose2~\cite{BenjaminSappCVPR11}, Poses in the Wild~\cite{AnoopCherianCVPR14} and Outdoor
Pose~\cite{VarunRamakrishnaCVPR13}, and our new dataset ICDPose~\cite{ICDPose}.

{\bf VideoPose2 dataset\footnote{http://vision.grasp.upenn.edu/cgi-bin/index.php?n=VideoLearning.VideoPose2}: } This dataset is
created from the TV shows \textit{Friends} and \textit{Lost}. The dataset contains $44$ video clips with a total of $1286$ frames. The dataset focuses on only
the upper portion of body. Body parts such as torso, shoulders, elbows and
wrists are manually annotated in  all the frames. The authors have indicated the data partition for training (26 video clips) and test (14 video clips). We have followed this partition in our experiment.

{\bf Poses in the Wild dataset\footnote{https://lear.inrialpes.fr/research/posesinthewild/}: } This dataset consists of $30$ video clips
with a total of 830 frames. The authors have created this dataset from Hollywood movies \textit{Forrest Gump}, \textit{The Terminal}, and \textit{Cast Away}.
This dataset too focuses on the upper portion of body with manually annotated parts: neck, shoulders, elbows, wrists and mid-torso.

{\bf ICDPose dataset\footnote{http://www.isical.ac.in/$\sim$vlrg/sites/default/files/Soumitra/Site/ss\_icdpose.html}: } Our Indian
classical dance pose (ICDPose) dataset contains full body pose data. It has $60$ video clips covering six most popular Indian classical dance styles
(Bharatnatyam, Kathak, Kuchupudi, Mohiniyattam, Manipuri and Odissi). Each video clip has 45 frames with a total of 2700 frames in the whole dataset, which
is sufficiently larger than the other benchmark datasets. This dataset is created from YouTube video library, where all the videos depict
stage performance of Indian classical dancers. So our dataset has huge variations in respect of lighting condition, camera position, and clothing. We have
manually annotated the following $14$ body parts: head, neck, shoulders, elbows, wrists, hips, knees and feet. We have arbitrarily marked 5 video clips of each dance
style as train data (30 video clips altogether) and the remaining 30 video clips as test data.

{\bf Outdoor Pose dataset\footnote{http://www.cs.cmu.edu/~vramakri/Data/}: } This is a full body pose dataset.
It consists of $6$ videos where $4$ different actors perform different actions in a outdoor environment. It has total $828$ frames each with annotated $14$ body joints
(head, neck, shoulders, elbows, wrists, hips, knees and feet).

\subsection{Experimental settings}
\label{subsec:Experimental_settings}

Like all others, in our proposed human pose tracking algorithm some parameters need to be fixed. We choose these parameters experimentally based on training samples of VideoPose2 dataset and use most of these relevant parameter values for other datasets. For body part
descriptor we fix $m = 10$ (see section~\ref{subsec:Experimental_evaluation_and_discussion})  as the number of concentric annular regions for all the experiments. We have used maximum likelihood parameter
estimation method to estimate the parameters of the Gaussian distribution for temporal displacement of each part. For each part different datasets exibit different
relative displacement distribution (see Fig.~\ref{fig:chap5:videopose_upper_body_part_position},~\ref{fig:chap5:icdpose_upper_body_part_position}
and~\ref{fig:chap5:icdpose_lower_body_part_position}). We have experimentally seen that for each part over all datasets six Gaussian distribution functions can represent the relative displacement faithfully. So we fix the
number of clusters $k = 6$ and use maximum likelihood parameter estimation method to estimate
the Gaussian distribution parameters of each cluster. Two regularization parameters are fixed as $\lambda_{1} = 0.7$ and $\lambda_{2} = 0.2$ experimentally based on VideoPose2 dataset.

{\bf Evaluation metric: } We use the key point localization error as the evaluation metric~\cite{BenjaminSappCVPR11}. In this metric, for each body part (joint), we
calculate the pixel location deviation, i.e., the distance between the estimated location and corresponding ground truth location.
Then for a video we compute the percentage of frames, where this distance is less than an acceptable deviation
threshold $\Omega$  as average accuracy. Here we present the results for $\Omega = 5, 10, 15, 20, 25, 30, 35,$ and $40$ pixels.

\subsection{Experimental evaluation and discussion}
\label{subsec:Experimental_evaluation_and_discussion}
\begin{figure}
  \begin{center}
	\includegraphics[width = 0.60\textwidth]{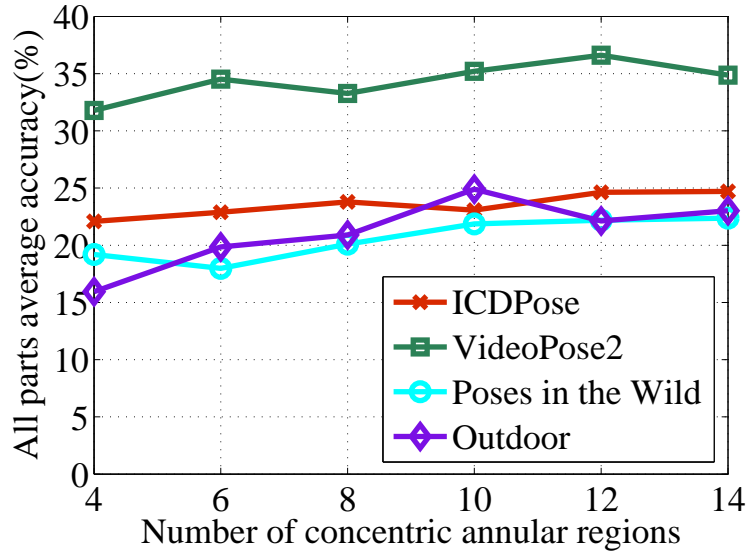} \\	
    \caption{Comparative results of tracking body parts using proposed descriptor with different number of concentric annular regions on different datasets (VideoPose2,  Poses in the Wild, ICDPose and Outdoor Pose)}
    \label{fig:chap5:local_nbr_diff_rect_size_feature_comparison}
  \end{center}
\end{figure}
\begin{figure}
  \begin{center}
    $\begin{array}{@{\hspace{1pt}}c@{\hspace{1pt}}c@{\hspace{1pt}}c}
	\includegraphics[width = 0.3\textwidth]{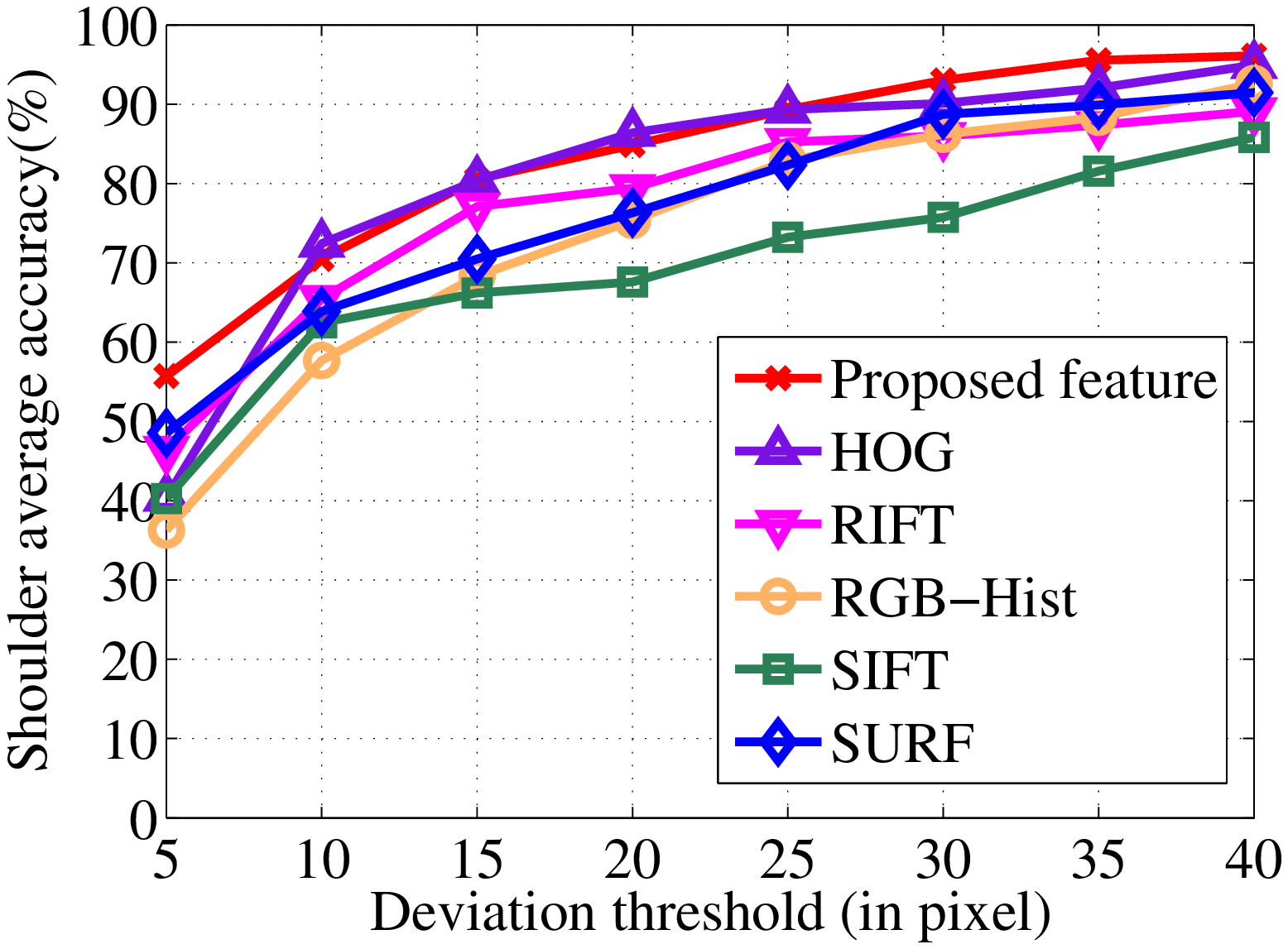} &
    \includegraphics[width = 0.3\textwidth]{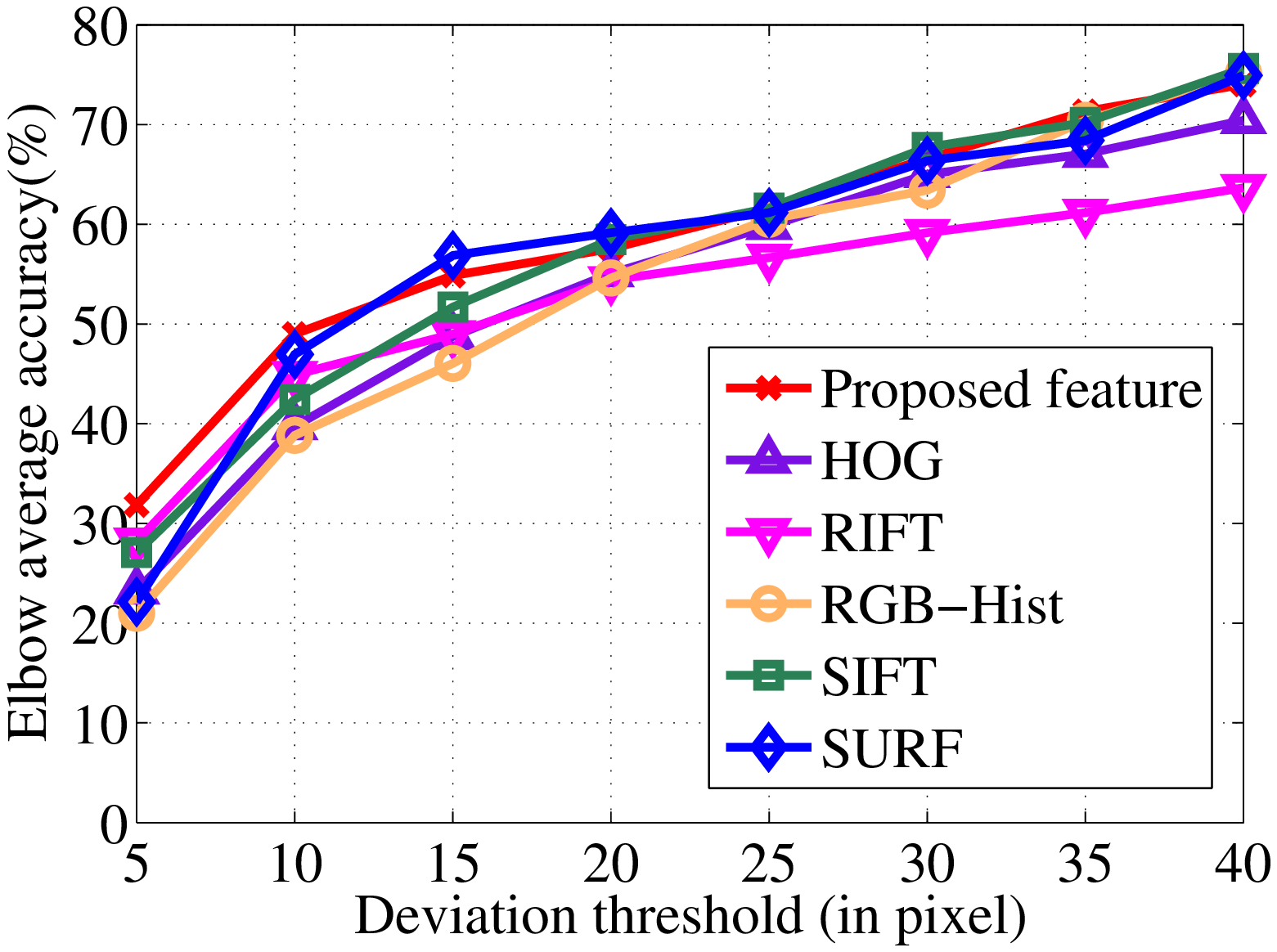} &
	\includegraphics[width = 0.3\textwidth]{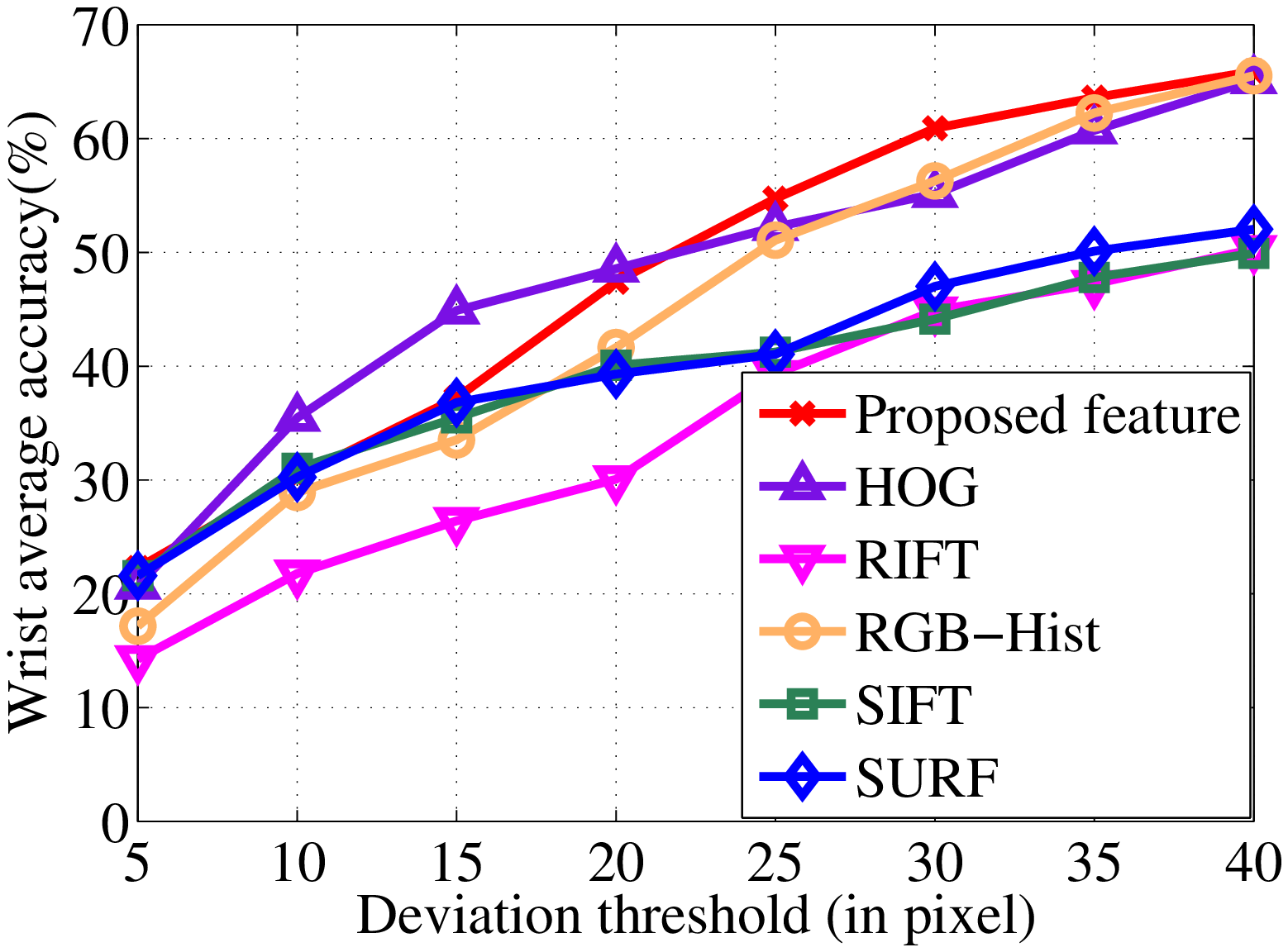} \\
	(a)	&	(b)  &   (c)
    \end{array} $
    \caption{Comparative results of tracking different body parts independently using different features on VideoPose2 dataset: (a) shoulder, (b) elbow, and (c) wrist.}
    \label{fig:chap5:videopose_local_descriptor_comparison}
  \end{center}
\end{figure}
\begin{figure}
  \begin{center}
    $\begin{array}{@{\hspace{1pt}}c@{\hspace{1pt}}c@{\hspace{1pt}}c}
	\includegraphics[width = 0.3\textwidth]{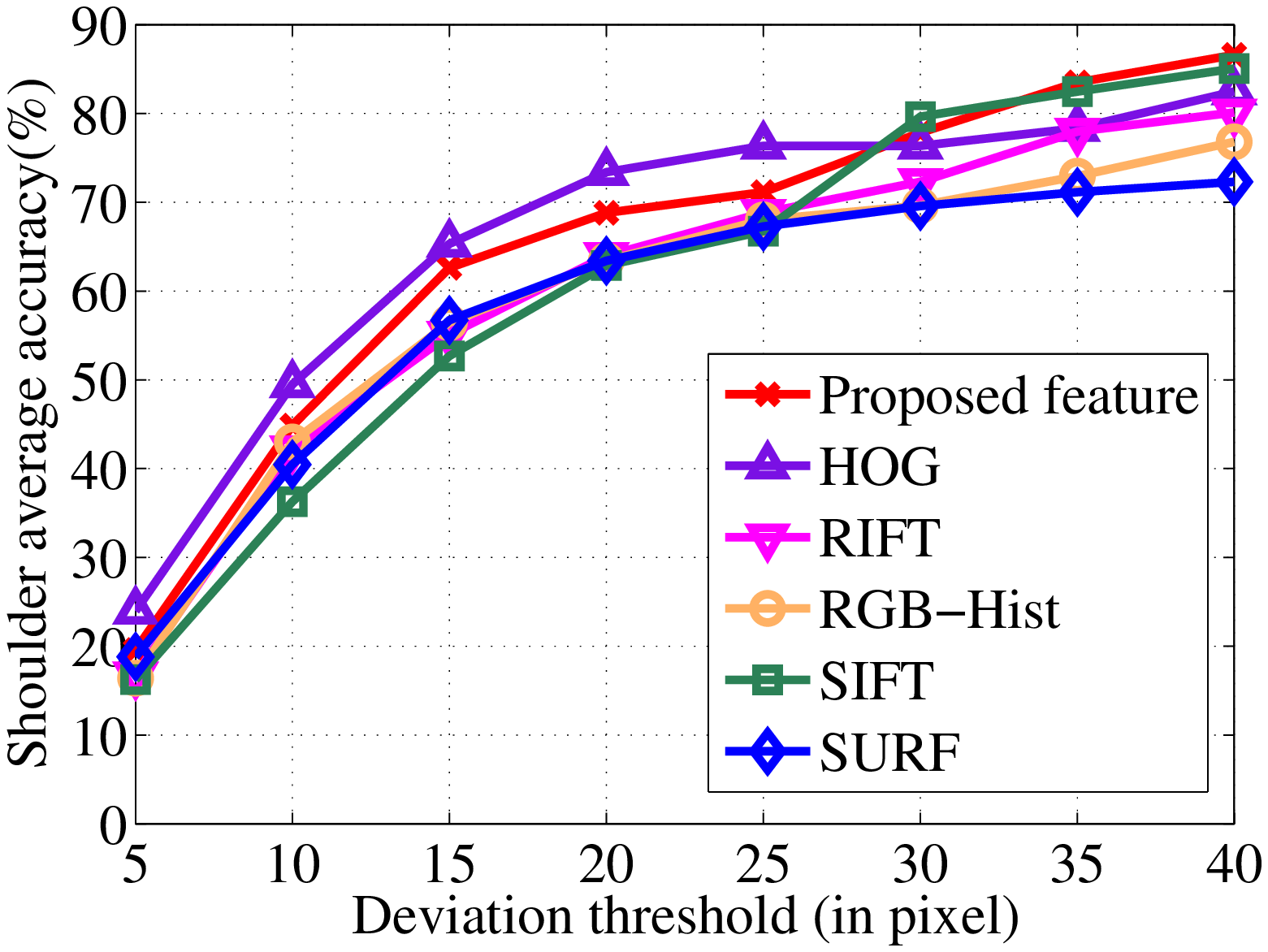} &
    \includegraphics[width = 0.3\textwidth]{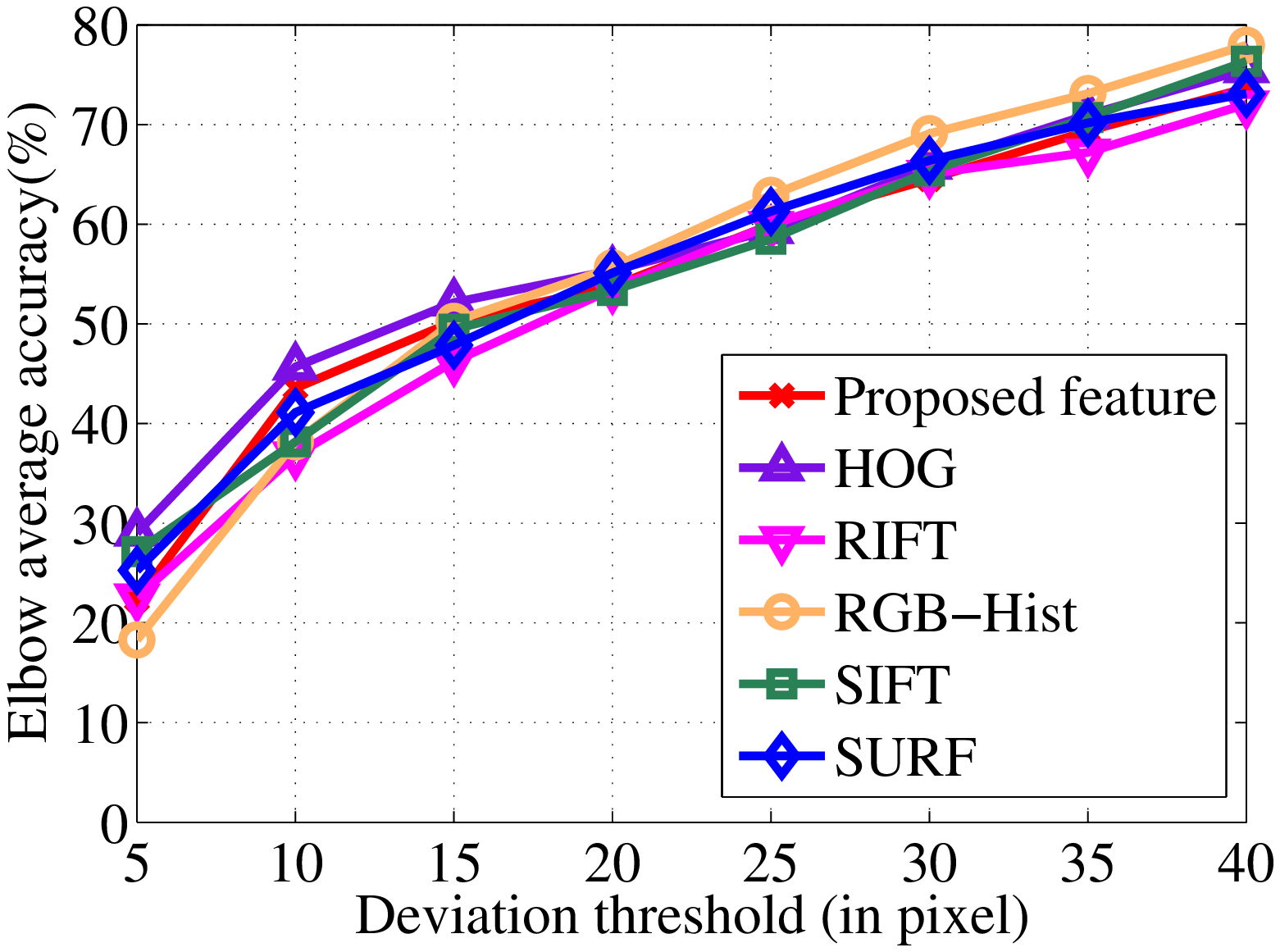} &
	\includegraphics[width = 0.3\textwidth]{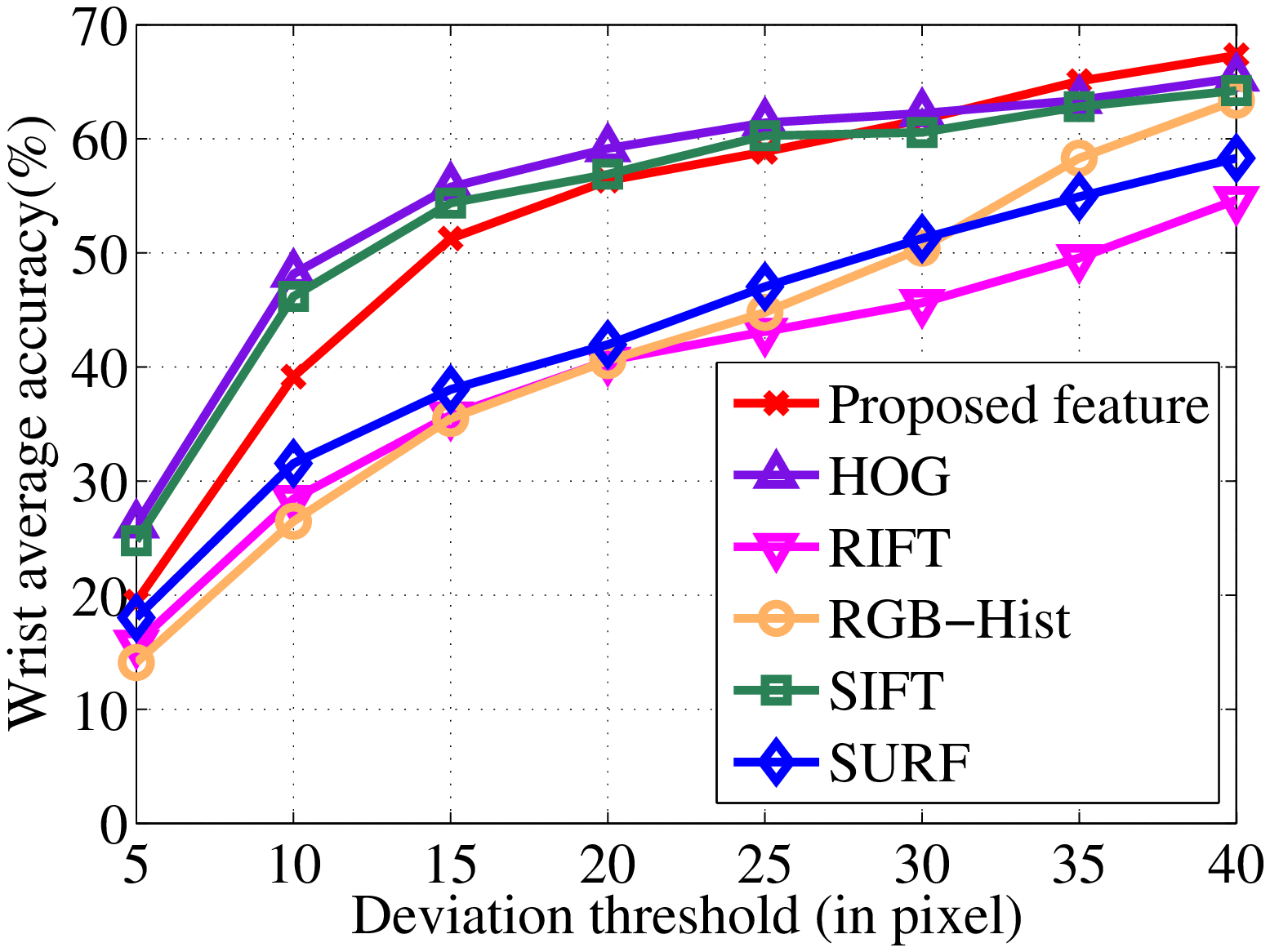} \\
	(a)	&	(b)  &   (c)
    \end{array} $
    \caption{Comparative results of tracking different body parts independently using different features on Poses in the Wild dataset: (a) shoulder, (b) elbow, and (c) wrist.}
    \label{fig:chap5:wildpose_local_descriptor_comparison}
  \end{center}
\end{figure}
\begin{figure}
  \begin{center}
    $\begin{array}{@{\hspace{1pt}}c@{\hspace{1pt}}c@{\hspace{1pt}}c}
	\includegraphics[width = 0.3\textwidth]{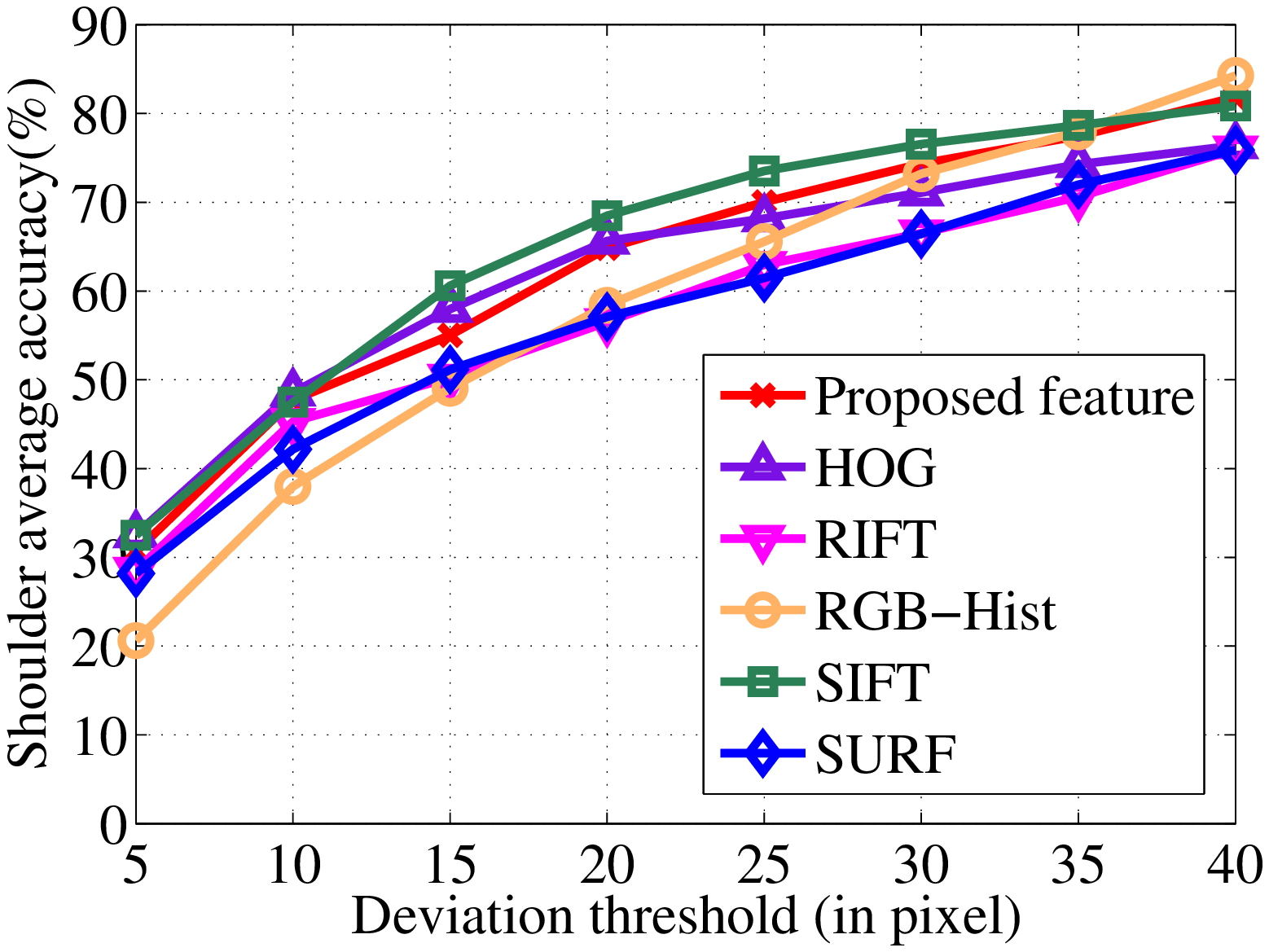} &
    \includegraphics[width = 0.3\textwidth]{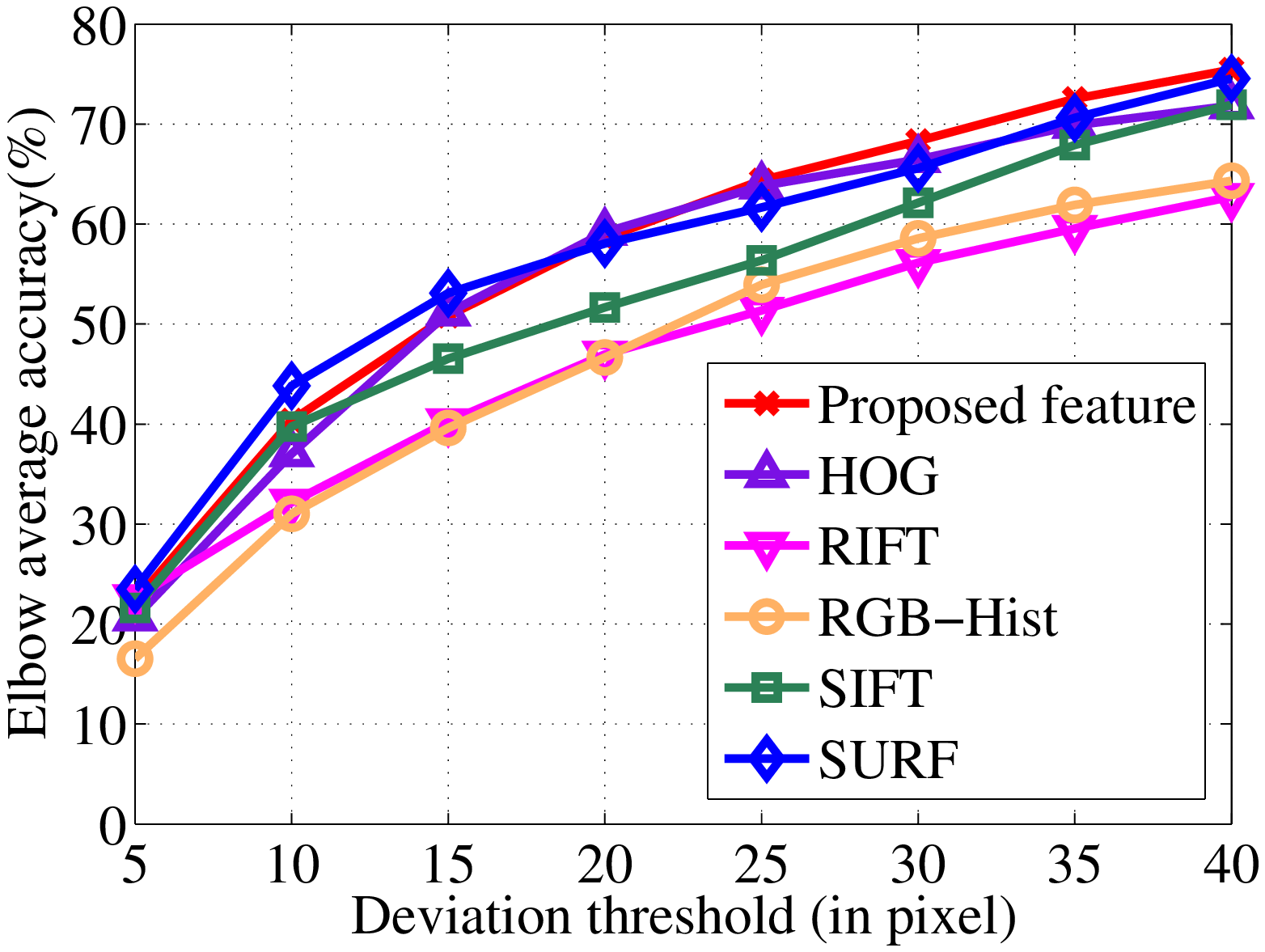} &
	\includegraphics[width = 0.3\textwidth]{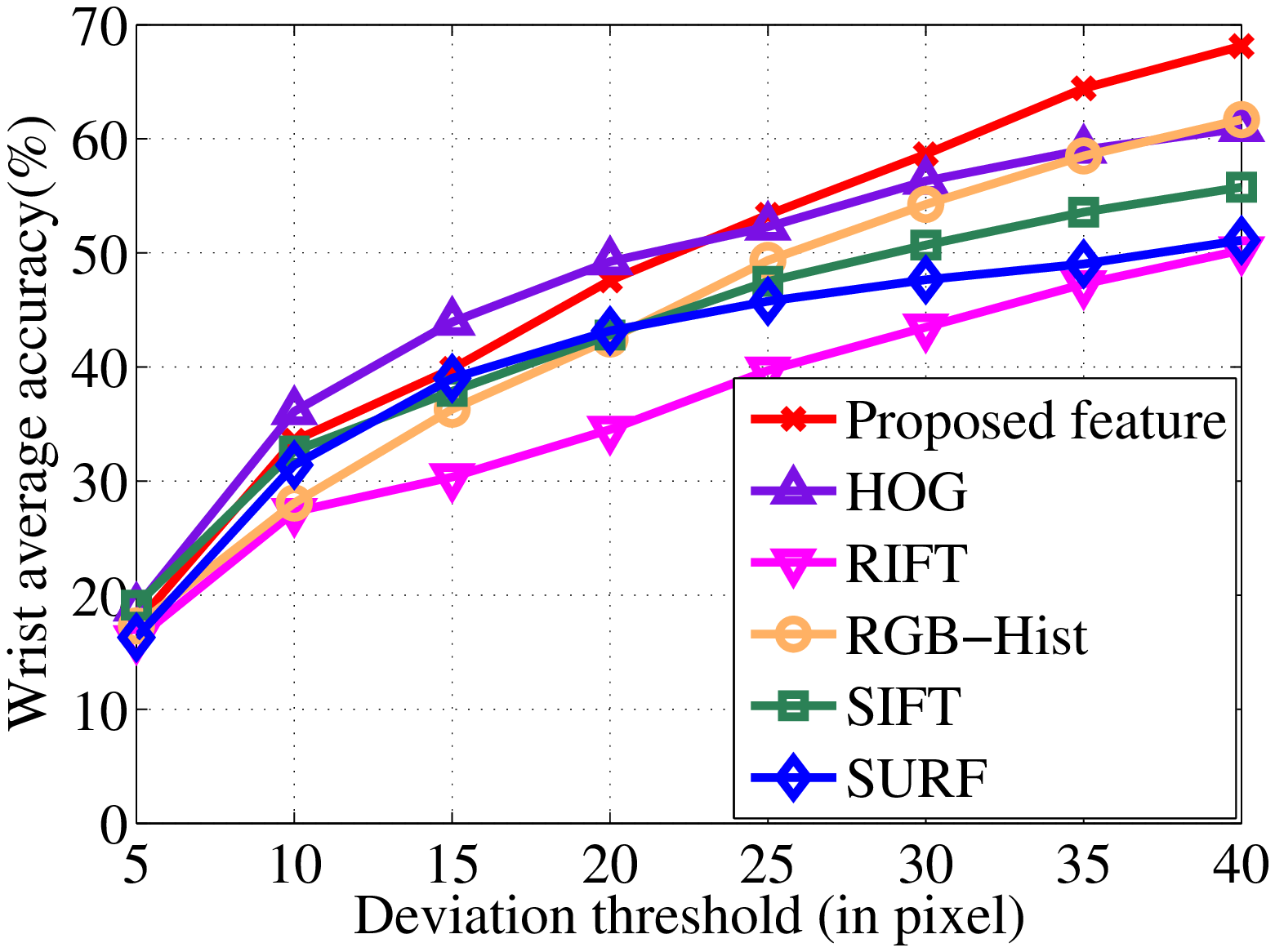} \\
	(a)	&	(b)  &   (c) \\
    \includegraphics[width = 0.3\textwidth]{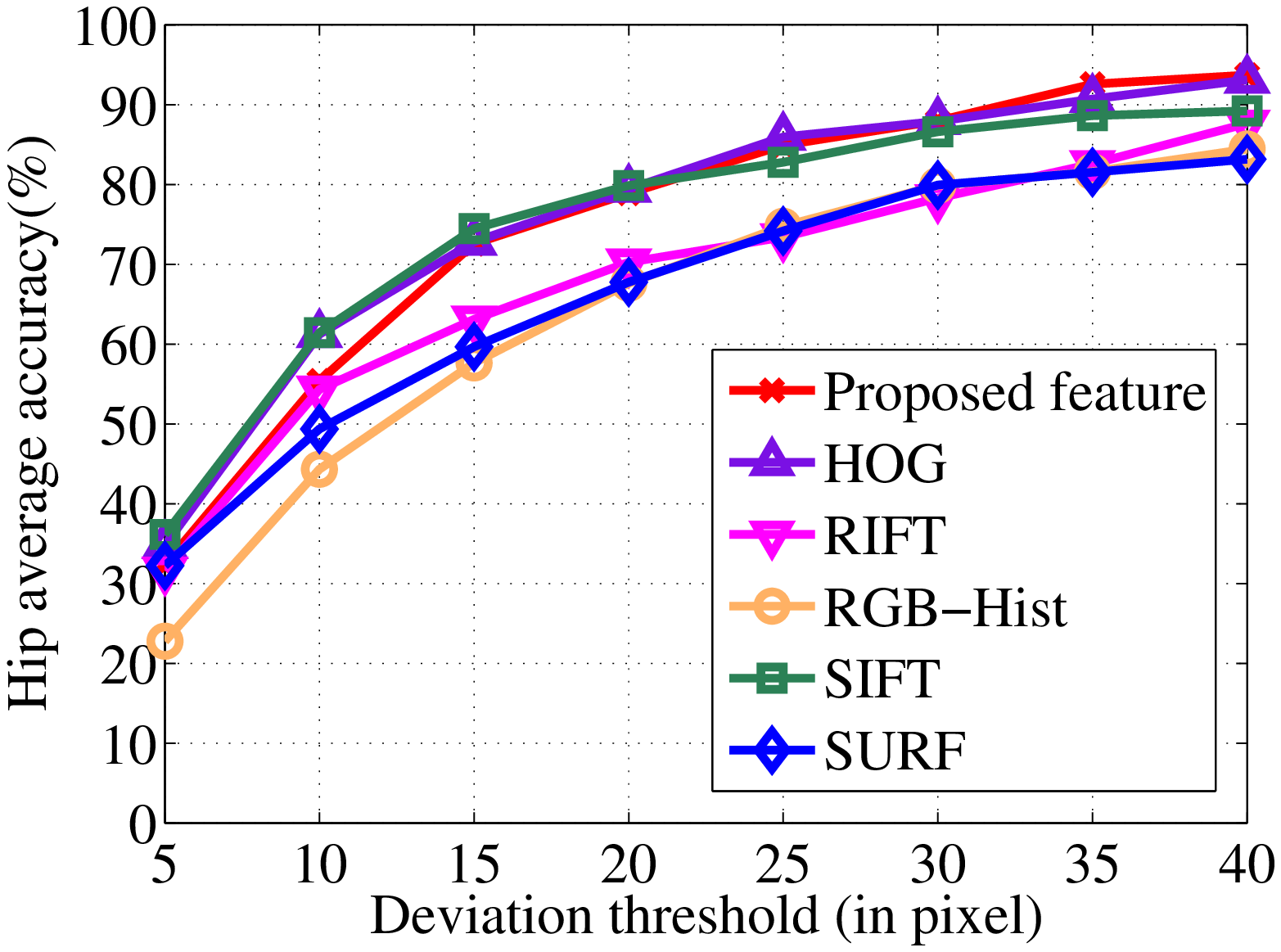} &
    \includegraphics[width = 0.3\textwidth]{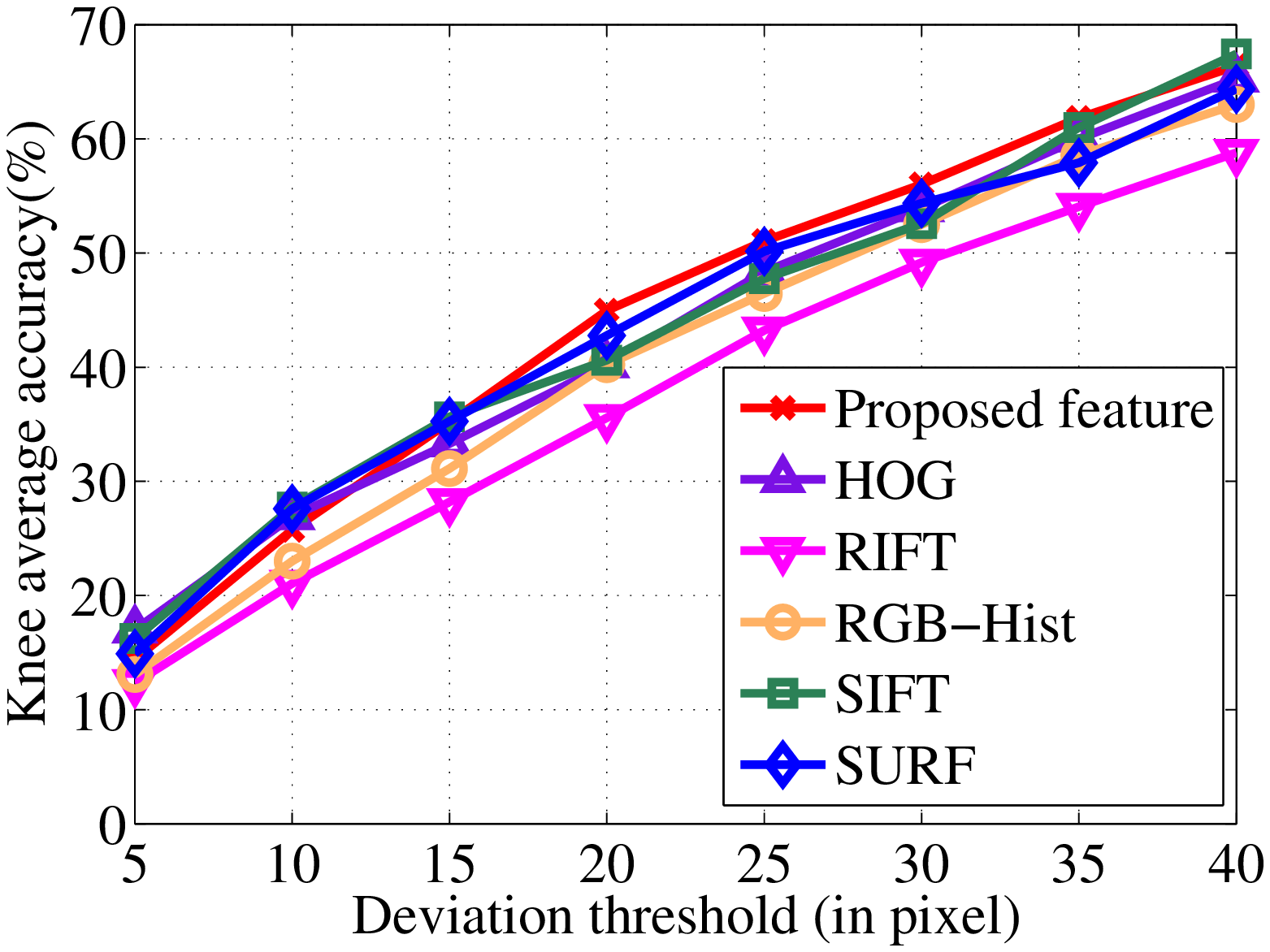} &
	\includegraphics[width = 0.3\textwidth]{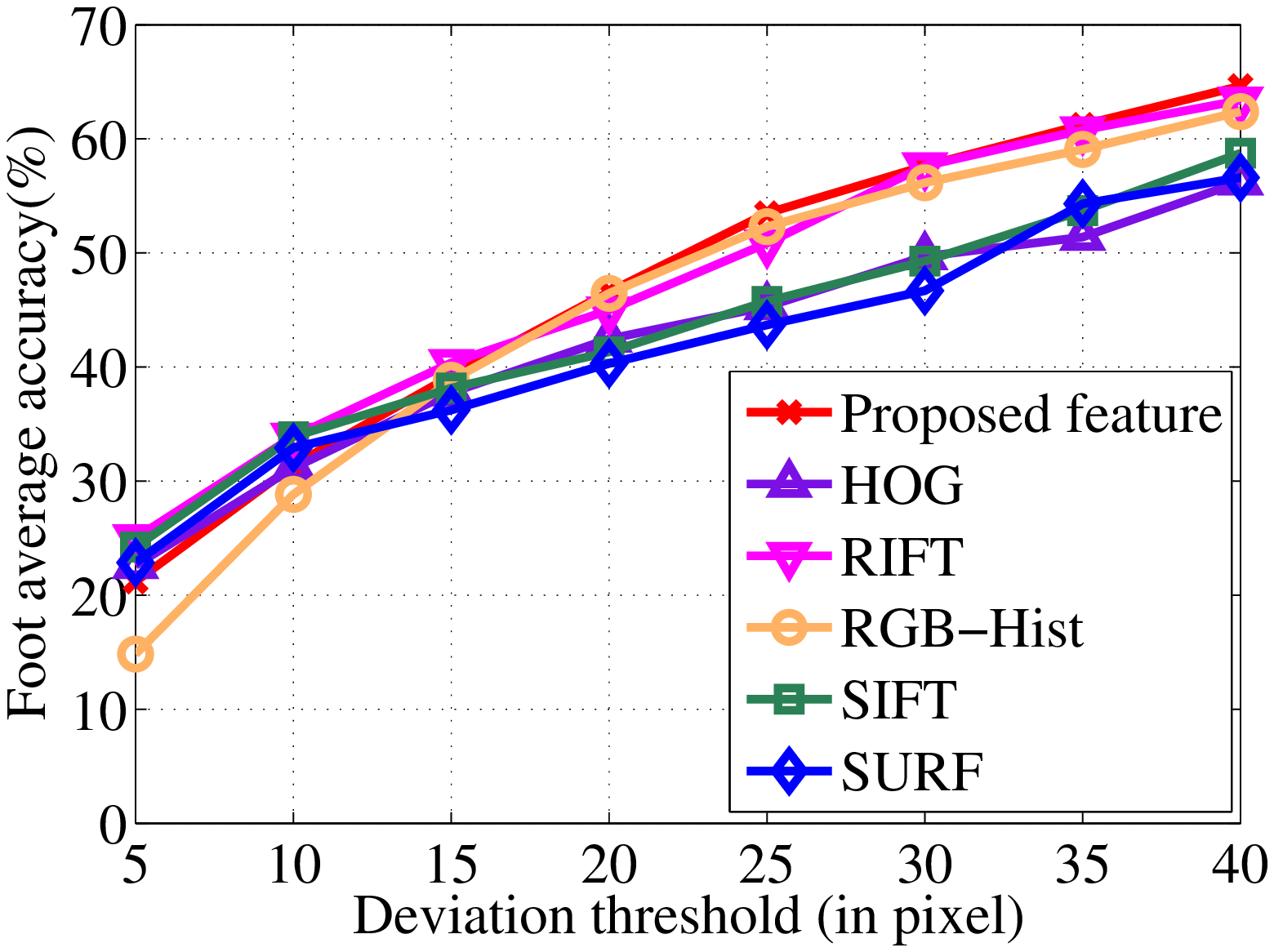} \\
	(d)	&	(e)  &   (f)
    \end{array} $
    \caption{Comparative results of tracking different body parts independently using different features on ICDPose dataset: (a) shoulder, (b) elbow, (c) wrist, (d) hip, (e) knee, and (f) foot.}
    \label{fig:chap5:icdpose_local_descriptor_comparison}
  \end{center}
\end{figure}
\begin{figure}
  \begin{center}
    $\begin{array}{@{\hspace{1pt}}c@{\hspace{1pt}}c@{\hspace{1pt}}c}
	\includegraphics[width = 0.3\textwidth]{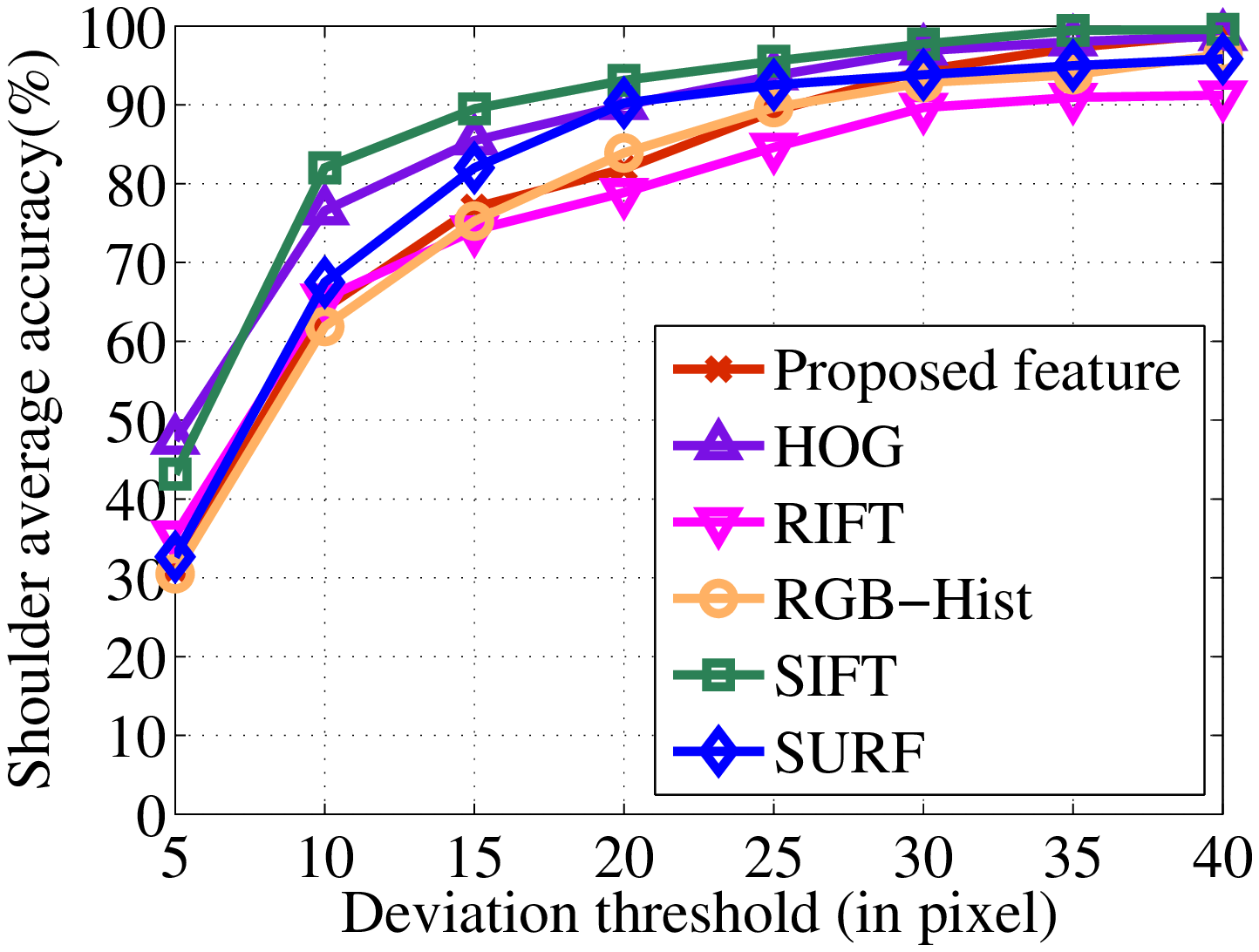} &
    \includegraphics[width = 0.3\textwidth]{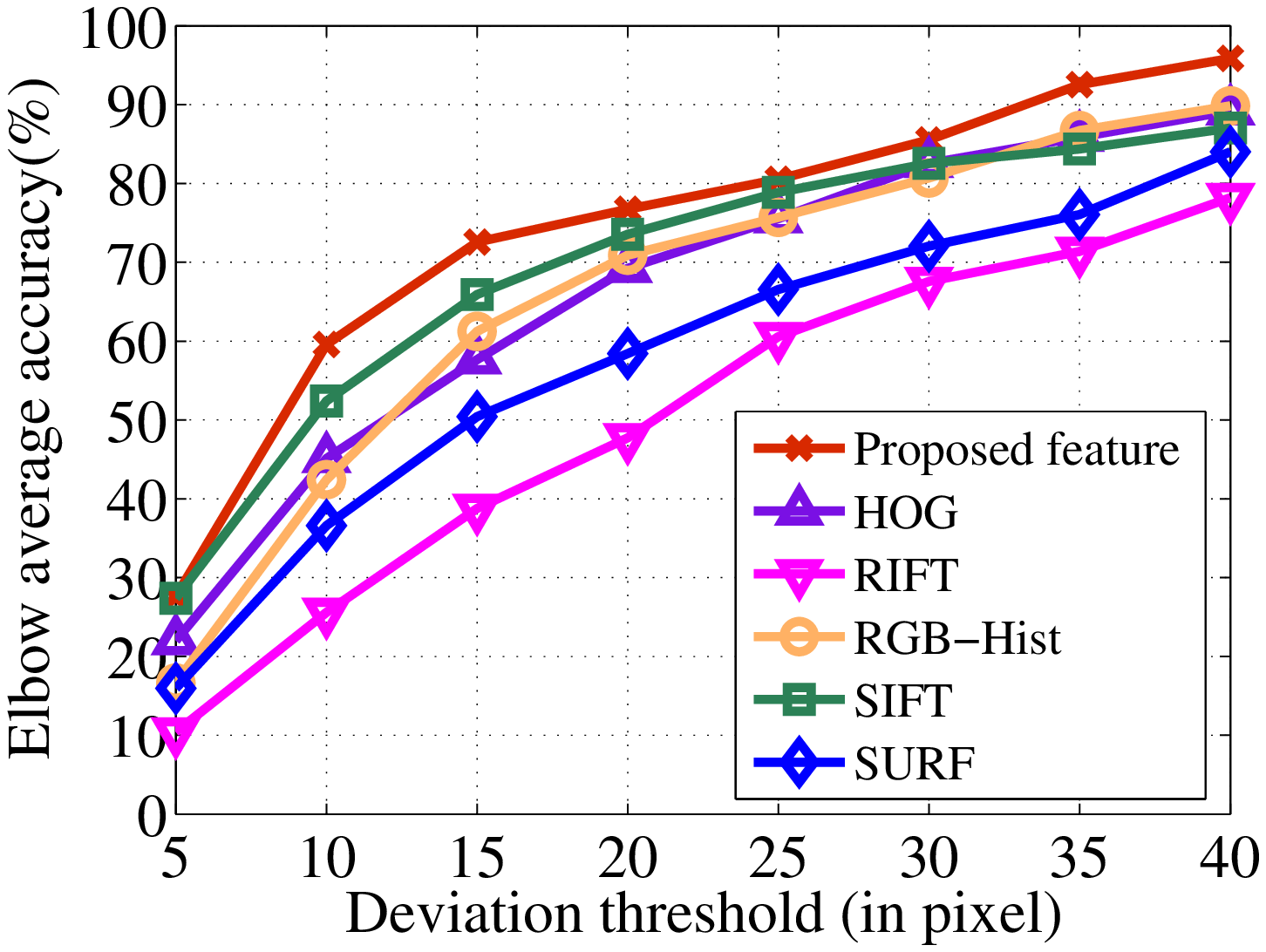} &
	\includegraphics[width = 0.3\textwidth]{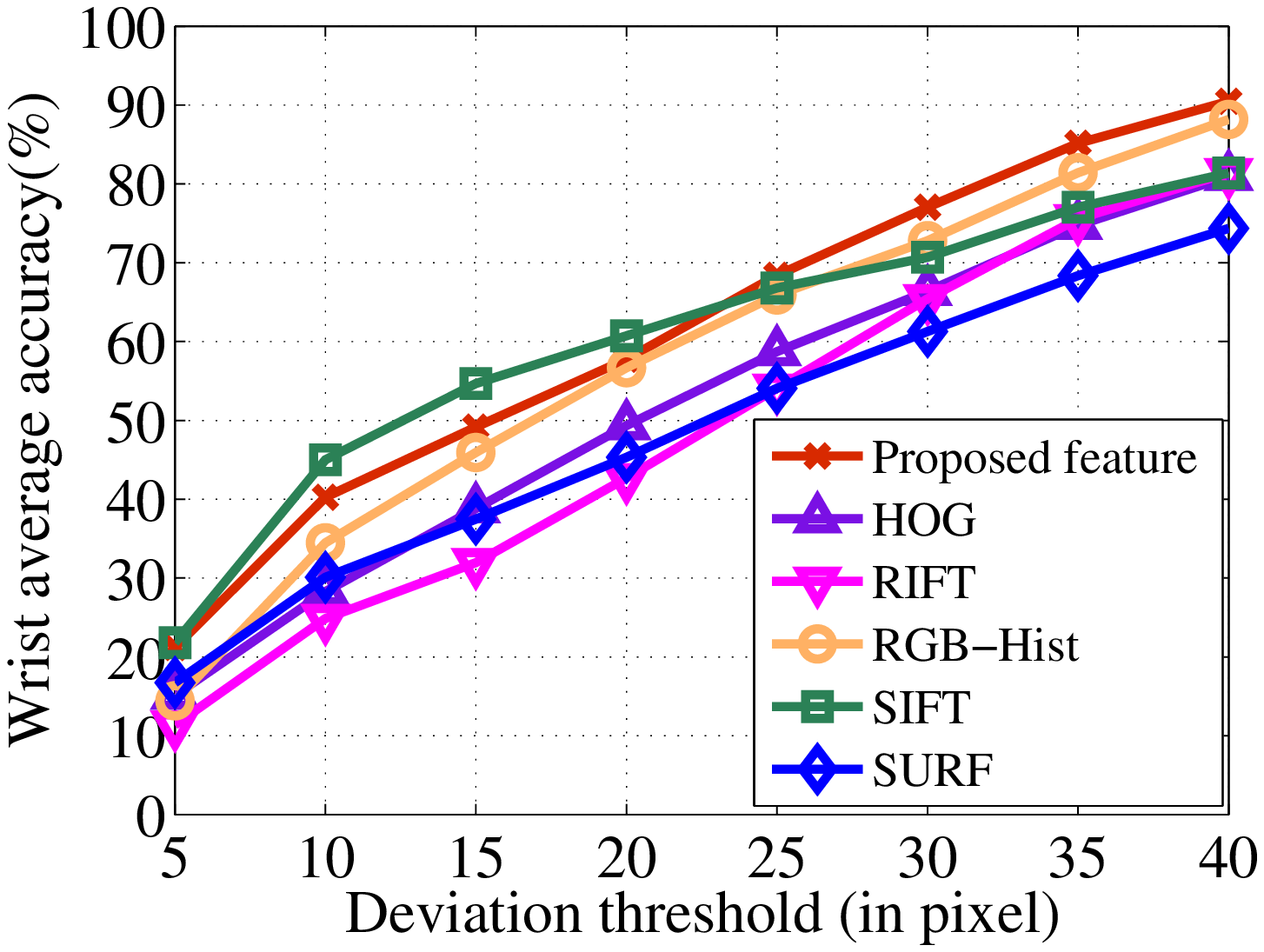} \\
	(a)	&	(b)  &   (c) \\
    \includegraphics[width = 0.3\textwidth]{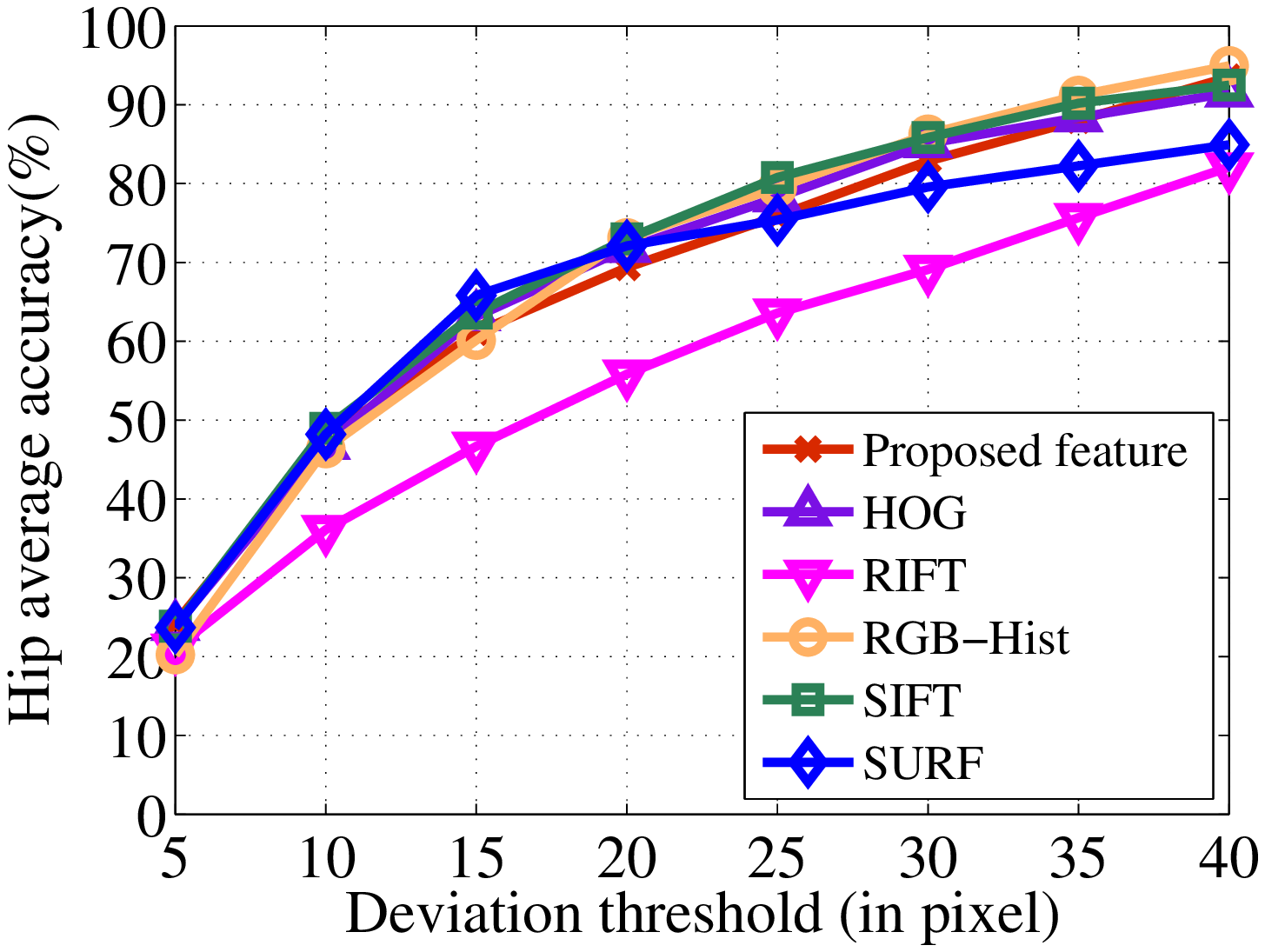} &
    \includegraphics[width = 0.3\textwidth]{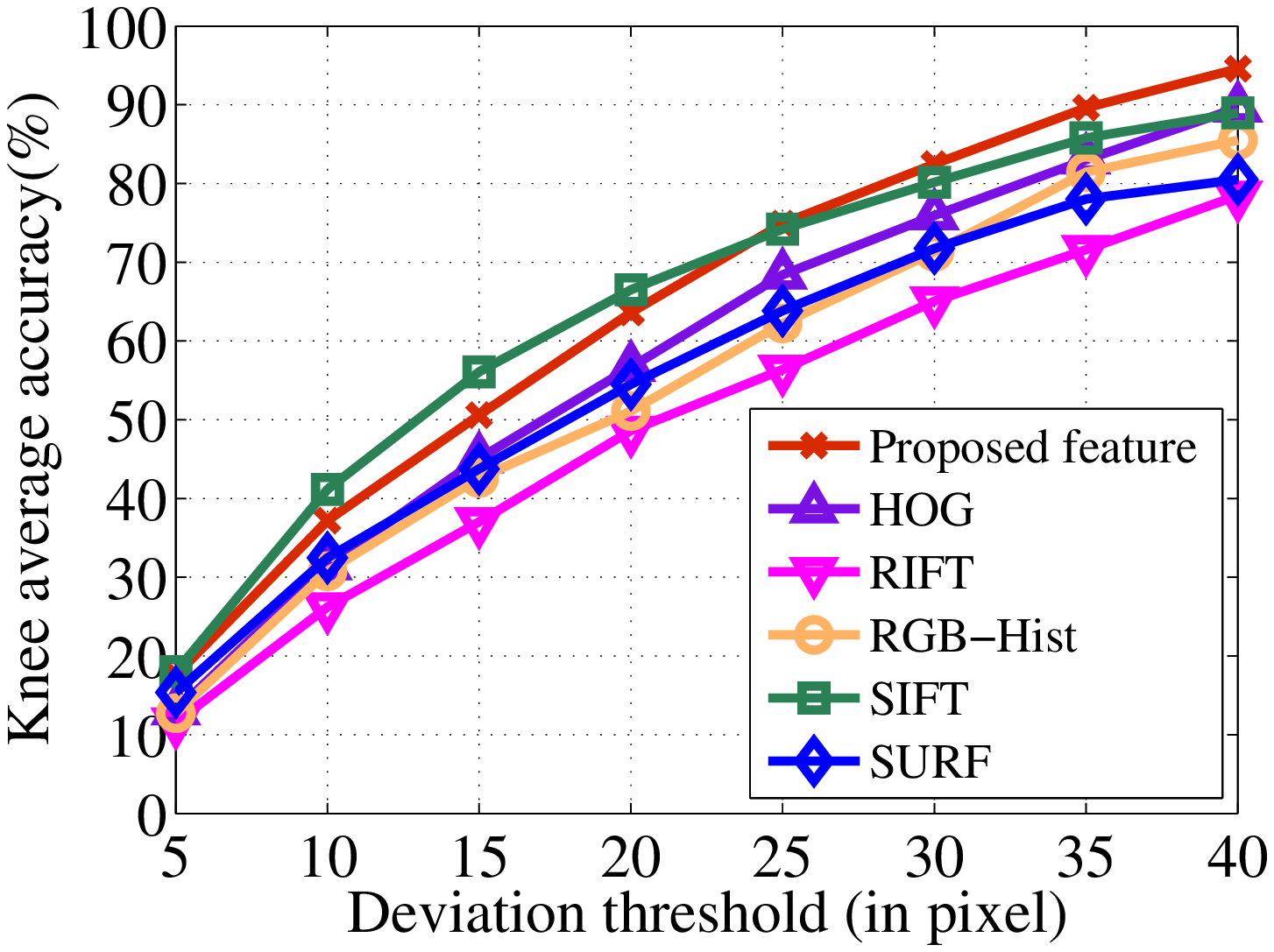} &
	\includegraphics[width = 0.3\textwidth]{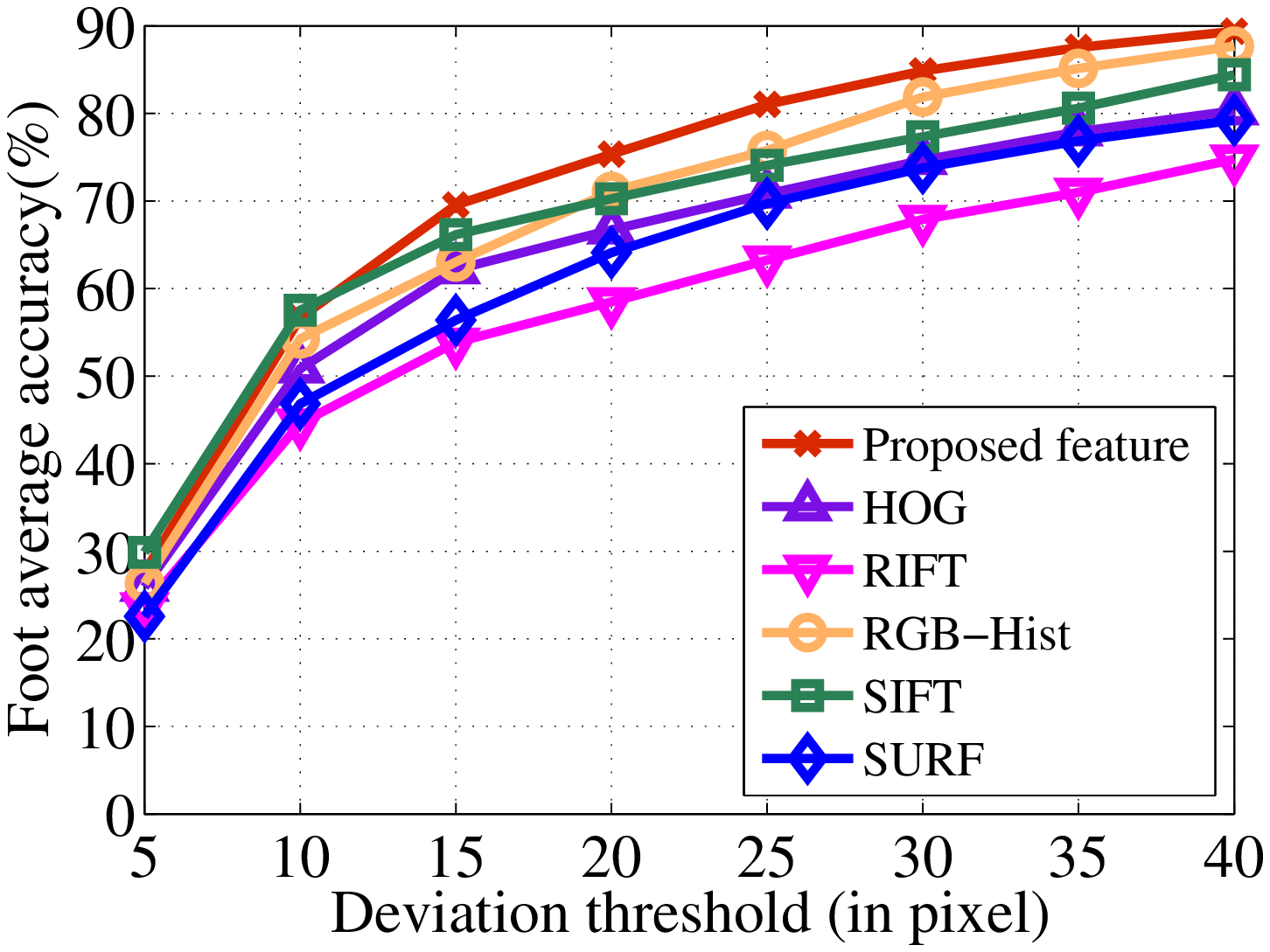} \\
	(d)	&	(e)  &   (f)
    \end{array} $
    \caption{Comparative results of tracking different body parts independently using different features on Outdoor Pose dataset: (a) shoulder, (b) elbow, (c) wrist, (d) hip, (e) knee, and (f) foot.}
    \label{fig:chap5:outdoor_local_descriptor_comparison}
  \end{center}
\end{figure}

{\bf Proposed feature evaluation:} First, we determine the number $m$ of concentric annular regions to describe body parts (joints) by proposed methods. We have tried different values of $m$ from $4$ to $14$, and calculated the average accuracy of locating all body parts in various datasets ($\Omega = 5$). The results are summarised in Fig.~\ref{fig:chap5:local_nbr_diff_rect_size_feature_comparison}. The experiment suggests $m=10$ considering cost and accuracy. We then evaluate the performance of proposed part descriptor for conventional object tracking and compare it with other standard descriptors like RGB-histogram, HOG~\cite{NavneetDalalCVPR05}, RIFT~\cite{SvetlanaLazebnikBMVC04}, SIFT~\cite{DavidGLoweIJCV04}, and SURF~\cite{HerbertBayCVIU08} features. We track each part of the human body independently over a video based on their appearance measure only. Here similarity is measured using Euclidean distance between the descriptor template and the relevant frame. Figs.~\ref{fig:chap5:videopose_local_descriptor_comparison},~\ref{fig:chap5:wildpose_local_descriptor_comparison}, ~\ref{fig:chap5:icdpose_local_descriptor_comparison}, and~\ref{fig:chap5:outdoor_local_descriptor_comparison} show the average accuracy in tracking different body parts using various features for VideoPose2, Poses in the Wild, ICDPose and Outdoor Pose datasets respectively.
From these figures we see that the proposed feature gives better or at least  comparable result compared to the others. These figures also suggest that the accuracy decreases if we move from slow moving part to faster moving parts (i.e., shoulder to elbow to wrist).

Table~\ref{tab:chap5:feature_time_comparison} shows the average time taken to process each frame (in second) of various datasets and reveals that the proposed feature computation is faster
than the others. We have already mentioned that (in first paragraph of Section~\ref{sec:Experimental_result}) our feature implementation has been done in MATLAB (with no mex file interface). For other feature like HOG, we have used Yang et al.~\cite{YiYangIEEETPAMI13} implementation with bin size $4$. We have implemented the RIFT feature in MATLAB with parameters suggested in~\cite{SvetlanaLazebnikBMVC04} (i.e., four concentric ring and eight histogram orientation).
For SURF feature,  we have used MATLAB R2013a SURF feature implementation. We have used Ce Liu's dense SIFT feature implementation with default parameters~\cite{SIFTflow}.
\noindent
\begin{table*}
\caption{Average time comparison (in sec./frame) of proposed feature with others on different datasets.}
\begin{center}
\begin{tabular}{|l|c|c|c|c|c|c|}
\hline
    Datasets                    	&  	RGB-hist   	&   HOG~\cite{NavneetDalalCVPR05}      &   RIFT~\cite{SvetlanaLazebnikBMVC04}      &   SIFT~\cite{DavidGLoweIJCV04}	&   SURF~\cite{HerbertBayCVIU08}	&  Proposed \\ \hline
  VideoPose2~\cite{BenjaminSappCVPR11}		            &   1.0173      &   1.1166      &   1.1147      &   0.2277      &   1.5360      &       \bf{0.1901}	\\ \hline
  Poses in the Wild~\cite{AnoopCherianCVPR14}         &   1.0527        &   1.1011      &   1.1848      &   	0.2365     &   1.5469      &	\bf{0.1993}	\\ \hline
  ICDPose~\cite{ICDPose}		                                    &	1.8711      &   1.6978      &   1.8157      &   0.5040      &   2.7237				&	\bf{0.4396}	\\ \hline
  Outdoor Pose~\cite{VarunRamakrishnaCVPR13}        &   1.7909      &   1.6121      &   1.7402      &   0.4144      &   2.5802      &   	\bf{0.3527}    \\
\hline
\end{tabular}
\end{center}
\label{tab:chap5:feature_time_comparison} 
\end{table*}
\begin{figure}
  \begin{center}
    $\begin{array}{@{\hspace{1pt}}c@{\hspace{1pt}}c@{\hspace{1pt}}c}
	\includegraphics[width = 0.3\textwidth]{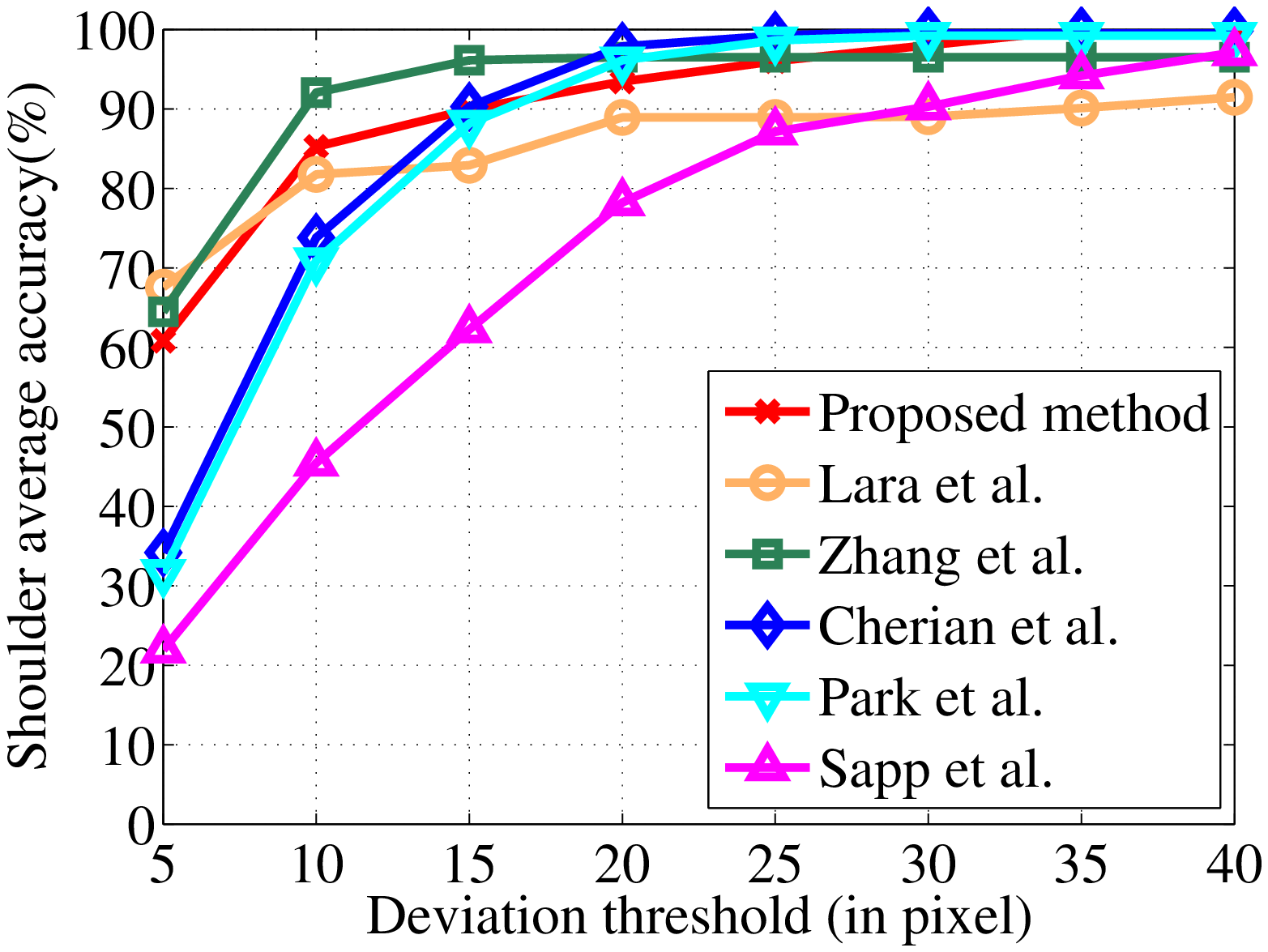} &
    \includegraphics[width = 0.3\textwidth]{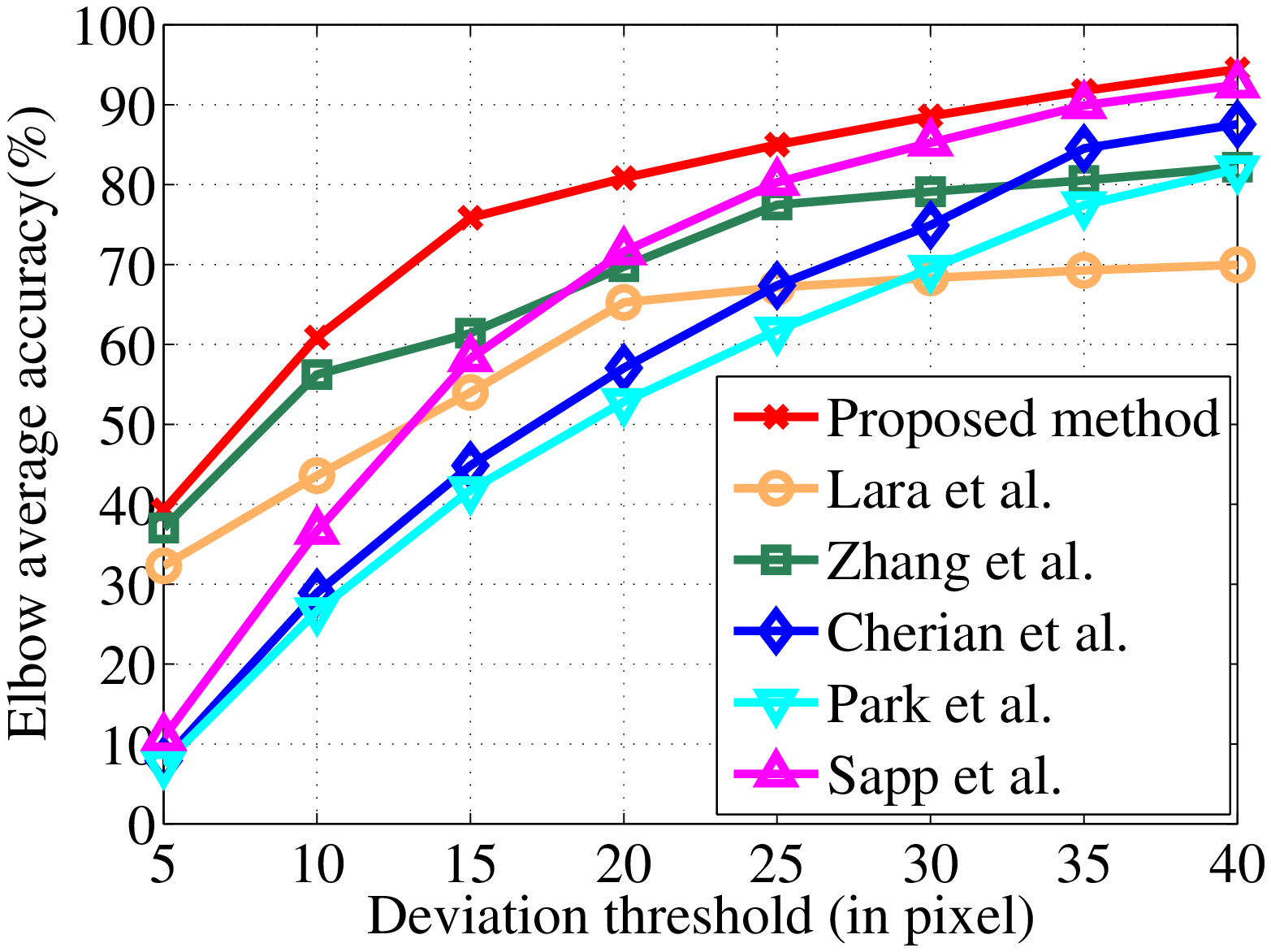} &
	\includegraphics[width = 0.3\textwidth]{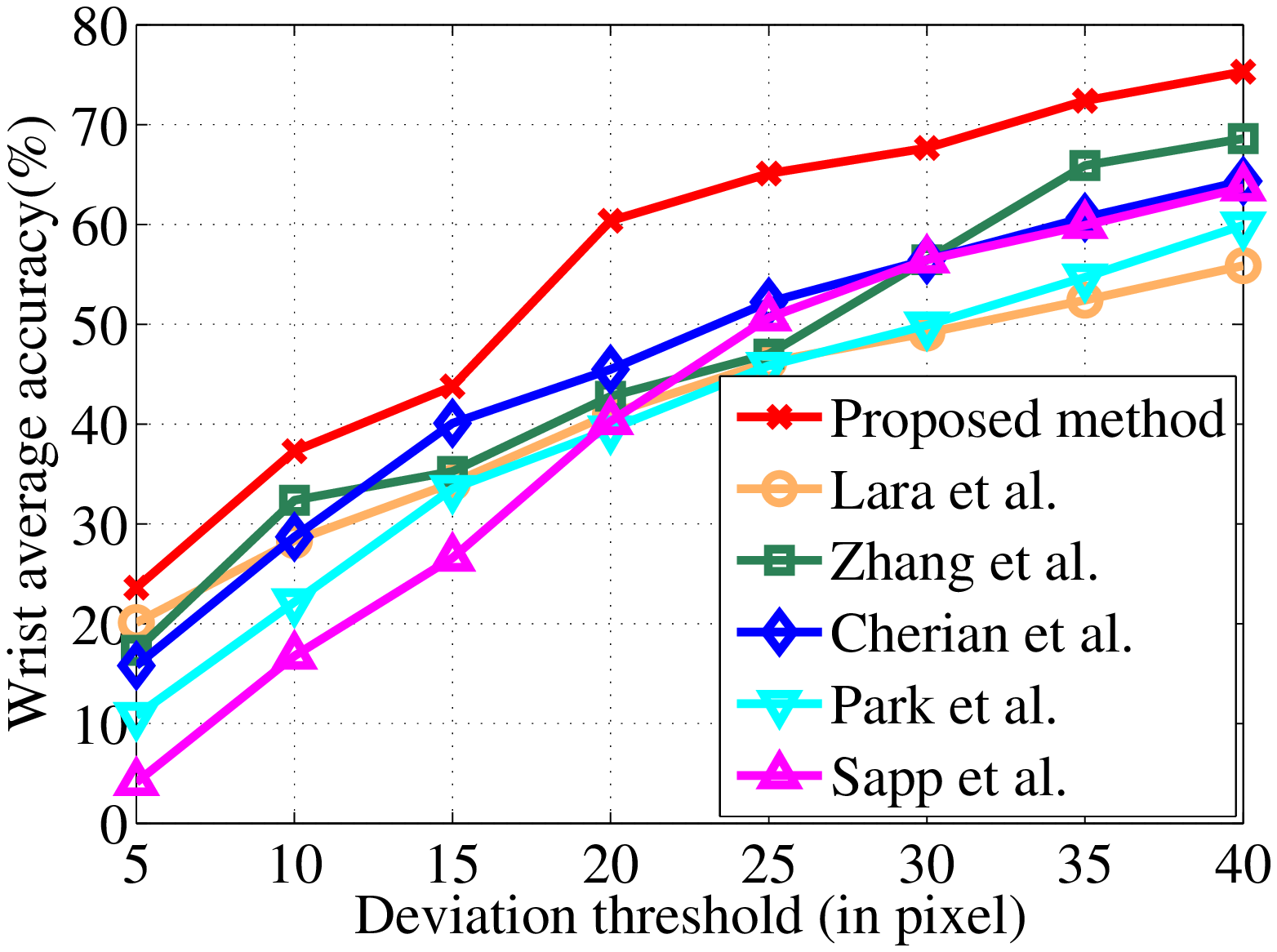} \\
	(a)	&	(b)  &   (c)
    \end{array} $
    \caption{Comparative results of proposed method with the state-of-the-art methods for human pose tracking on VideoPose2 dataset: (a) shoulder, (b) elbow, and (c) wrist.}
    \label{fig:chap5:videopose_state_of_the_art_comparison}
  \end{center}
\end{figure}

So based on the experimental observations stated in the previous paragraph we use the proposed descriptor to track the human pose in all the datasets by optimizing the objective functions~(\ref{eqn:chap5:regularized_x_i_t_*}) and~(\ref{eqn:chap5:regularized_x_root_t_*}). We compare our result for
human pose tracking with that of the state-of-the-art pose estimation methods~\cite{BenjaminSappCVPR11}, \cite{DennisParkICCV11} and \cite{AnoopCherianCVPR14}. We also compare our results with that of the state-of-the-art individual object tracking methods~\cite{LauraSevilla-LaraCVPR12, KaihuaZhangECCV14}, where each of the body parts separately fed to these methods and each body part is treated as a single object. These two methods are brought into comparison to show the effect of spatial constraints provided by the connected body parts over the independent individual part tracking. Note that for all these methods we have used their implementation with default parameter settings.

{\bf Experimental results on VideoPose2 dataset:} For VideoPose2 dataset we use the authors suggested training and test data partition. Fig.~\ref{fig:chap5:videopose_state_of_the_art_comparison} shows the comparison for different body parts like shoulder, elbow, and wrist with the state-of-the-art methods. We see that in most of the cases our method gives the superior results. Note that since motion is less in the videos of this dataset, individual part tracking methods~\cite{LauraSevilla-LaraCVPR12}, \cite{KaihuaZhangECCV14} perform better than pose tracking methods~\cite{BenjaminSappCVPR11}, \cite{DennisParkICCV11}, \cite{AnoopCherianCVPR14}. However, the latter overtakes the formers when motion is more, i.e., for higher value of threshold. Aggregated results for all the parts and all the methods are shown in Table~\ref{tab:chap5:videopose_all_part_state_of_the_art_comparison}.
\noindent
\begin{table*}[t]
\small
\caption{Average accuracy (in \%) of pose tracking comprising three parts (shoulder, elbow, and wrist) together using different methods on VideoPose2 dataset.}
\begin{center}
\begin{tabular}{|c|c|c|c|c|c|c|}
\hline
    Dev. thrs.			&   \multicolumn{6}{c|}{Different methods}\\\cline{2-7}
    (in pixel)                        &   Lara~\cite{LauraSevilla-LaraCVPR12}  &  Zhang~\cite{KaihuaZhangECCV14}    &   Sapp~\cite{BenjaminSappCVPR11}   &   Park~\cite{DennisParkICCV11}  &  Cherian~\cite{AnoopCherianCVPR14}  &   \textbf{Proposed} \\ \hline
    5                        &   39.99       &   39.61       &   12.28       &   16.62       &   19.29       &   \bf{41.22}               \\ \hline
    10                      &   51.21       &   60.20       &   32.94       &   39.88       &   43.80       &   \bf{61.13}               \\ \hline
    15                      &   57.00       &   64.26       &   49.02       &   54.48       &   58.40       &   \bf{69.84}               \\ \hline
    20                      &   65.08       &   69.66       &   63.36       &   62.81       &   66.79       &   \bf{78.21}               \\ \hline
    25                      &   67.44       &   73.66       &   72.70       &   68.74       &   72.94       &   \bf{82.02}               \\ \hline
    30                      &   68.81       &   77.35       &   77.32       &   72.87       &   77.05       &   \bf{84.73}               \\ \hline
    35                      &   70.58       &   80.97       &   81.33       &   77.12       &   81.67       &   \bf{88.04}               \\ \hline
    40                      &   72.42       &   82.42       &   84.42       &   80.35       &   83.91       &   \bf{89.90}               \\
\hline
\end{tabular}
\end{center}
\label{tab:chap5:videopose_all_part_state_of_the_art_comparison} 
\end{table*}

\begin{figure}
  \begin{center}
    $\begin{array}{@{\hspace{1pt}}c@{\hspace{1pt}}c@{\hspace{1pt}}c}
	\includegraphics[width = 0.3\textwidth]{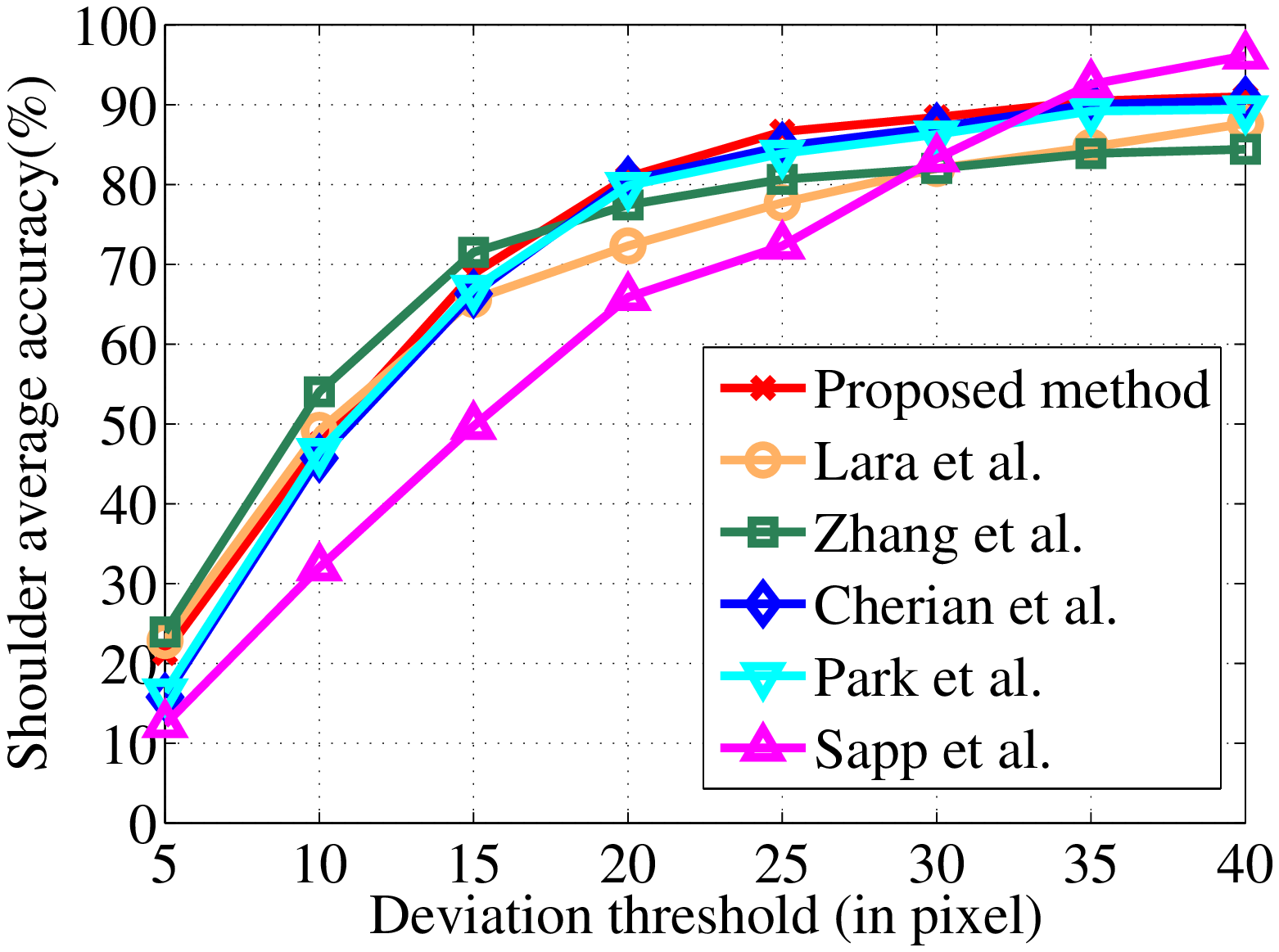} &
    \includegraphics[width = 0.3\textwidth]{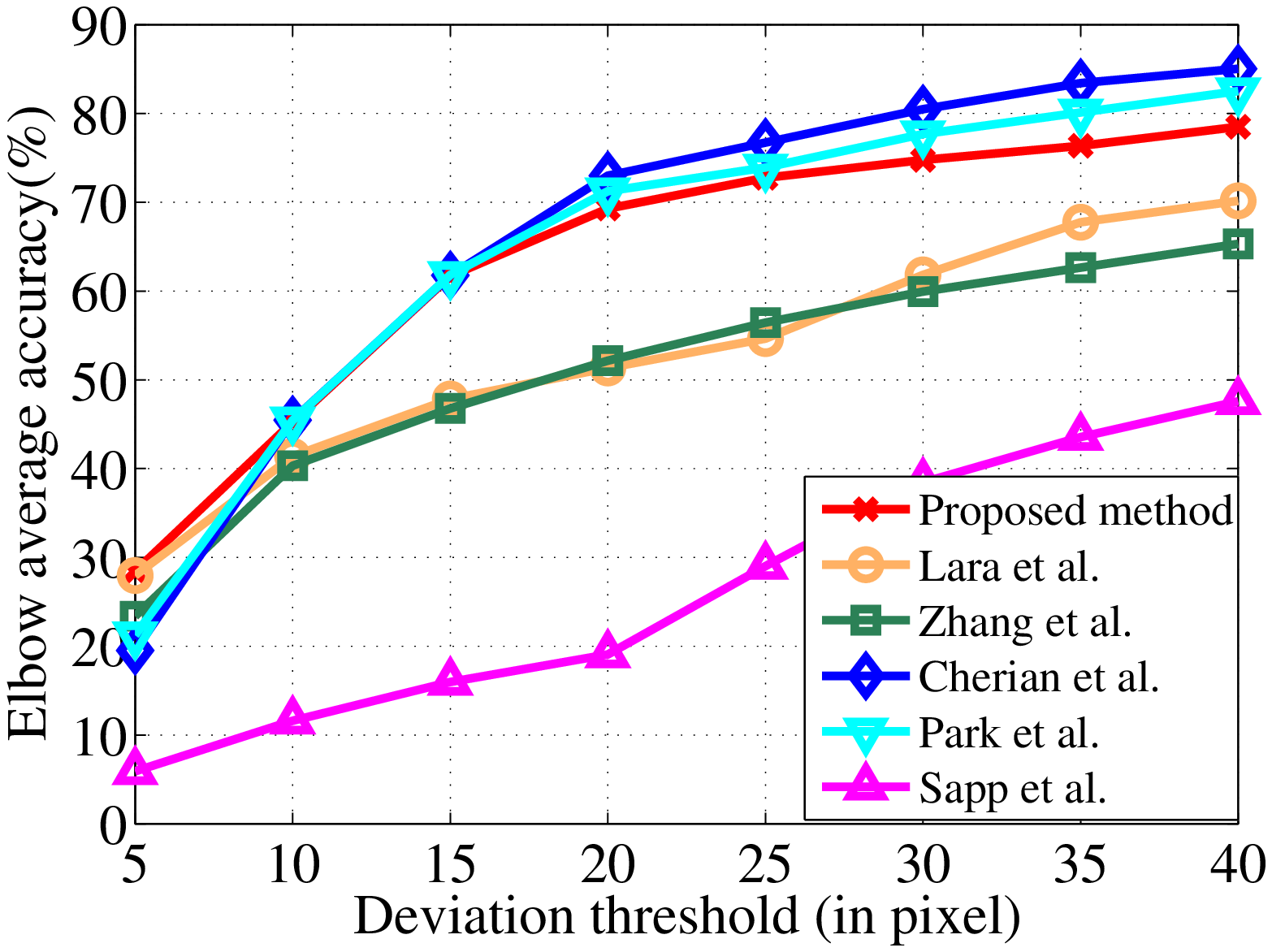} &
	\includegraphics[width = 0.3\textwidth]{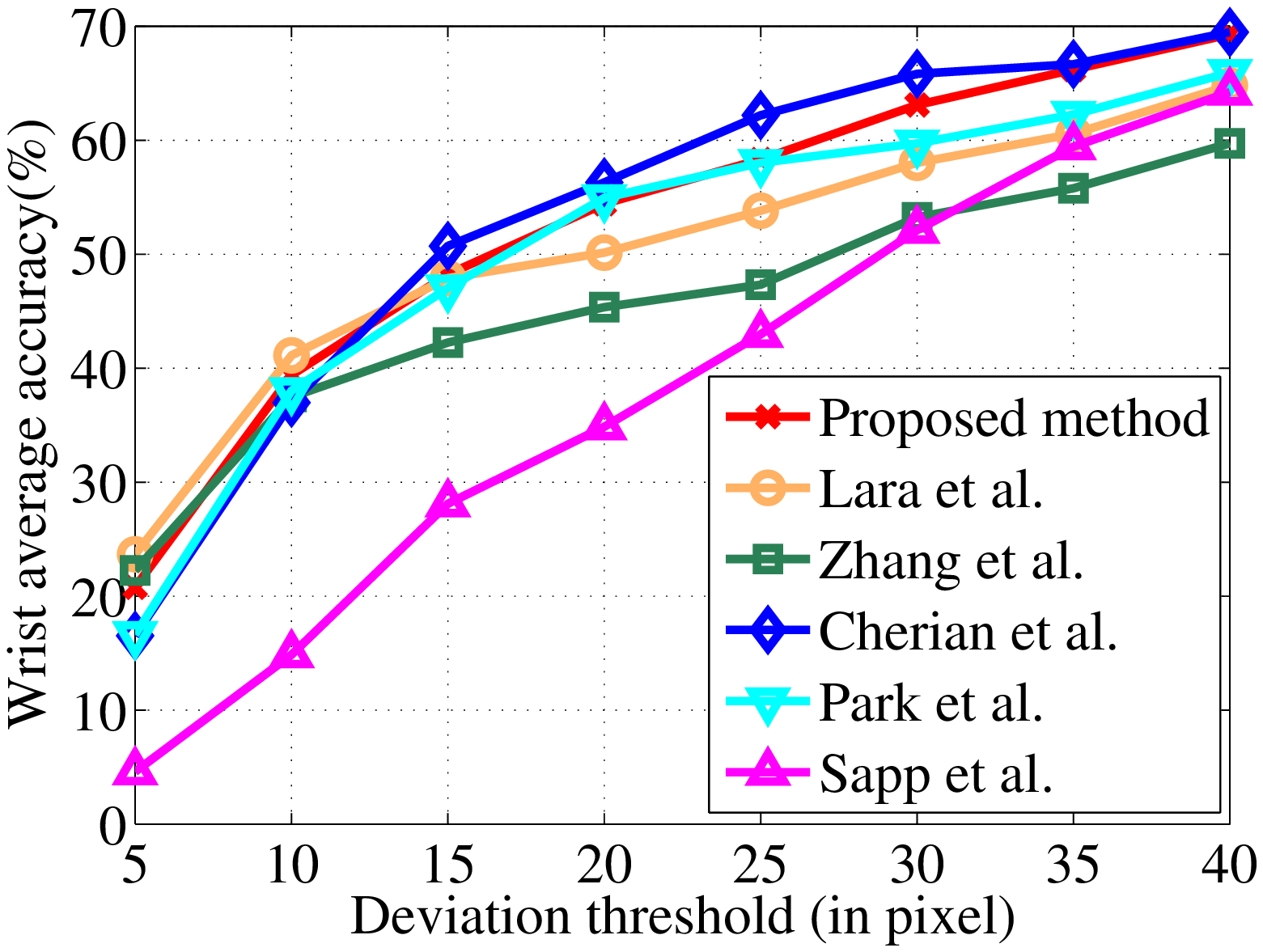} \\
	(a)	&	(b)  &   (c)
    \end{array} $
    \caption{Comparison results of different parts with the state-of-the-art methods on Poses in The Wild dataset: (a) shoulder, (b) elbow, and (c) wrist.}
    \label{fig:chap5:wildpose_state_of_the_art_comparison}
  \end{center}
\end{figure}
{\bf Experimental results on Poses in the Wild dataset:} Poses in the Wild dataset has no training and test data partition. We use first
15 video clips to train our system and the remaining 15 video clips to test. Note that we have followed the same data partition for all the methods.
Fig.~\ref{fig:chap5:wildpose_state_of_the_art_comparison}
shows the result of different methods on three individual body parts for comparison, and Table~\ref{tab:chap5:wildpose_all_part_satate_of_the_art_comparison} shows average accuracy (\%) for all three parts.
Fig.~\ref{fig:chap5:wildpose_state_of_the_art_comparison}(a) shows that our method gives better result
for almost all the values of deviation threshold as the movement of this part is small. As the movement increases, performance of proposed method reduces, but it still remains within top two methods and far better than the individual object tracking methods~\cite{LauraSevilla-LaraCVPR12, KaihuaZhangECCV14} as shown in Fig.~\ref{fig:chap5:wildpose_state_of_the_art_comparison}(b) and (c) and Table~\ref{tab:chap5:wildpose_all_part_satate_of_the_art_comparison}. Note that time complexity of the closest competitors~\cite{DennisParkICCV11, AnoopCherianCVPR14} is much higher than the proposed method (see Subsection~\ref{subsec:Computational_complexity}).
\noindent
\begin{table*}[t]
\small
\caption{Average accuracy (in \%) of pose tracking comprising three parts (shoulder, elbow, and wrist) together using different methods on Poses in the Wild dataset.}
\begin{center}
\begin{tabular}{|c|c|c|c|c|c|c|}
\hline
    Dev. thrs.			&   \multicolumn{6}{c|}{Different methods}\\\cline{2-7}
    (in pixel)                        &   Lara~\cite{LauraSevilla-LaraCVPR12}  &  Zhang~\cite{KaihuaZhangECCV14}    &   Sapp~\cite{BenjaminSappCVPR11}  &     Park~\cite{DennisParkICCV11}    &  Cherian~\cite{AnoopCherianCVPR14}  &   \textbf{Proposed} \\ \hline
    5                       &   \bf{24.82}      &   23.19       &   7.63        &   18.08       &   17.27       &   23.50                    \\ \hline
    10                      &   43.91       &   43.94       &   19.48       &   43.32       &   42.72       &   \bf{44.02}                     \\ \hline
    15                      &   53.78       &   53.51       &   31.27       &   58.60       &   \bf{59.59}      &   59.57                       \\ \hline
    20                      &   57.93       &   58.31       &   39.93       &   68.67       &   \bf{69.93}      &   68.21                       \\ \hline
    25                      &   62.04       &   61.47       &   48.11       &   71.94       &   \bf{74.56}      &   72.55                       \\ \hline
    30                      &   67.28       &   65.06       &   57.93       &   74.56       &   \bf{77.82}      &   75.42                       \\ \hline
    35                      &   70.99       &   67.43       &   65.17       &   77.19       &   \bf{80.03}      &   77.67                       \\  \hline
    40                      &   74.19       &   69.82       &   69.31       &   79.29       &   \bf{81.67}      &   79.59                       \\
\hline
\end{tabular}
\end{center}
\label{tab:chap5:wildpose_all_part_satate_of_the_art_comparison} 
\end{table*}

\begin{figure}
  \begin{center}
    $\begin{array}{@{\hspace{1pt}}c@{\hspace{1pt}}c@{\hspace{1pt}}c}
	\includegraphics[width = 0.3\textwidth]{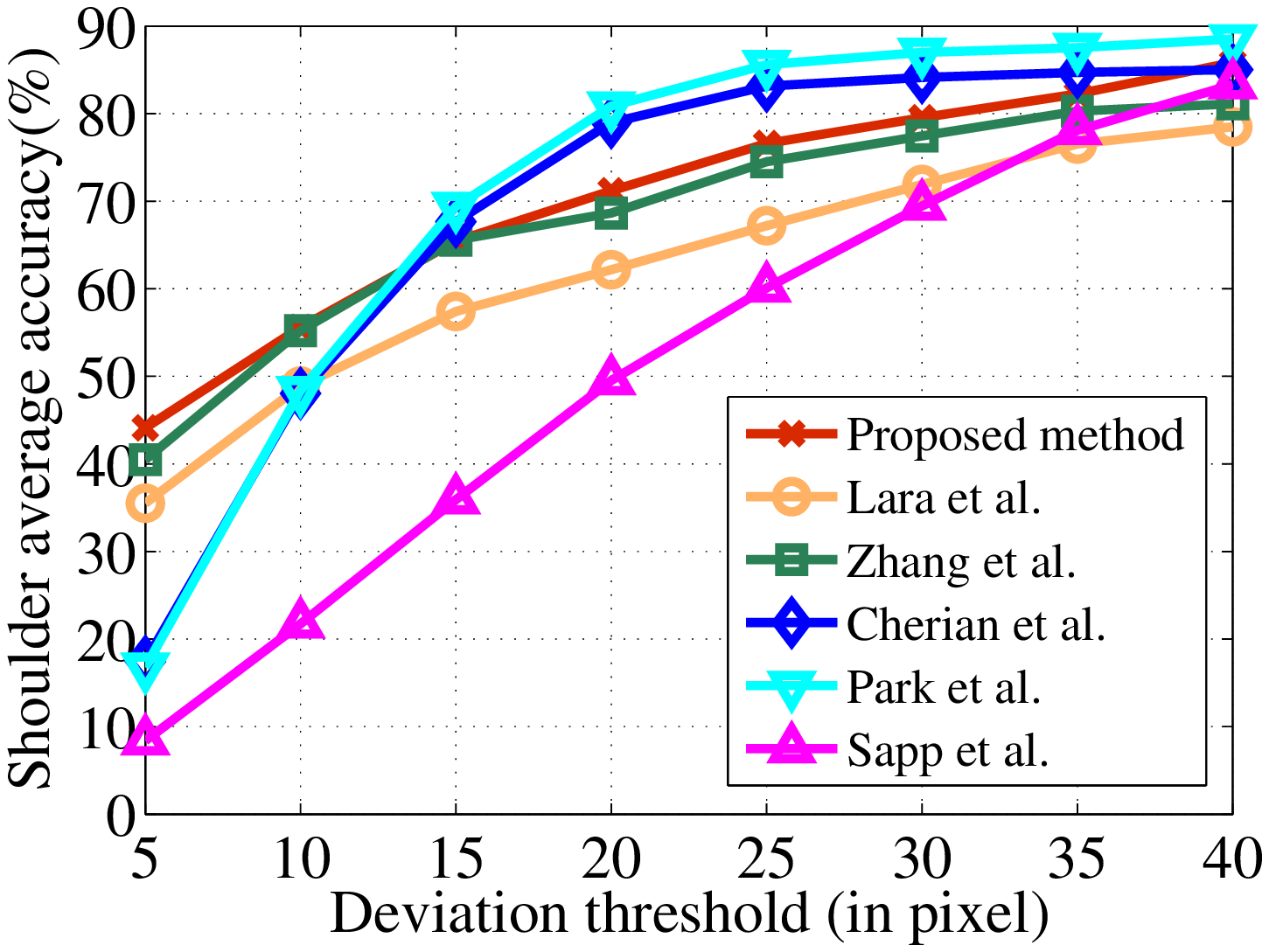} &
    \includegraphics[width = 0.3\textwidth]{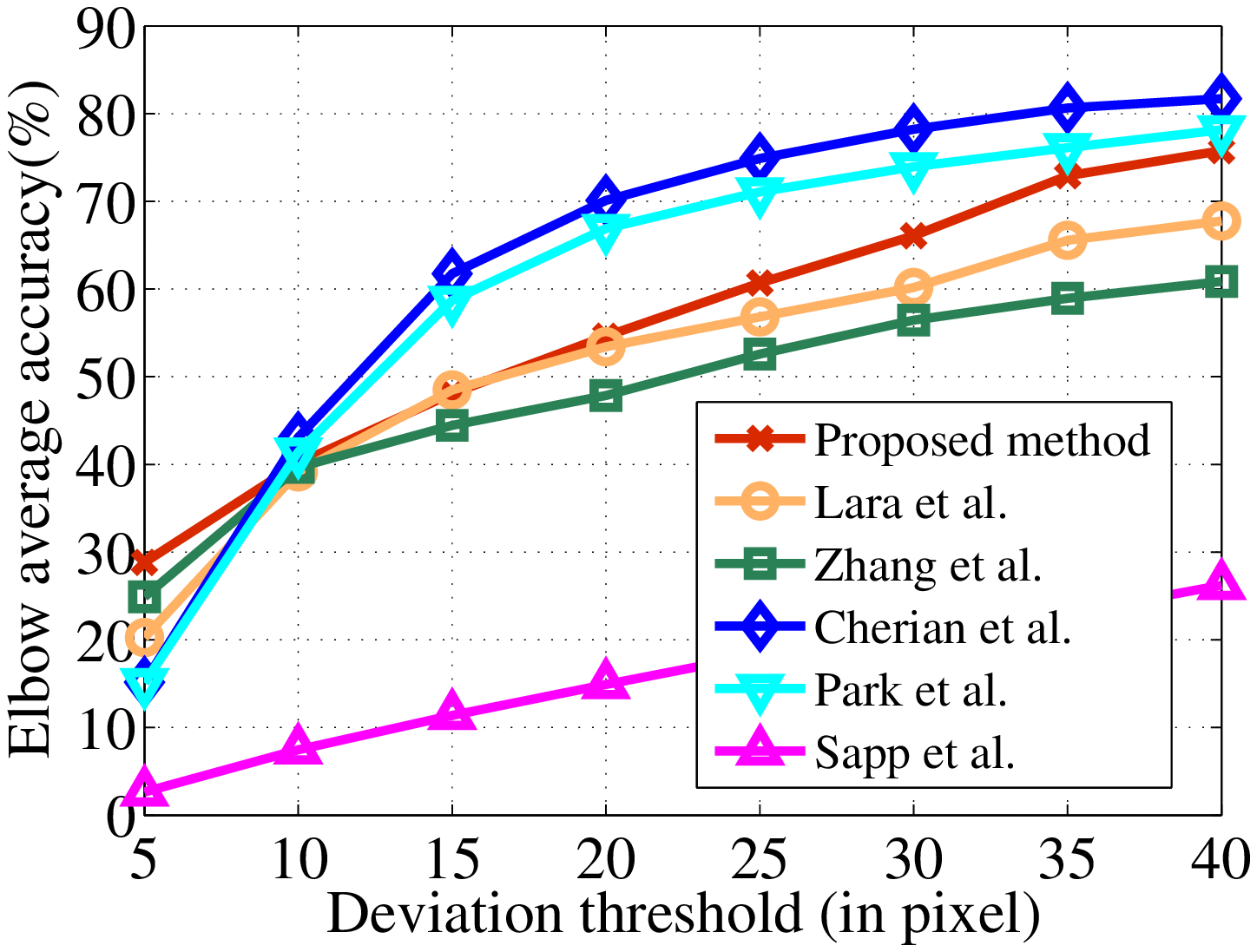} &
	\includegraphics[width = 0.3\textwidth]{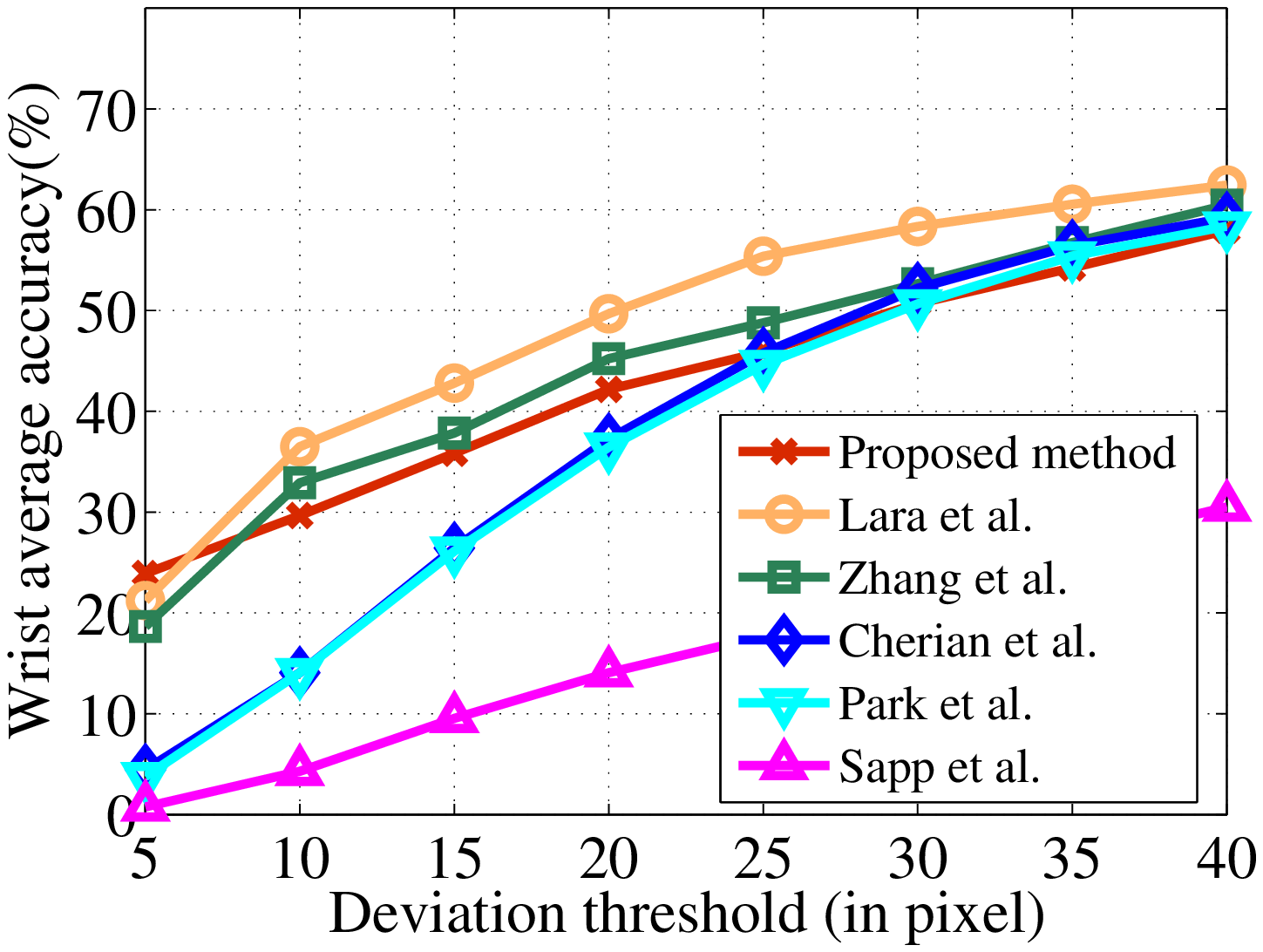} \\
	(a)	&	(b)  &   (c) \\
    \includegraphics[width = 0.3\textwidth]{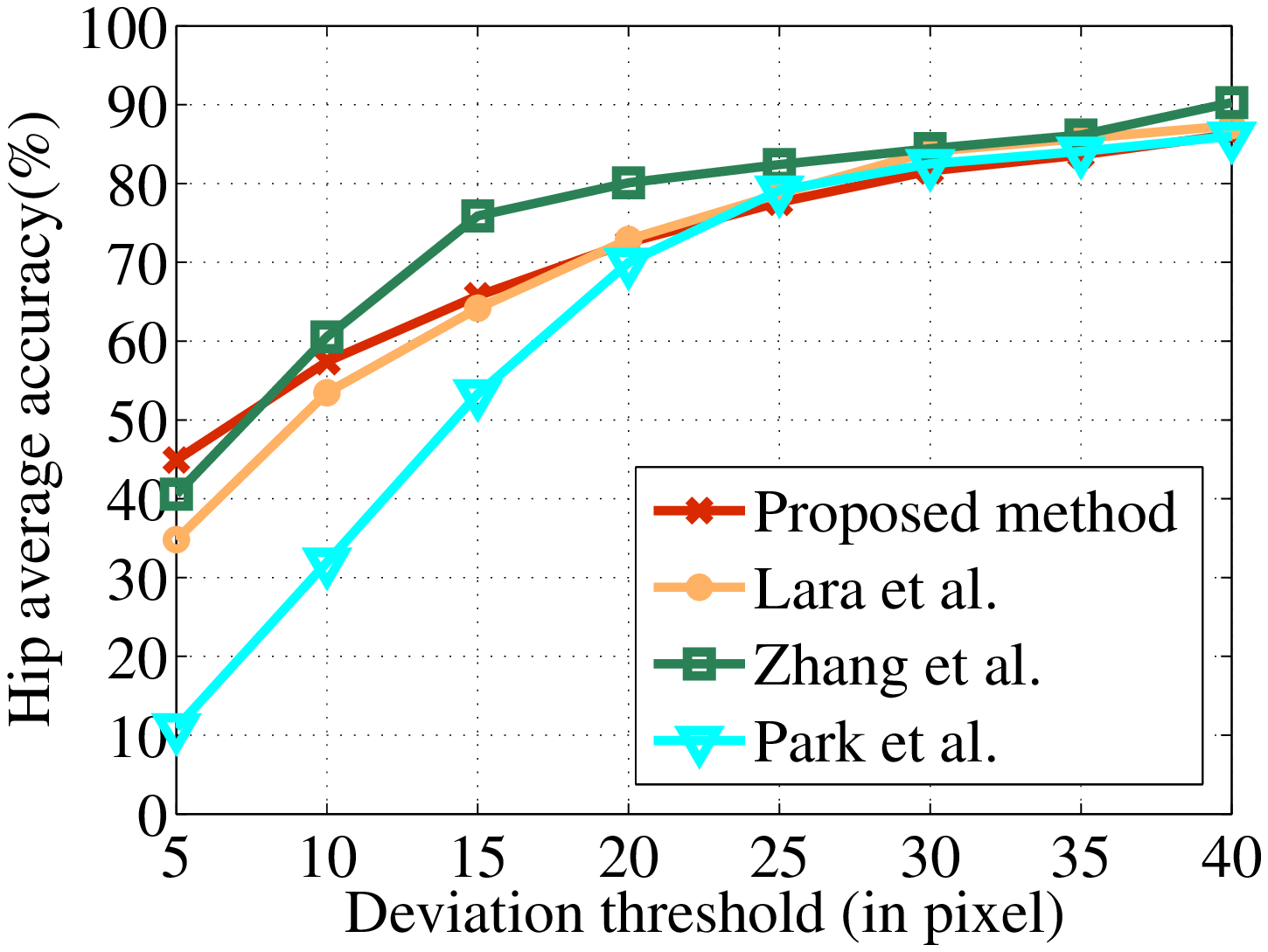} &
    \includegraphics[width = 0.3\textwidth]{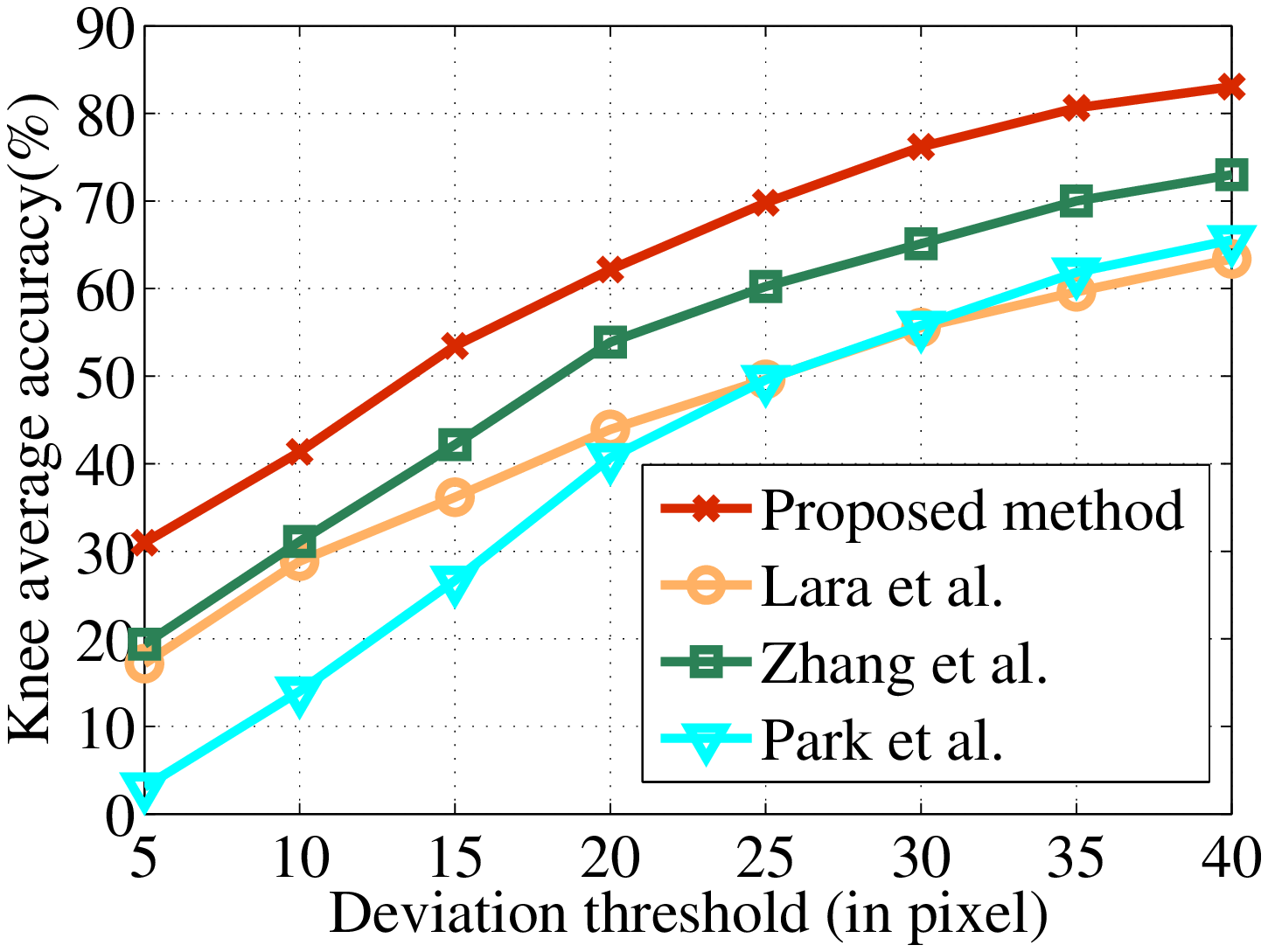} &
	\includegraphics[width = 0.3\textwidth]{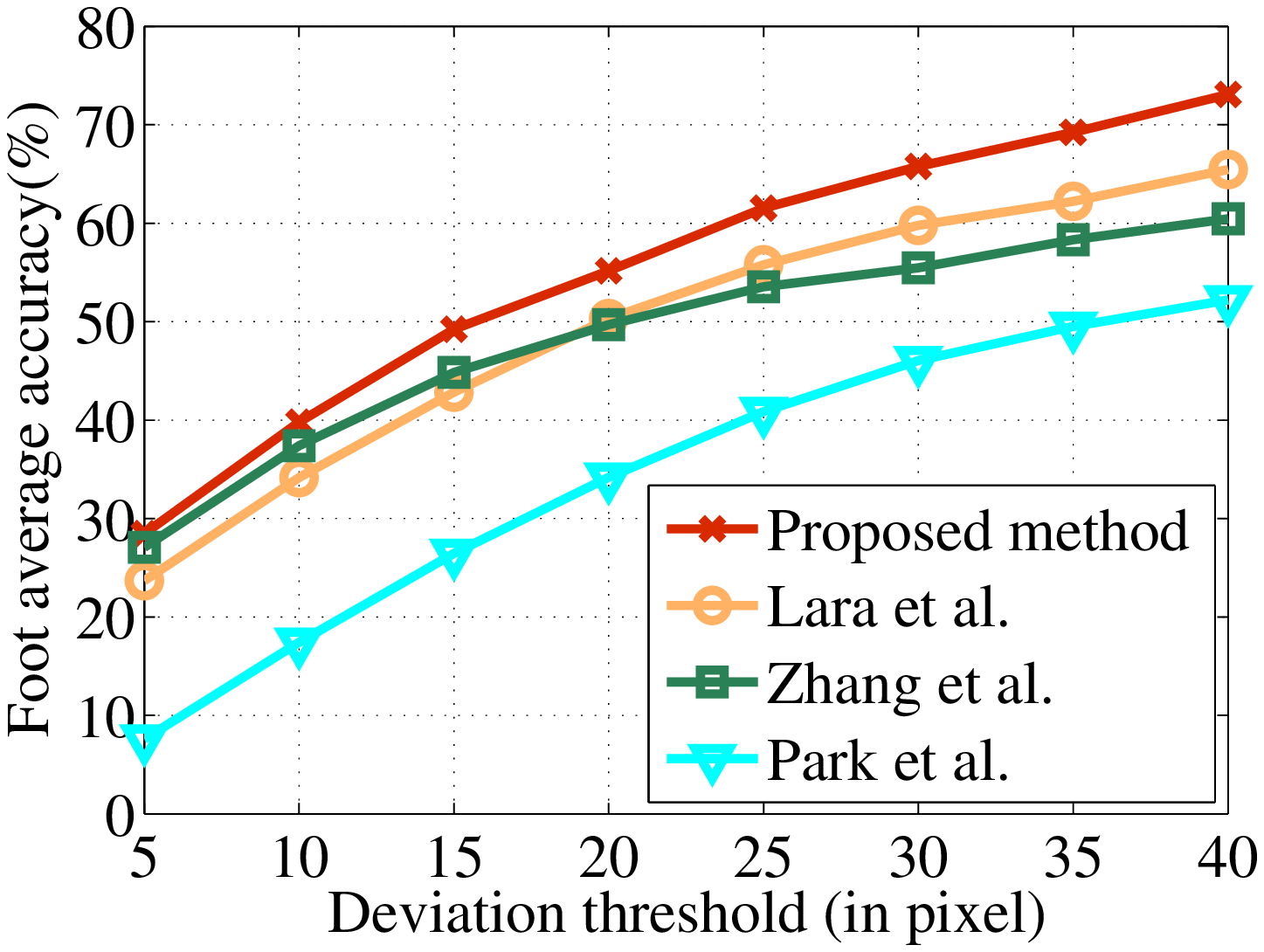} \\
	(d)	&	(e)  &   (f)
    \end{array} $
    \caption{Comparison results of different body parts with the state-of-the-art methods on ICDPose dataset: (a) shoulder, (b) elbow, (c) wrist, (d) hip, (e) knee, and (f) foot.}
    \label{fig:chap5:icdpose_state_of_the_art_comparison}
  \end{center}
\end{figure}
{\bf Experimental results on ICDPose dataset: } We have tested the proposed method on 30 video clips of ICDPose dataset using remaining 30 
video clips as training data. Comparison of our results with the state-of-the-art methods are reported in Fig.~\ref{fig:chap5:icdpose_state_of_the_art_comparison} 
for different body parts. Figs.~\ref{fig:chap5:icdpose_state_of_the_art_comparison}(e) and (f) show that the proposed method achieves highest scores for knee 
and foot, and for other parts it consistently remains among the top performers. However, for wrist part the proposed method falls marginally behind most of 
other methods as seen in Fig.~\ref{fig:chap5:icdpose_state_of_the_art_comparison}(c). If we consider average accuracy computed over all body parts, the 
proposed method stands superior. However, for only upper portion of the body it stands third position (shown in 
Table~\ref{tab:chap5:icdpose_all_part_satate_of_the_art_comparison}). 
\noindent
\begin{table*}
\small
\caption{Average accuracy (in \%) of pose tracking comprising upper body parts (without bracket) and full body parts (within the bracket) together using different methods on ICDPose dataset.}
\begin{center}
\begin{tabular}{|c|c|c|c|c|c|c|}
\hline
    Dev. thrs.			&   \multicolumn{6}{c|}{Different methods}\\\cline{2-7}
        (in pixel)                        &   Lara~\cite{LauraSevilla-LaraCVPR12}  &  Zhang~\cite{KaihuaZhangECCV14}    &   Sapp~\cite{BenjaminSappCVPR11}  &     Park~\cite{DennisParkICCV11}    &  Cherian~\cite{AnoopCherianCVPR14}  &   \textbf{Proposed} \\ \hline

    5                        &   25.67 (25.44)       &   28.01 (28.50)       &   03.93       &   11.89 (09.54)       &   12.35       &   \bf{32.28} (\bf{33.53})               \\ \hline
    10                      &   41.50 (40.16)       &   \bf{42.61} (42.81)       &   11.16       &   34.62 (27.88)       &   35.07       &   41.79 (\bf{43.98})               \\ \hline
    15                      &   49.56 (48.64)       &   49.29 (51.80)       &   18.95       &   51.42 (43.44)       &   \bf{51.96}       &   49.80 (\bf{52.97})               \\ \hline
    20                      &   55.10 (55.40)       &   53.91 (57.56)       &   26.17       &   61.41 (54.84)       &   \bf{62.12}       &   56.00 (\bf{59.63})               \\ \hline
    25                      &   59.80 (60.57)       &   58.63 (62.00)       &   32.03       &   67.11 (61.81)       &   \bf{67.94}       &   61.06 (\bf{65.35})               \\ \hline
    30                      &   63.47 (64.96)       &   62.18 (65.25)       &   37.46       &   70.53 (65.98)       &   \bf{71.52}       &   65.44 (\bf{69.97})               \\ \hline
    35                      &   67.54 (68.36)       &   65.33 (68.40)       &   42.63       &   73.03 (69.08)       &   \bf{73.93}       &   69.80 (\bf{73.80})               \\ \hline
    40                      &   69.57 (70.81)       &   67.47 (71.02)       &   46.68       &   75.02 (71.47)       &   \bf{75.31}       &   73.20 (\bf{76.99})               \\
\hline
\end{tabular}
\end{center}
\label{tab:chap5:icdpose_all_part_satate_of_the_art_comparison} 
\end{table*}

\begin{figure}
  \begin{center}
    $\begin{array}{@{\hspace{1pt}}c@{\hspace{1pt}}c@{\hspace{1pt}}c}
	\includegraphics[width = 0.3\textwidth]{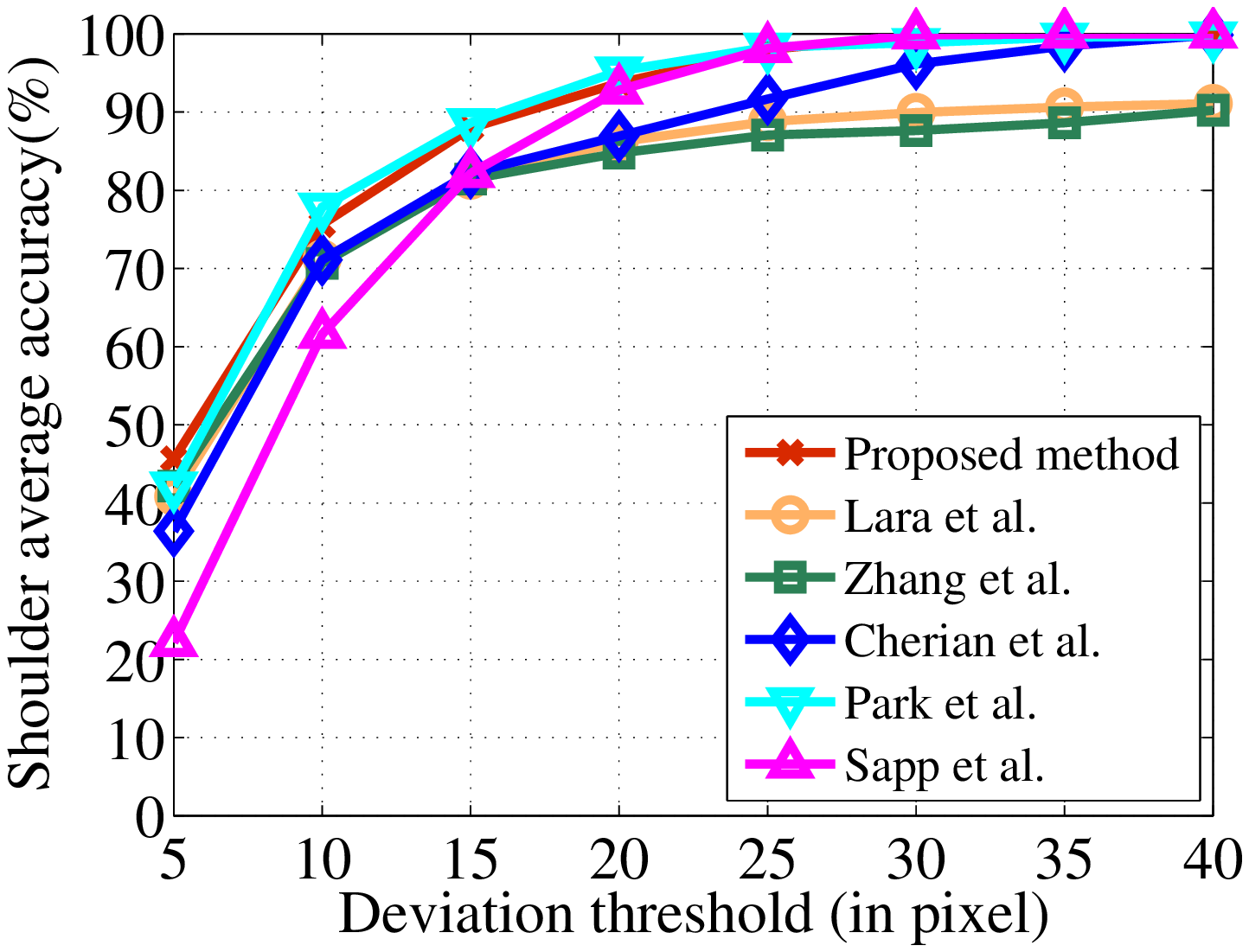} &
    \includegraphics[width = 0.3\textwidth]{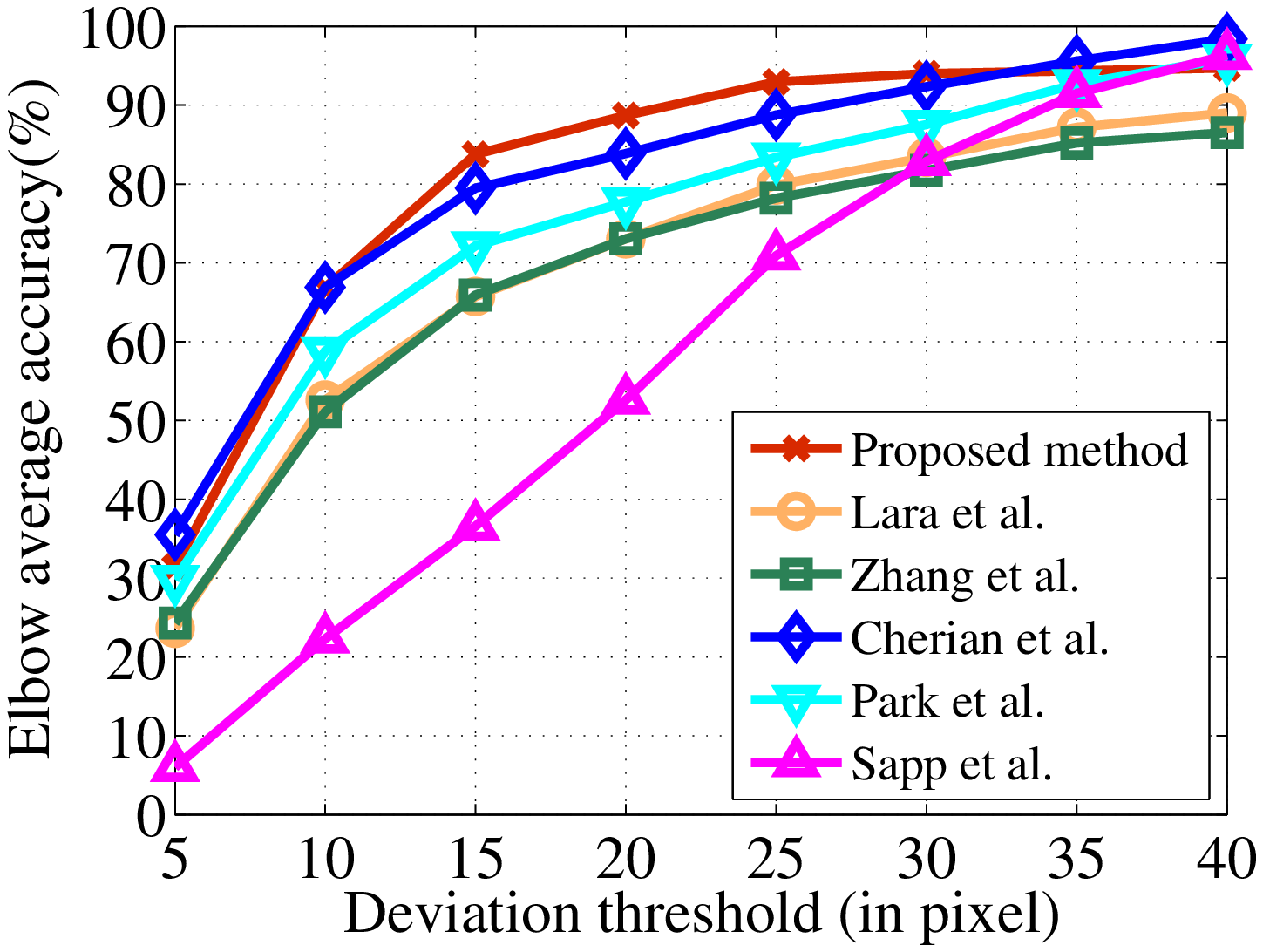} &
	\includegraphics[width = 0.3\textwidth]{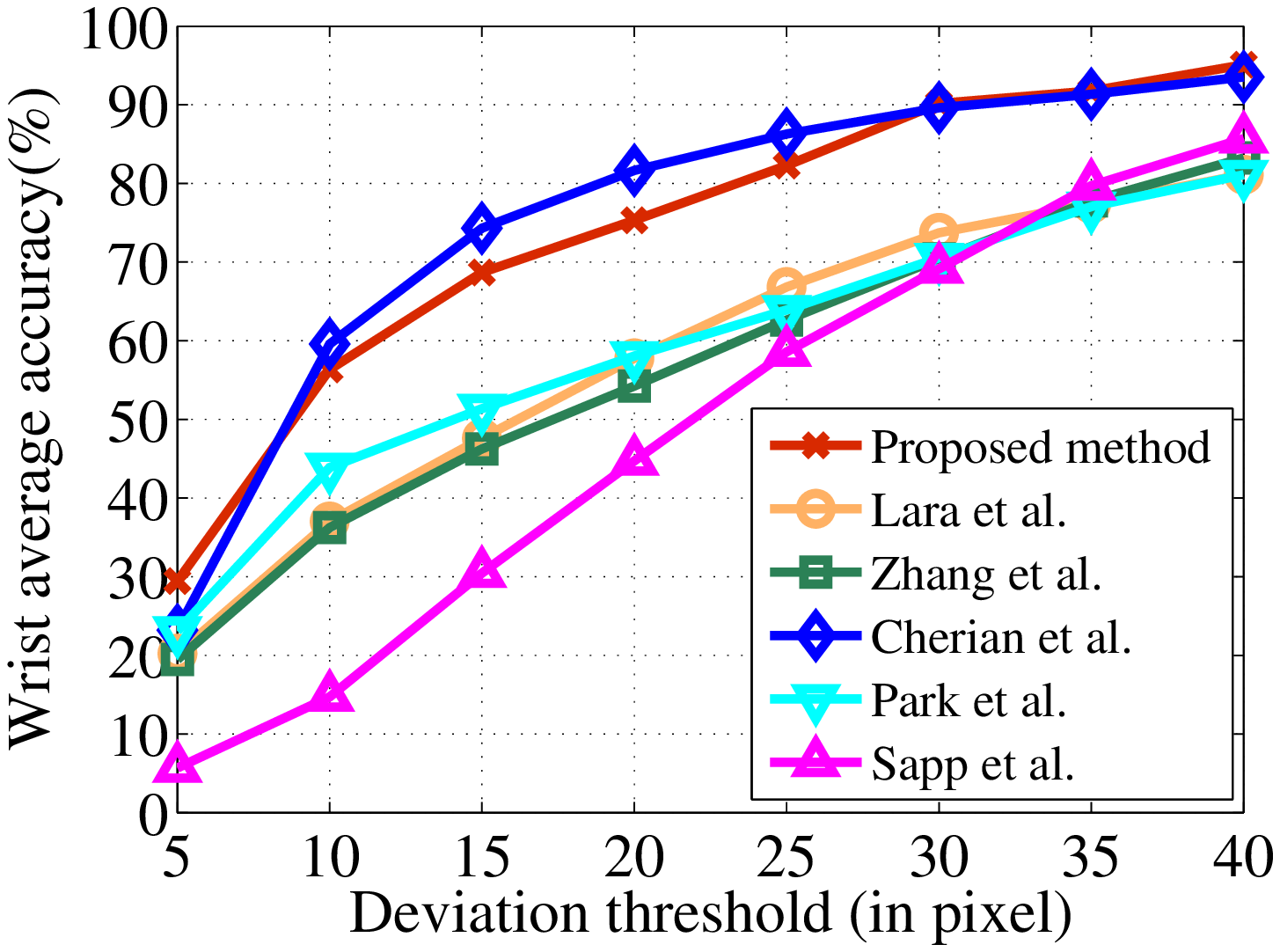} \\
	(a)	&	(b)  &   (c) \\
    \includegraphics[width = 0.3\textwidth]{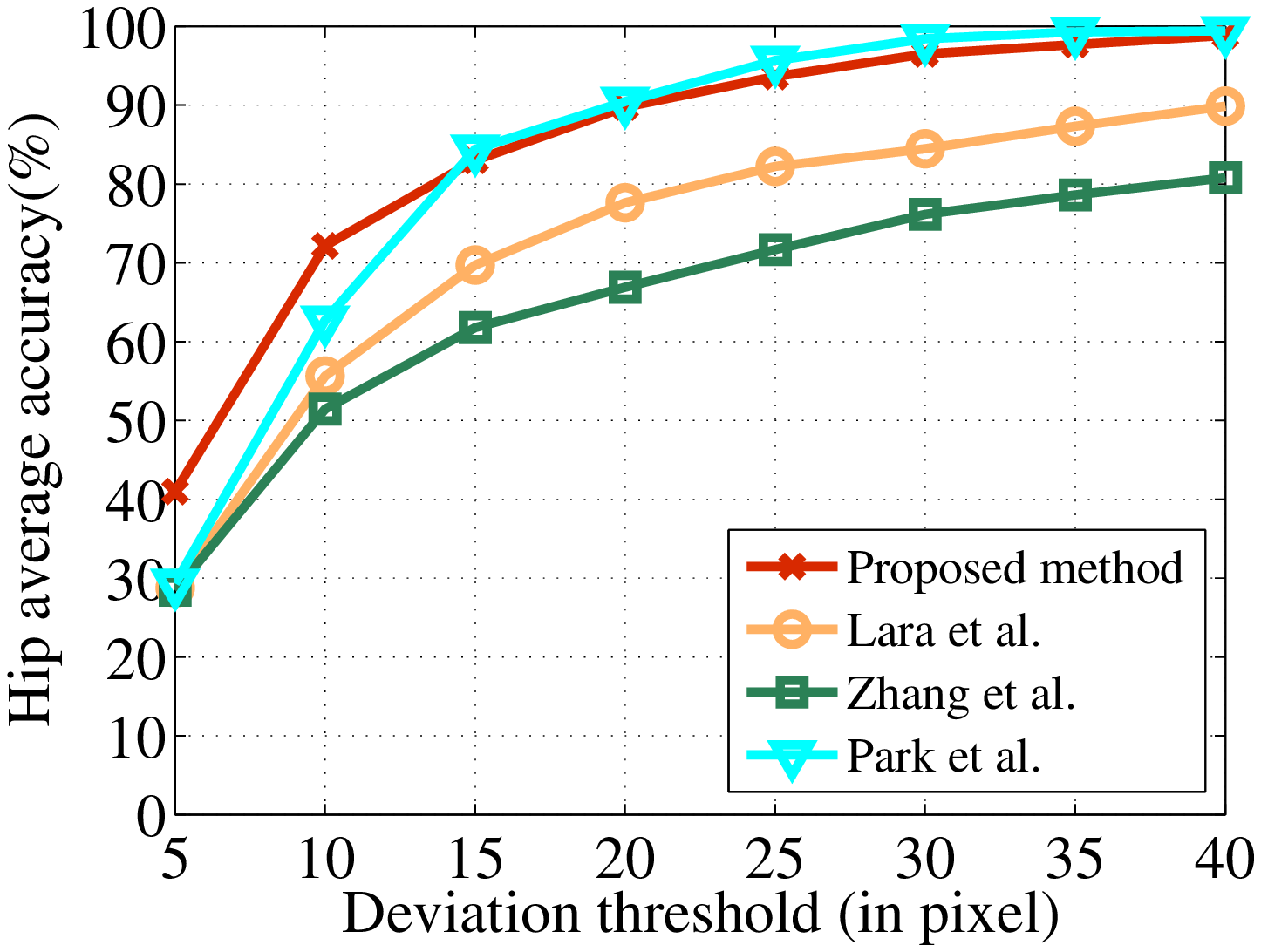} &
    \includegraphics[width = 0.3\textwidth]{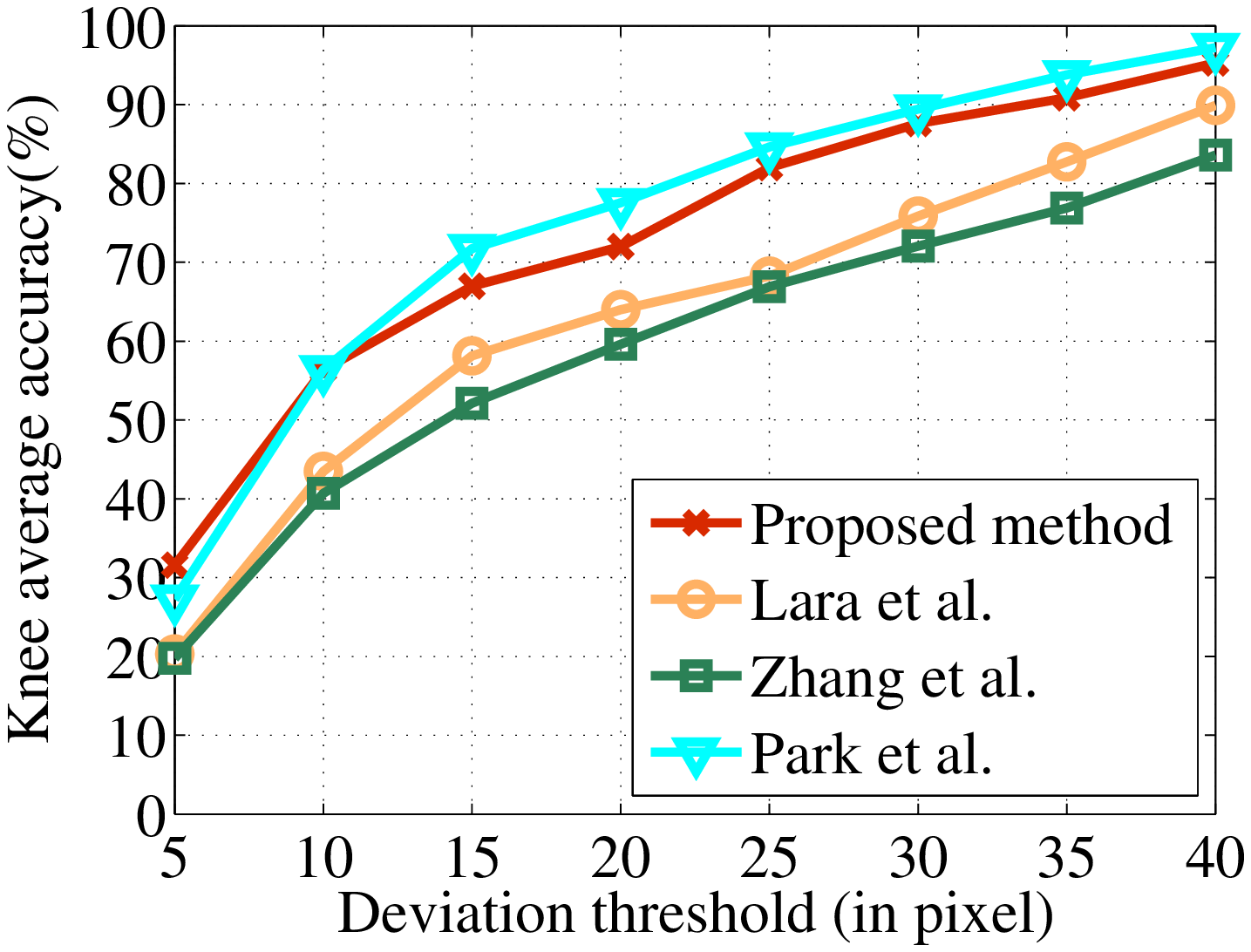} &
	\includegraphics[width = 0.3\textwidth]{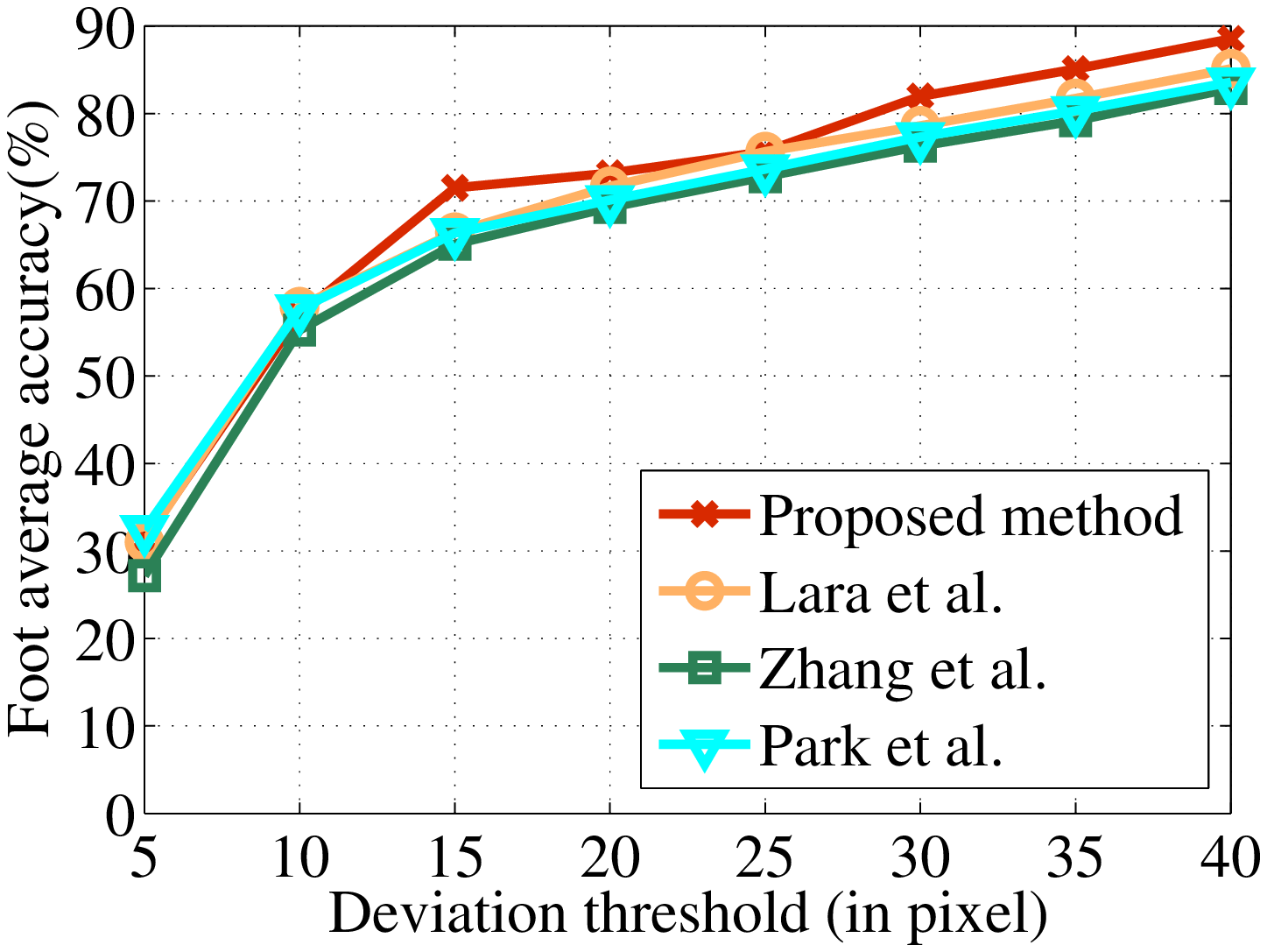} \\
	(d)	&	(e)  &   (f)
    \end{array} $
    \caption{Comparison results of different body parts with the state-of-the-art methods on Outdoor Pose dataset: (a) shoulder, (b) elbow, (c) wrist, (d) hip, (e) knee, and (f) foot.}
    \label{fig:chap5:outdoor_state_of_the_art_comparison}
  \end{center}
\end{figure}

{\bf Experimental results on Outdoor Pose dataset:} Outdoor Pose dataset has also no training and test data partition. As it is a full body pose dataset, we obtain parameter values from ICDPose training dataset and use all the six videos as test data. This dataset has on an average $138$ frames per videos, which is large compared to other datasets like VideoPose2, Poses in the Wild and ICDPose. So we initialize our tracking method after every $60$ frames. Fig.~\ref{fig:chap5:outdoor_state_of_the_art_comparison} shows the results of different body parts (joints) tracking using different methods for comparison.
From this figure we see that performance of the proposed method is at least second best for individual parts, and is best considering average accuracy over all the parts as shown in  Table~\ref{tab:chap5:outdoor_all_part_state_of_the_art_comparison}.
Ramakrishna et al.~\cite{VarunRamakrishnaCVPR13} have reported their result using \textit{Percentage of Correct Parts} (PCP)~\cite{VittorioFerrariCVPR08} evaluation metric. So in Table~\ref{tab:chap5:outdoor_all_part_pcp_state_of_the_art_comparison} we present PCP score of our method for comparison with \cite{VarunRamakrishnaCVPR13}, which reveals that on an average our method is superior.

\noindent
\begin{table*}
\small
\caption{Average accuracy (in \%) of pose tracking comprising six parts (shoulder, elbow, wrist, hip, knee, and foot) together using different methods on Outdoor Pose dataset.}
\begin{center}
\begin{tabular}{|c|c|c|c|c|c|c|}
\hline
    Dev. thrs.			&   \multicolumn{6}{c|}{Different methods}\\\cline{2-7}
        (in pixel)                        &   Lara~\cite{LauraSevilla-LaraCVPR12}  &  Zhang~\cite{KaihuaZhangECCV14}    &   Sapp~\cite{BenjaminSappCVPR11}  &     Park~\cite{DennisParkICCV11}    &  Cherian~\cite{AnoopCherianCVPR14}  &   \textbf{Proposed} \\ \hline

    5                        &   28.23 (27.46)       &   28.64 (26.89)       &   11.28       &   31.68 (30.68)       &   31.70       &   \bf{35.50} (\bf{35.05})               \\ \hline
    10                      &   53.61 (52.98)       &   52.66 (50.90)       &   32.87       &   60.17 (59.54)       &   65.85       &   \bf{66.25} (\bf{64.04})               \\ \hline
    15                      &   64.88 (64.82)       &   64.52 (62.09)       &   49.74       &   70.72 (72.44)       &   78.67       &   \bf{80.13} (\bf{76.97})               \\ \hline
    20                      &   72.33 (71.71)       &   70.68 (67.97)       &   63.37       &   77.01 (78.19)       &   84.15       &   \bf{85.90} (\bf{82.10})               \\ \hline
    25                      &   78.49 (76.93)       &   75.96 (73.21)       &   75.87       &   81.85 (83.25)       &   88.90       &   \bf{91.06} (\bf{87.41})               \\ \hline
    30                      &   82.34 (80.99)       &   79.91 (77.37)       &   83.95       &   85.64 (87.02)       &   92.70       &   \bf{94.63} (\bf{91.66})               \\ \hline
    35                      &   85.08 (84.50)       &   83.87 (81.04)       &   90.41       &   89.77 (90.44)       &   95.13       &   \bf{95.37} (\bf{93.29})               \\ \hline
    40                      &   87.07 (87.69)       &   86.67 (84.54)       &   94.02       &   92.21 (92.81)       &   \bf{97.26}       &   96.60 (\bf{95.41})               \\
\hline
\end{tabular}
\end{center}
\label{tab:chap5:outdoor_all_part_state_of_the_art_comparison} 
\end{table*}
\noindent
\begin{table}
\small
\caption{Comparison with~\cite{VarunRamakrishnaCVPR13} using percentage of correct parts (PCP)
on Outdoor Pose dataset.}
\begin{center}
\begin{tabular}{|c|c|c|}
\hline
    Different body parts        &   Ramakrishna~\cite{VarunRamakrishnaCVPR13}     &      \textbf{Proposed} \\ \hline
    upper arm                       &   0.86        &    \bf{0.94}    \\ \hline
    lower arm                       &   0.52        &    \bf{0.81}    \\ \hline
    upper leg                        &   \bf{0.95}        &    0.94    \\ \hline
    lower leg                        &   \bf{0.96}        &    0.92     \\ \hline
    Average                         &   0.82                &   \bf{0.90}   \\
\hline
\end{tabular}
\end{center}
\label{tab:chap5:outdoor_all_part_pcp_state_of_the_art_comparison} 
\end{table}

To show the strength of the spatial constraints we run our experiment on four datasets (VideoPose2, Poses in the Wild, Outdoor Pose and ICDPose) without spatial deformation (WOSD) constraints (or $\lambda_{2} = 0$ in Eq.~(\ref{eqn:chap5:regularized_x_i_t_*})). Table~\ref{tab:chap5:icdpose_all_part_satate_of_the_art_comparison} shows the average accuracy (\%) of WOSD constraint and with spatial deformation (WSD) constraint for different values of $\Omega$. We have done a computational complexity analysis and time comparison with the other methods in the next Subsection~\ref{subsec:Computational_complexity}.
\noindent
\begin{table*}
\small
\caption{Comparative results between without spatial deformation (WOSD) and with spatial deformation (WSD)  constraint on different datasets ( in average accuracy of all the parts)}
\begin{center}
\begin{tabular}{|c|c|c|c|c|c|c|c|c|}
\hline
    Dev. thrs.			&   \multicolumn{2}{c|}{VideoPose2~\cite{BenjaminSappCVPR11}}     &   \multicolumn{2}{c|}{Poses in the Wild~\cite{AnoopCherianCVPR14}}  &   \multicolumn{2}{c|}{Outdoor Pose~\cite{VarunRamakrishnaCVPR13}}  &   \multicolumn{2}{c|}{ICDPose}\\\cline{2-9}
        (in pixel)                        &   WOSD  &  WSD    &   WOSD  &  WSD    &  WOSD  &  WSD   &  WOSD  &  WSD \\ \hline

    5                        &   41.08       &   41.22       &   22.20       &   23.50       &  30.87       &   35.05       &   33.20       &  33.53               \\ \hline
    10                      &   58.60       &   61.13       &   41.94       &   44.02       &   61.60       &   64.04       &   43.36       &   43.98               \\ \hline
    15                      &   65.90       &   69.84       &   54.60       &   59.58       &   74.45       &   76.97       &   51.80       &   52.97               \\ \hline
    20                      &   77.20       &   78.21       &   62.43       &   68.21       &   80.81       &   82.10       &   58.73       &   59.64               \\ \hline
    25                      &   77.47       &   82.03       &   66.51       &   72.55       &   86.16       &   87.41       &   64.70       &   65.36              \\ \hline
    30                      &   83.17       &   84.74       &   70.80       &   75.42       &   90.41       &   91.66       &   68.73       &   69.97               \\ \hline
    35                      &   86.94       &   88.04       &   73.81       &   77.67       &   92.78       &   93.29       &   71.90       &   73.81              \\ \hline
    40                      &   87.82       &   89.90       &   76.68       &   79.59       &   94.69       &   95.41       &   75.25       &   76.99               \\
\hline
\end{tabular}
\end{center}
\label{tab:chap5:icdpose_all_part_satate_of_the_art_comparison} 
\end{table*}

\subsection{Computational complexity}
\label{subsec:Computational_complexity}

In Subsection~\ref{subsec:tracking_human_pose_in_a_video} we have mentioned that the order of time complexity of our proposed method is $O(nMN)$, where $n$ is the number of body parts under consideration, $M$ the plausible locations of each part and $N$ is the number of clusters for each of these locations due to relative spatial displacement. On the other hand, the time complexity of pose estimation method given in~\cite{DennisParkICCV11} is $O(nMN^{2})$ and that of Cherian et al.~\cite{AnoopCherianCVPR14} is even higher. So the proposed method is at least $N$ times faster than the state-of-the-art pose estimation methods presented in~\cite{DennisParkICCV11} and~\cite{AnoopCherianCVPR14}.

We have already compared the average time required by feature computation methods in Subsection~\ref{subsec:Experimental_evaluation_and_discussion} (Table~\ref{tab:chap5:feature_time_comparison}).
We have also compared the execution time of the proposed tracking method with that of the state-of-the-art individual object tracking methods~\cite{LauraSevilla-LaraCVPR12, KaihuaZhangECCV14}
in Table~\ref{tab:chap5:tracking_time_comparison}, which shows that our method is slower than that of Zhang et al.~\cite{KaihuaZhangECCV14}. However, the proposed method gives much higher accuracy compared to Zhang et al.'s method (see Tables~\ref{tab:chap5:videopose_all_part_state_of_the_art_comparison},~\ref{tab:chap5:wildpose_all_part_satate_of_the_art_comparison},~\ref{tab:chap5:icdpose_all_part_satate_of_the_art_comparison}, and~\ref{tab:chap5:outdoor_all_part_state_of_the_art_comparison}).
\noindent
\begin{table}
\small
\caption{Average time comparison (in sec./frame) of proposed method with the state-of-the-art individual tracking methods on different datasets.}
\begin{center}
\begin{tabular}{|c|c|c|c|}
\hline
  Datasets				&	Lara~\cite{LauraSevilla-LaraCVPR12}	&	Zhang~\cite{KaihuaZhangECCV14}	&	Proposed \\ \hline
  VideoPose2~\cite{BenjaminSappCVPR11}		&	0.6293					&	\bf{0.0411}				&	0.2268		\\ \hline
  Poses in the Wild~\cite{AnoopCherianCVPR14}	&	0.6265					&	\bf{0.0518}				&	0.2256		\\ \hline
  ICDPose~\cite{ICDPose}		&	1.1666					&	\bf{0.2375}				&	0.4638		\\ \hline
  Outdoor Pose~\cite{VarunRamakrishnaCVPR13}    &   1.1720      &   \bf{0.2033}      &   0.3904\\
\hline
\end{tabular}
\end{center}
\label{tab:chap5:tracking_time_comparison} 
\end{table}

\section{Conclusion}
\label{sec:Conclusion}

In this paper we have presented a human pose tracking method by introducing a novel body part descriptor.
We have considered human pose as a graphical tree structure model and formulated the human pose tracking problem as a
discrete optimization problem by combing the following terms: likeliness of appearance of a part within a frame, temporal
displacement of the part from previous frame to the current frame, and spatial dependency of a part with its parent in the graph structure.
The first and third terms take care of pose estimation in single frame or image, while the second term deals with object tracking in subsequent
frames. More precisely the first term measures the degree of the presence of a body part at a location and the third term maintains the
global structure of the human body. Thus the proposed method becomes robust by incorporating advantages of both approaches. We have proposed a greedy
approach to solve the optimization problem and consequently to track the human pose efficiently. Experimental
results on benchmark datasets (VideoPose2, Poses in the Wild and Outdoor Pose) as well as on our newly developed full human body pose dataset,
called ICDPose, show the efficacy of the proposed method.

\bibliographystyle{plain}
\bibliography{ss_all_references_abbreviation}

\begin{thebibliography}{10}

\bibitem{SIFTflow}
http://people.csail.mit.edu/celiu/siftflow/.

\bibitem{Seung-HwanBaeIEEETIP14}
S.-H. Bae and K.-J. Yoon.
\newblock Robust online multiobject tracking with data association and track
  management.
\newblock {\em IEEE Trans. on Image Processing}, 23(7):2820--2833, 2014.

\bibitem{LambertoBallanIEEETM12}
L.~Ballan, M.~Bertini, A.~D. Bimbo, L.~Seidenari, and G.~Serra.
\newblock Effective codebooks for human action representation and
  classification in unconstrained videos.
\newblock {\em IEEE Trans. on Multimedia}, 14(4):1234--1245, 2012.

\bibitem{JanBandouchIJCV12}
J.~Bandouch, O.~C. Jenkins, and M.~Beetz.
\newblock A self-training approach for visual tracking and recognition of
  complex human activity patterns.
\newblock {\em IJCV}, 99(2):166--189, 2012.

\bibitem{MarkBarnardIEEETM14}
M.~Barnard, P.~Koniusz, W.~Wang, J.~Kittler, S.~M. Naqvi, and J.~Chambers.
\newblock Robust multi-speaker tracking via dictionary learning and identity
  modeling.
\newblock {\em IEEE Trans. on Multimedia}, 16(3):864--880, 2014.

\bibitem{HerbertBayCVIU08}
H.~Bay, A.~Ess, T.~Tuytelaars, and L.~V. Gool.
\newblock Surf: Speeded up robust features.
\newblock {\em CVIU}, 110(3):346--359, 2008.

\bibitem{TewodrosABiresawNeurocomputing15}
T.~A. Biresaw, A.~Cavallaro, and C.~S. Regazzoni.
\newblock Correlation-based self-correcting tracking.
\newblock {\em Neurocomputing}, 152(3):345--358, 2015.

\bibitem{ZhaoweiCaiIEEETIP14}
Z.~Cai, L.~Wen, Z.~Lei, N.~Vasconcelos, and S.~Z. Li.
\newblock Robust deformable and occluded object tracking with dynamic graph.
\newblock {\em IEEE Trans. on Image Processing}, 23(12):5497--5509, 2014.

\bibitem{AnoopCherianCVPR14}
A.~Cherian, J.~Mairal, K.~Alahari, and C.~Schmid.
\newblock Mixing body-part sequences for human pose estimation.
\newblock In {\em CVPR}, June 2014.

\bibitem{ChunTeChuIEEETM13}
C.-T. Chu, J.-N. Hwang, H.-I. Pai, and K.-M. Lan.
\newblock Tracking human under occlusion based on adaptive multiple kernels
  with projected gradients.
\newblock {\em IEEE Trans. on Multimedia}, 15(7):1602--1615, 2013.

\bibitem{DorinComaniciuCVPR00}
D.~Comaniciu, V.~Ramesh, and P~Meer.
\newblock Real-time tracking of non-rigid objects using mean shift.
\newblock In {\em CVPR}, 2000.

\bibitem{YangCongIEEETCSVT15}
Y.~Cong, B.~Fan, J.~Liu, J.~Luo, and H.~Yu.
\newblock Speeded up low rank online metric learning for object tracking.
\newblock {\em IEEE Trans. on CSVT}, 25(6):922--934, 2015.

\bibitem{FranklinCCrowACMSIGGRAPH84}
F.~C. Crow.
\newblock Summed-area tables for texture mapping.
\newblock In {\em Proceedings of the ACM SIGGRAPH}, 1984.

\bibitem{NavneetDalalCVPR05}
N.~Dalal and B.~Triggs.
\newblock Histograms of oriented gradients for human detection.
\newblock In {\em CVPR}, 2005.

\bibitem{MatthiasDantoneIEEETPAMI14}
M.~Dantone, J.~Gall, C.~Leistner, and L.~V. Gool.
\newblock Body parts dependent joint regressors for human pose estimation in
  still images.
\newblock {\em IEEE Trans. on PAMI}, 36(11):2131--2143, 2014.

\bibitem{JonathanDeutscherIJCV05}
J.~Deutscher and I.~Reid.
\newblock Articulated body motion capture by stochastic search.
\newblock {\em IJCV}, 61(2):185--205, 2005.

\bibitem{PedroFFelzenszwalbIJCV05}
P.~F. Felzenszwalb and D.~P. Huttenlocher.
\newblock Pictorial structures for object recognition.
\newblock {\em IJCV}, 61(1):55--79, 2005.

\bibitem{VittorioFerrariCVPR08}
V.~Ferrari, M.~Mar\'{\i}n-Jim\'{e}nez, and A.~Zisserman.
\newblock Progressive search space reduction for human pose estimation.
\newblock In {\em CVPR}, 2008.

\bibitem{JingMingGuoIEEETCSVT12}
J.-M. Guo, Y.-F. Liu, C.-H. Chang, and H.-S. Nguyen.
\newblock Improved hand tracking system.
\newblock {\em IEEE Trans. on CSVT}, 22(5):693--701, 2012.

\bibitem{AlexandreHeiliIEEETIP14}
A.~Heili, A.~L\'{o}pez-M\'{e}ndez, and J.-M. Odobez.
\newblock Exploiting long-term connectivity and visual motion in crf-based
  multi-person tracking.
\newblock {\em IEEE Trans. on Image Processing}, 23(7):3040--3056, 2014.

\bibitem{RaduHoraudIEEETPAMI08}
R.~Horaud, M.~Niskanen, G.~Dewaele, and E.~Boyer.
\newblock Human motion tracking by registering an articulated surface to 3-d
  points and normals.
\newblock {\em IEEE Trans. on PAMI}, 31(11):158--163, 2008.

\bibitem{YuGangJiangIEEETM15}
Y.-G. Jiang, Q.~Dai, T.~Mei, Y.~Rui, and S.-F. Chang.
\newblock Super fast event recognition in internet videos.
\newblock {\em IEEE Trans. on Multimedia}, 17(8):1174--1186, 2015.

\bibitem{ShanonXJuICAFGR96}
S.~X. Ju, M.~J. Black, and Y.~Yacoob.
\newblock Cardboard people: A parameterized model of articulated image motion.
\newblock In {\em 2nd Int. Conf. on Automatic Face- and Gesture-Recognition},
  1996.

\bibitem{MartinKiefelECCV14}
M.~Kiefel and P.~V. Gehler.
\newblock Human pose estimation with fields of parts.
\newblock In {\em ECCV}, pages 331--346, 2014.

\bibitem{JunseokKwonIEEETPAMI13a}
J.~Kwon and K.~M. Lee.
\newblock Highly nonrigid object tracking via patch-based dynamic appearance
  modeling.
\newblock {\em IEEE Trans. on PAMI}, 35(10):2427--2441, 2013.

\bibitem{SvetlanaLazebnikBMVC04}
S.~Lazebnik, C.~Schmid, and J.~Ponce.
\newblock Semi-local affine parts for object recognition.
\newblock In {\em BMVC}, 2004.

\bibitem{AndreasMLehrmannICCV13}
A.~M. Lehrmann, P.~V. Gehler, and S.~Nowozin.
\newblock A non-parametric bayesian network prior of human pose.
\newblock In {\em ICCV}, pages 1281--1288, 2013.

\bibitem{DavidGLoweIJCV04}
D.~G. Lowe.
\newblock Distinctive image features from scale-invariant keypoints.
\newblock {\em IJCV}, 60(2):91--110, 2004.

\bibitem{EmilioMaggioWiley11}
E.~Maggio and A.~Cavallaro.
\newblock {\em Video Tracking: Theory and Practice}.
\newblock Wiley, 2011.

\bibitem{DennisParkICCV11}
D.~Park and D.~Ramanan.
\newblock N-best maximal decoders for part models.
\newblock In {\em ICCV}, 2011.

\bibitem{HyunSooParkICCV11}
H.~S. Park and Y.~Sheikh.
\newblock 3d reconstruction of a smooth articulated trajectory from a monocular
  image sequence.
\newblock In {\em ICCV}, 2011.

\bibitem{TomasPfisterACCV14}
T.~Pfister, K.~Simonyan, J.~Charles, and A.~Zisserman.
\newblock Deep convolutional neural networks for efficient pose estimation in
  gesture videos.
\newblock In {\em ACCV}, 2014.

\bibitem{Lai-ManPoTENCON96}
L.-M. Po and C.~Cheung.
\newblock A new center-biased orthogonal search algorithm for fast block motion
  estimation.
\newblock In {\em TENCON}, 1996.

\bibitem{JensPuweinECCV14}
J.~Puwein, L.~Ballan, R.~Ziegler, and M.~Pollefeys.
\newblock Foreground consistent human pose estimation using branch and bound.
\newblock In {\em ECCV}, pages 315--330, 2014.

\bibitem{VarunRamakrishnaCVPR13}
V.~Ramakrishna, T.~Kanade, and Y.~Sheikh.
\newblock Tracking human pose by tracking symmetric parts.
\newblock In {\em CVPR}, pages 3728--3735, June 2013.

\bibitem{VarunRamakrishnaECCV14}
V.~Ramakrishna, D.~Munoz, M.~Hebert, J.~A. Bagnell, , and Y.~Sheikh.
\newblock Pose machines: Articulated pose estimation via inference machines.
\newblock In {\em ECCV}, pages 33--47, 2014.

\bibitem{DevaRamananCVPR05}
D.~Ramanan, D.~A. Forsyth, and A.~Zisserman.
\newblock Strike a pose: Tracking people by finding stylized poses.
\newblock In {\em CVPR}, 2005.

\bibitem{ICDPose}
Soumitra Samanta and Bhabatosh Chanda.
\newblock
  http://www.isical.ac.in/$\sim$vlrg/sites/default/files/soumitra/site/ss\_icdpose.html.

\bibitem{BenjaminSappCVPR11}
B.~Sapp, D.~Weiss, and B.~Taskar.
\newblock Parsing human motion with stretchable models.
\newblock In {\em CVPR}, 2011.

\bibitem{ChristianSchmaltzMVA12}
C.~Schmaltz, B.~Rosenhahn, T.~Brox, and J.~Weickert.
\newblock Region-based pose tracking with occlusions using 3d models.
\newblock {\em MVA}, 23(3):557--577, 2012.

\bibitem{LauraSevilla-LaraCVPR12}
L.~Sevilla-Lara and E.~Learned-Miller.
\newblock Distribution fields for tracking.
\newblock In {\em CVPR}, 2012.

\bibitem{LeonidSigalIJCV12}
L.~Sigal, M.~Isard, H.~Haussecker, and M.~J. Black.
\newblock Loose-limbed people: Estimating 3d human pose and motion using
  non-parametric belief propagation.
\newblock {\em IJCV}, 98(1):15--48, 2012.

\bibitem{ArnoldWMSmeuldersIEEETPAMI14}
A.~W.~M. Smeulders, D.~M. Chu, R.~Cucchiara, S.~Calderara, A.~Dehghan, and
  M.~Shah.
\newblock Visual tracking: An experimental survey.
\newblock {\em IEEE Trans. on PAMI}, 36(7):1442--1468, 2014.

\bibitem{JiTaoIJIG07}
J.~Tao, Y.-P. Tan, and W.~Lu.
\newblock Robust color object tracking with application to people monitoring.
\newblock {\em International Journal of Image and Graphics}, 7(2):227--254,
  2007.

\bibitem{AlexanderToshevCVPR14}
A.~Toshev and C.~Szegedy.
\newblock Deeppose: Human pose estimation via deep neural networks.
\newblock In {\em CVPR}, June 2014.

\bibitem{PaulViolaCVPR01}
P.~Viola and M.~Jones.
\newblock Rapid object detection using a boosted cascade of simple features.
\newblock In {\em CVPR}, 2001.

\bibitem{ShuWangIEEETIP14}
S.~Wang, H.~Lu, F.~Yang, and M.-H. Yang.
\newblock Robust superpixel tracking.
\newblock {\em IEEE Trans. on Image Processing}, 23(4):1639--1651, 2014.

\bibitem{YuweiWuIEEETIP15}
Y.~Wu, M.~Pei, M.~Yang, J.~Yuan, and Y.~Jia.
\newblock Robust discriminative tracking via landmark-based label propagation.
\newblock {\em IEEE Trans. on Image Processing}, 24(5):1510--1523, 2015.

\bibitem{LexingXieIEEETM13}
L.~Xie, A.~Natsev, X.~He, J.~Kender, M.~Hill, and J.~R. Smith.
\newblock Tracking large-scale video remix in real-world events.
\newblock {\em IEEE Trans. on Multimedia}, 15(6):1244--1254, 2013.

\bibitem{TengXuAVSS12}
T.~Xu, P.~Peng, X.~Fang, C.~Su, Y.~Wang, Y.~Tian, W.~Zeng, and T.~Huang.
\newblock Single and multiple view detection, tracking and video analysis in
  crowded environments.
\newblock In {\em AVSS}, pages 494--499, 2012.

\bibitem{YiYangIEEETPAMI13}
Y.~Yang and D.~Ramanan.
\newblock Articulated human detection with flexible mixtures of parts.
\newblock {\em IEEE Trans. on PAMI}, 35(12):2878--2890, 2013.

\bibitem{BangpengYaoECCV12}
B.~Yao and F.-F. Li.
\newblock Action recognition with exemplar-based 2.5d graph matching.
\newblock In {\em ECCV}, 2012.

\bibitem{XinguoYuIEEETM06}
X.~Yu, H.~W. Leong, C.~Xu, and Q.~Tian.
\newblock Trajectory-based ball detection and tracking in broadcast soccer
  video.
\newblock {\em IEEE Trans. on Multimedia}, 8(6):1164--1178, 2006.

\bibitem{YuanYuanIEEETM15}
Y.~Yuan, H.~Yang, Y.~Fang, and W.~Lin.
\newblock Visual object tracking by structure complexity coefficients.
\newblock {\em IEEE Trans. on Multimedia}, 17(8):1125--1136, 2015.

\bibitem{ChaZhangICPR06}
C.~Zhang and Y.~Rui.
\newblock Robust visual tracking via pixel classification and integration.
\newblock In {\em ICPR}, pages 37--42, 2006.

\bibitem{KaihuaZhangECCV14}
K.~Zhang, L.~Zhang, Q.~Liu, D.~Zhang, and M.-H. Yang.
\newblock Fast visual tracking via dense spatio-temporal context learning.
\newblock In {\em ECCV}, 2014.

\bibitem{KaihuaZhangIEEETPAMI14}
K.~Zhang, L.~Zhang, and M.-H. Yang.
\newblock Fast compressive tracking.
\newblock {\em IEEE Trans. on PAMI}, 36(10):2002--2015, 2014.

\bibitem{ShunliZhangIEEETM15}
S.~Zhang, X.~Yu, Y.~Sui, S.~Zhao, and L.~Zhang.
\newblock Object tracking with multi-view support vector machines.
\newblock {\em IEEE Trans. on Multimedia}, 17(3):265--278, 2015.

\bibitem{ZhengZhangIEEETM13}
Z.~Zhang, H.~S. Seah, C.~K. Quah, and J.~Sun.
\newblock Gpu-accelerated real-time tracking of full-body motion with
  multi-layer search.
\newblock {\em IEEE Trans. on Multimedia}, 15(1):106--119, 2013.

\bibitem{WeiZhongIEEETIP14}
W.~Zhong, H.~Lu, and M.-H. Yang.
\newblock Robust object tracking via sparse collaborative appearance model.
\newblock {\em IEEE Trans. on Image Processing}, 23(5):2356--2368, 2014.

\end{thebibliography}
\end{document}